\documentclass[preprint,11pt]{article}
\usepackage{geometry}
\newgeometry{vmargin={30mm}, hmargin={30mm,30mm}}
\usepackage{hyperref}
\usepackage{graphicx} %% for loading jpg figures
\usepackage{multirow}
\usepackage{mathtools}
\usepackage{amsmath}
\usepackage{amsfonts}
\usepackage{subcaption}
\usepackage{xcolor}
\usepackage{caption}
\usepackage{subcaption}
\usepackage{booktabs}

\title{Beyond Statistical Similarity: Rethinking Metrics for Deep Generative Models in Engineering Design}

\author{
  Lyle Regenwetter$^1$, Akash Srivastava$^2$, Dan Gutfreund$^2$, Faez Ahmed$^1$\\
  Massachusetts Institute of Technology$^1$, MIT-IBM Watson AI Lab$^2$ \\
  \texttt{regenwet@mit.edu, Akash.Srivastava@ibm.com,}\\ \texttt{dgutfre@us.ibm.com, faez@mit.edu} \\
}

\date{October 17, 2023}

\newcommand{\etal}{{\em et~al.}}

\newcommand{\RNum}[1]{\uppercase\expandafter{\romannumeral #1\relax}}

\begin{document}

\maketitle
%%%%%%%%%%%%%%%%%%%%%%%%%%%%%%%%%%%%%%%%%%%%%%%%%%%%%%%%%%%%%%%%%%%%%%
\begin{abstract}
Deep generative models such as Variational Autoencoders (VAEs), Generative Adversarial Networks
(GANs), Diffusion Models, and Transformers, have shown great promise in a variety of applications,
including image and speech synthesis, natural language processing, and drug discovery. However, when
applied to engineering design problems, evaluating the performance of these models can be challeng-
ing, as traditional statistical metrics based on likelihood may not fully capture the requirements of
engineering applications. This paper doubles as a review and practical guide to evaluation metrics
for deep generative models (DGMs) in engineering design. We first summarize the well-accepted
‘classic’ evaluation metrics for deep generative models grounded in machine learning theory. Using
case studies, we then highlight why these metrics seldom translate well to design problems but see
frequent use due to the lack of established alternatives. Next, we curate a set of design-specific metrics
which have been proposed across different research communities and can be used for evaluating
deep generative models. These metrics focus on unique requirements in design and engineering,
such as constraint satisfaction, functional performance, novelty, and conditioning. Throughout our
discussion, we apply the metrics to models trained on simple-to-visualize 2-dimensional example
problems. Finally, we evaluate four deep generative models on a bicycle frame design problem and
structural topology generation problem. In particular, we showcase the use of proposed metrics to
quantify performance target achievement, design novelty, and geometric constraints. We publicly
release the code for the datasets, models, and metrics used throughout the paper at \href{http://decode.mit.edu/projects/metrics/}{decode.mit.edu/projects/metrics/}. 

\end{abstract}

\section{Introduction}
Deep generative models (DGMs) have seen explosive growth across engineering design disciplines in recent years. DGMs like Generative Adversarial Networks (GAN)~\cite{goodfellow2014generative} and Variational Autoencoders (VAE)\cite{kingma2013auto} have dominated image generation problems since 2014, but only bridged the gap to the design community in 2016~\cite{regenwetter2022deep}. DGMs have since been applied across design domains to problems such as optimal topology generation, airfoil synthesis, and metamaterial design. As promising new methods for image synthesis like diffusion models~\cite{croitoru2022diffusion, yang2022diffusion} are introduced in other machine learning fields, researchers in design adapt them to solve challenging design problems~\cite{maze2022topodiff}. Similarly, transformers, a leading class of generative models for sequences, have dominated natural language generation for years~\cite{devlin2018bert, raffel2020exploring, brown2020language}, and have seen extensive use in the textual generation of design concepts~\cite{zhu2023generative, siddharth2022natural}. 

DGMs are powerful learners, boasting an unparalleled ability to process and understand complex data distributions and mimic them through batches of synthetic data. In the context of data-driven design, these `data distributions' are often comprised of a collection of existing designs that lie in some multi-dimensional design manifold in the same way that a collection of points would form a density distribution in a Euclidean space. From this perspective, it's clear why DGMs are promising data-driven designers. They can study collections of existing designs, understand their distribution, and generate new ones that should belong in the same manifold but do not yet exist. This ability of a DGM to learn and match a distribution is often measured using statistical similarity (i.e., how similar is the distribution of generated designs to the dataset?). 

\begin{figure}[!htb]
     \centering
     \begin{subfigure}[b]{0.40\textwidth}
         \centering
         \includegraphics[width=\textwidth]{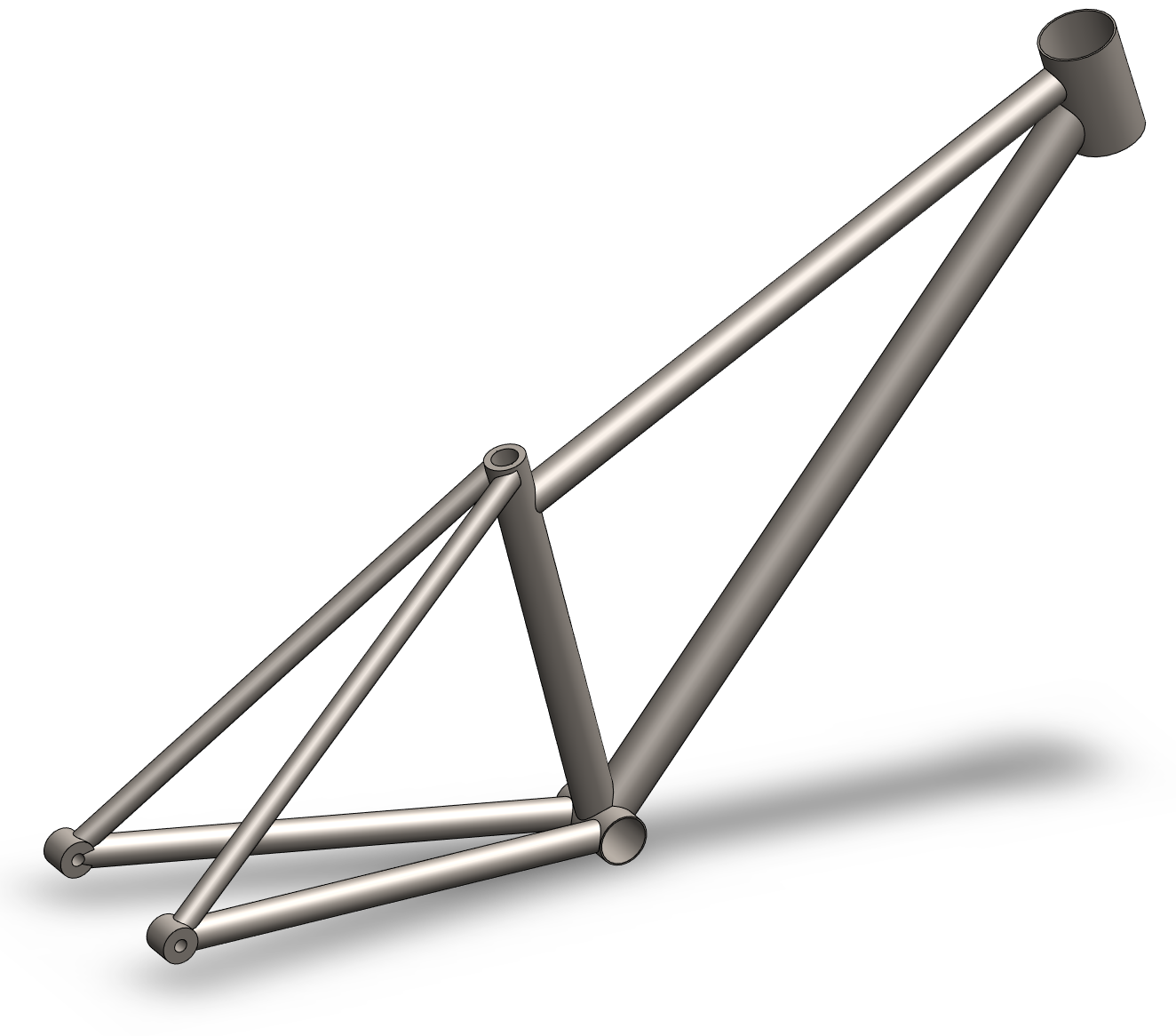}
         \caption{Deflection: $0.05\,cm$}
         \label{fig:frame1}
     \end{subfigure}
     \begin{subfigure}[b]{0.40\textwidth}
         \centering
         \includegraphics[width=\textwidth]{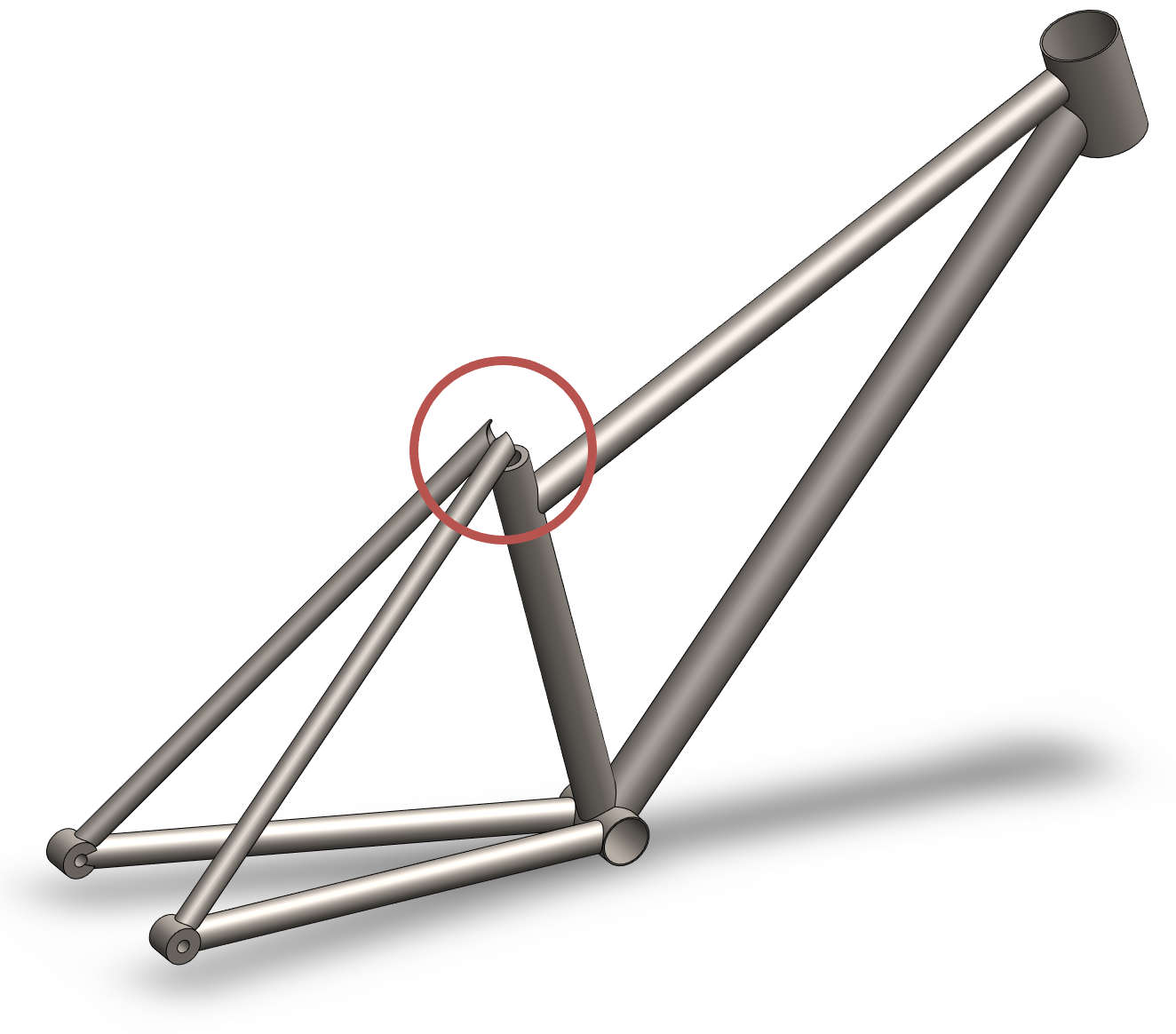}
         \caption{Deflection: $10.46\,cm$}
         \label{fig:frame2}
     \end{subfigure}

    \caption{Two very similar bike frames adapted from~\cite{regenwetter2022framed} with drastically different structural performance. By most distance metrics, these bike frames would be the most similar designs among a dataset of thousands. Yet, due to the disconnected geometry highlighted, they experience deflections that differ by over two orders of magnitude when subjected to the in-plane loading scenario.}
    \label{fig:twobikes}
\end{figure}

While DGMs' amazing distribution-matching ability is often beneficial, it is limited because, in many design problems, we desire designs to be significantly unique or distinct from existing designs. Additionally, even if distribution-matching is desirable, it is often secondary to meeting problem constraints and achieving functional performance targets. As a result, relying solely on similarity as an objective can lead to misguided design efforts, since very similar designs can have drastically different performances, as illustrated in Figure~\ref{fig:twobikes}. Historically, design researchers have failed to account for this gap when selecting evaluation metrics, often opting for the classic similarity-based metrics that are prominent in other machine learning domains. In this paper, we provide an exposition of evaluation metrics for DGMs, which we show are important for design practitioners and design automation researchers. We intend for the metrics presented to apply to a variety of DGMs for design, and we do not go into detail about specific model architectures. 
For an introduction to DGMs and an overview of design engineering research using DGMs to date, we refer readers to a review paper on DGMs in engineering design~\cite{regenwetter2022deep}. 

We broadly divide metrics into five categories which each address an important facet of engineering design. The first of these is similarity, the class of metrics commonly used to evaluate DGMs. Though often overused, similarity is still important, and we present several perspectives and metrics to consider various aspects of similarity. The second category is design exploration, the idea that generated designs are often desired to be unique and varied. The third category is constraint satisfaction, the idea that generated designs must often adhere to a set of explicit or implicit constraints to be considered a valid design. The fourth is design quality, the idea that generated designs often have associated functional performance attributes that designers want to optimize. Finally, the last category is conditioning, the idea that conditional generative models should respect and adhere to their conditioning information. Although we discuss a wide variety of evaluation criteria within the five focus areas, this is not an exhaustive review. There are other aspects of DGM performance and evaluation techniques that are particularly important in design, such as latent space disentanglement and human evaluation, which we discuss in Appendix~\ref{sec:misc}. This appendix also lists other common considerations for DGMs broadly, which may also be relevant in design problems. 

% \textcolor{red}{Mention the obvious metrics somewhere--- Mean Absolute Error (MAE),
% Mean Squared Error (MSE),
% Root Mean Squared Error (RMSE),
% R² (R-Squared).}

\section{Relevant Reviews of Evaluation Metrics}
Review papers for DGM evaluation metrics are numerous in image and text generation~\cite{borji2019pros, borji2022pros, gatt2018survey, dong2022survey}. Within engineering design, a few research papers~\cite{maze2022topodiff, chen2022inverse} and code repositories\footnote{https://pypi.org/project/midbench/} discuss metrics for DGMs in specific design sub-disciplines, such as topology optimization and aerodynamic design. To our knowledge, however, the existing body of research lacks a dedicated discussion of applicable metrics for DGMs which generally apply across design disciplines. This paper aims to address this gap and highlight the importance of applying the right metrics for Generative AI applications. We will organize our discussion of relevant work according to several focus areas of this paper: statistical similarity, design exploration, constraint satisfaction, and design quality.

%\subsection{Reviews of Metrics in Vision and Natural Language}
\paragraph{Existing Literature in Similarity Metrics}
Image generation~\cite{goodfellow2014generative,  kingma2013auto, karras2019style, karras2020analyzing, brock2018large} and natural language generation~\cite{devlin2018bert, raffel2020exploring, brown2020language, openai2023gpt4} remain dominant research thrusts for DGMs. In these fields, the overwhelming majoring of metrics focus on statistical similarity. As such, similarity-related metrics for generative models in these domains have received much attention and careful consideration. Borji~\cite{borji2019pros, borji2022pros} provides two complementary reviews which outline a few evaluation metrics for GANs, though most of these metrics generalize to other generative models in computer vision problems. Gatt \& Krahmer~\cite{gatt2018survey} and Dong~\etal~\cite{dong2022survey} review the state of natural language generation, including detailed discussions of similarity-based evaluation metrics. 

Design domains heavily focused on image-based data have also seen reviews of metrics focusing on statistical similarity. For example, Shah~\etal~\cite{shah2022survey} reviews evaluation methods for generative models in synthetic microstructure images. While the metrics discussed in the mentioned papers focus on statistical similarity, this paper argues that thinking beyond statistical similarity to include factors such as performance, diversity, and constraints is crucial for design and engineering.

\paragraph{Existing Literature in Design Exploration Metrics}
The ability of a model to explore is often captured through diversity and novelty. Diversity and novelty are prevalent concepts in design ideation and can be challenging to evaluate. Mueller \& Ochsendorf~\cite{mueller2015combining} analyze numerous metrics for design diversity quantification. While useful, these design metrics are often set up to evaluate ``generic'' designs and may not be sufficient to measure ``AI'' generated designs. In particular, many metrics struggle with the extreme non-convexity and high dimensionality of the complicated design spaces typically learned by DGMs. This makes their use as an evaluation metric challenging. 

Diversity metrics are also found in optimization literature~\cite{Riquelme2015performance}. However, many metrics from optimization, such as the hypervolume metric, could be used to evaluate multiple objectives, such as diversity and functional performance. Specialized metrics introduced for DGMs may be better able to evaluate diversity and novelty for complex design distributions~\cite{chen2021padgan} and decouple diversity from other performance considerations. 

\paragraph{Existing Literature in Constraint Satisfaction Metrics}
The broad diversity of design constraints across various domains has led to a largely domain-specific approach to constraint satisfaction. 
For example, Bilodeau~\etal~\cite{bilodeau2022generative} present several interesting constraint-related metrics for DGMs in their review of generative models for molecular discovery\footnote{This paper also discusses other metrics, such as rediscovery, which we discuss in this paper}. In some design subdisciplines, similarity-based metrics are being used to indirectly infer constraint satisfaction~\cite{shah2022survey}.
In our paper, however, we introduce a series of generalized, domain-independent constraint satisfaction metrics that surpass the simple use of similarity.

\paragraph{Existing Literature in Design Quality Metrics} Design quality (functional performance) is a ubiquitous consideration across design disciplines. Many relevant functional performance metrics can be found in design optimization literature, as optimization is heavily focused on maximizing functional performance. Riquelme~\etal~\cite{Riquelme2015performance} review evaluation metrics for multi-objective optimization, documenting their popularity in optimization research and classifying them by type. In this paper, we show that many metrics developed to evaluate optimization algorithms are easily adapted to evaluate DGMs and should be adopted by researchers. To better handle inverse design problems, Regenwetter \& Ahmed~\cite{regenwetter2022design} propose several evaluation metrics focused on evaluating functional performance in design problems where a performance target is given.

\paragraph{Contributions}
In the context of existing work, this paper makes the following contributions:
\begin{enumerate}
    \item We provide a structured review and practical guide on metrics used to evaluate deep generative models in engineering design problems.
    \item We draw attention to the inherent shortcomings of statistical similarity and advocate for a collection of metrics to assess design exploration, constraints, quality, and conditioning needs. These metrics are derived from a broad range of metrics proposed in diverse fields, such as multi-objective optimization, natural image generation, and molecule synthesis.
    \item We train a variety of deep generative models, including GANs, VAEs, MO-PaDGANs, DTAI-GANs, cVAEs, cGANs, and Denoising Diffusion Probabilistic Models on two-dimensional examples to illustrate the diverse ways deep generative models can be evaluated under various conditions. 
    \item We introduce a design case study focusing on the synthesis of bike frames, with an emphasis on performance, diversity, and constraints. In this context, we discuss the selection and application of the most appropriate metrics.
    \item We present an additional design case study related to structural topology generation, where we demonstrate how to evaluate and compare two cutting-edge conditional diffusion models using the discussed evaluation metrics.
\end{enumerate}

\section{Background} \label{background}
% Many data-driven design practitioners are intimately familiar with the challenges of model selection.
% Say a designer wants to train a model to generate a set of optimal structural topologies that meet several manufacturability constraints and performance targets. The designer trains a BigGAN~\cite{brock2018large} and a StyleGAN2~\cite{karras2020analyzing} and generates thousands of design options using each model. How should the designer select which model to use? The go-to evaluation method for many practitioners using these types of GANs would be the Fr\'echet Inception Distance (FID)~\cite{heusel2017gans}, a metric that attempts to quantify similarity to the training dataset. FID can indicate which model generates sets of designs that are most similar to the dataset, but it doesn't provide information on how well they meet specific design constraints, performance targets, or diversity goals. Additionally, FID is calibrated for natural image datasets, so it may not accurately measure similarity for other types of data or structures. 
% For design researchers looking to train deep generative models (DGMs), commonly used evaluation metrics should be revisited. 
In this paper, we present a structured guide to help researchers select better evaluation metrics for design problems, focusing on five main facets: similarity, diversity, constraint satisfaction, functional performance, and conditioning. However, before introducing individual metrics in Section~\ref{similarity}, we will discuss terminology and several background concepts. Notably, we will discuss common requirements and prerequisites needed to apply the metrics as well as classifications to categorize metrics. We include a tabular summary of the metrics, their requirements, and their categorizations in Table~\ref{tab:overview}. 

\begin{table}[!htb]
\centering
\caption{Overview of metrics discussed in the paper. For each metric, we discuss: 1. The requirements and auxiliary computational cost necessary to evaluate the metric (key: \textbf{Aux} -- Auxiliary predictive task \& training, \textbf{CL} -- Clustering method, \textbf{CFC} -- Closed-Form Constraints, \textbf{Cond} -- Condition parameter calculation, \textbf{Const} -- Constraint violation test, \textbf{Dist} -- Distance metric, \textbf{DP} -- Differentiable method to calculate design performance, \textbf{Emb} -- Vector embedding, \textbf{Inv} -- Dataset of invalid designs, \textbf{Perf} -- Method to calculate design performance). 2. Whether the metric characterizes \textbf{points} or \textbf{sets} of generated samples. 3. Whether the metric uses a reference set (\textbf{binary}) or not (\textbf{unary}). 4. Relevant \textbf{spaces} to which the metric can reasonably apply. 5. Whether the metric depends on \textbf{hyperparameters}. 6. The metric's \textbf{bounds}. 7. The optimal \textbf{direction} of the metric. Notes: *Assuming L2 Norm (Euclidean) distance kernel. Distance kernel impacts hyperparameters, bounds, and objective direction. For example, switching to an RBF kernel would typically result in bounds of $[0,1]$, flipping the direction of the objective, as well as adding a hyperparameter to tune. ** Depending on the clustering method used, a distance metric may suffice instead of a full embedding. *** Problem parameters such as reference points or sets, objective weights, and DTAI priority parameters are not considered to be hyperparameters.}
\resizebox{\textwidth}{!}{
\begin{tabular}{ccccccccc}
\toprule
Category & Metric: & \begin{tabular}[c]{@{}c@{}}Requirements/ \\ Eval Costs:\end{tabular} & \begin{tabular}[c]{@{}c@{}}Point/ \\ Set:\end{tabular} & \begin{tabular}[c]{@{}c@{}}Unary/ \\ Binary:\end{tabular} & Space: & \begin{tabular}[c]{@{}c@{}}Hyperpara-\\ meters***:\end{tabular} & \begin{tabular}[c]{@{}c@{}}Bounds/ \\ Values:\end{tabular} & Direction: \\
\midrule
\multirow{8}{*}{\begin{tabular}[c]{@{}c@{}}Similarity and \\ Distribution \\ Matching\end{tabular}} & Statistical Distance/Divergence & Depends & Set & Binary & Either & Depends & $[0,\infty\}$ & Minimize \\
 & Precision-Recall Curves & Emb**, CL & Set & Binary & Either & Yes & N/A & N/A \\
 & Precision ($F_{\beta<<1}$) & Emb**, CL & Set & Binary & Either & Yes & $[0,1]$ & Maximize \\
 & Nearest Datapoint* & Dist & Point & Binary & Either & No & $[0,\infty\}$ & Minimize \\
 & Recall ($F_{\beta>>1}$) & Emb**, CL & Set & Binary & Either & Yes & $[0,1]$ & Maximize \\
 & Nearest Generated Sample* & Dist & Set & Binary & Either & Yes & $[0,\infty\}$ & Minimize \\
 & Rediscovery* & Dist & Set & Binary & Either & Yes & $[0,\infty\}$ & Minimize \\
 & ML Efficacy & Aux & Set & Binary & Either & Yes & Varies & Varies \\
 \midrule
\multirow{7}{*}{\begin{tabular}[c]{@{}c@{}}Novelty \& \\ Diversity\end{tabular}} & Inter-Sample Distance* & Dist & Point & Unary & Either & No & $[0,\infty\}$ & Maximize \\
 & Nearest Datapoint* & Dist & Point & Binary & Either & No & $[0,\infty\}$ & Maximize \\
 & Distance to Centroid* & Emb & Point & Unary & Either & No & $[0,\infty\}$ & Maximize \\
 & Entropy & Depends & Set & Unary & Either & Depends & $[0,\infty\}$ & Minimize \\
 & DPP Diversity Score & Dist & Set & Unary & Either & Yes & $[0,\infty\}$ & Minimize \\
 & Smallest Enclosing Hypersphere & Emb & Set & Unary & Either & No & $[0,\infty\}$ & Maximize \\
 & Convex Hull & Emb & Set & Unary & Either & No & $[0,\infty\}$ & Maximize \\
 \midrule
\multirow{5}{*}{\begin{tabular}[c]{@{}c@{}}Design \\ Constraints\end{tabular}} & Constraint Satisfaction & Const & Point & Unary & N/A & No & $[0,1]$ & Maximize \\
 & Constraint Satisfaction Rate & Const & Point & Unary & N/A & No & $[0,1]$ & Maximize \\
 & Signed Dist. to Constraint Boundary* & CFC & Point & Unary & Design & No & $\{-\infty,\infty\}$ & Maximize \\
 & Predicted Constraint Satisfaction & Inv, Aux & Point & Binary & Design & Yes & $[0,1]$ & Maximize \\
 & Nearest Invalid Datapoint* & Inv, Dist & Point & Binary & Design & No & $[0,\infty\}$ & Maximize \\
 \midrule
\multirow{7}{*}{\begin{tabular}[c]{@{}c@{}}Performance \\ and Target \\ Achievement\end{tabular}} & Hypervolume & Perf & Set & Unary & Perf. & No & $[0,k]$ & Maximize \\
 & Target Achievement & Perf & Point & Unary & Perf. & No & $[0,1]$ & Maximize \\
 & Target Achievement Rate & Perf & Point & Unary & Perf. & No & $[0,1]$ & Maximize \\
 & Signed Distance to Target* & Perf & Point & Unary & Perf. & No & $\{-\infty,\infty\}$ & Maximize \\
 & Design Target Achievement Index & Perf & Point & Unary & Perf. & No & $[0,1\}$ & Maximize \\
 & Generational Distance* & Perf & Point & Binary & Perf. & No & $[0,\infty\}$ & Minimize \\
 & Inst/Cum. Optimality Gap* & DP & Point & Unary & Perf. & Yes & $[0,\infty\}$ & Minimize \\
 \midrule
\multirow{2}{*}{Conditioning} & Conditioning Adherence* & Cond & Point & Unary & Either & No & $[0,\infty\}$ & Minimize \\
 & Conditioning Reconstruction & Aux & Point & Binary & Either & Yes & $[0,\infty\}$ & Minimize\\
 \bottomrule
\end{tabular}
}
\label{tab:overview}

\end{table}

\subsection{A Note on Terminology}
In this paper, we broadly use the term `metrics' to loosely refer to `evaluation criteria,' not specifically distance metrics. To avoid loss of generality, we broadly use the term `samples' to refer to the output of a generative model. We also refer to the original data entries as `datapoints.' Though in generative design problems, `datapoints' are often existing designs and `samples' are often generated designs, this may not be universally true. In Section~\ref{Constraints}, we also refer to constraint-violating data entries as `invalid datapoints,' which typically correspond to existing design concepts that fail to meet some set of constraints or design requirements. 

\subsection{Calculating Distance Between Designs}
Many distance-based metrics presented in this paper require the ability to calculate distances amongst and in between datapoints and samples (Tab.~\ref{tab:overview}, col. 3). In some data modalities, calculating distances between designs can be a significant challenge. We present some strategies to calculate distances in different representation schemes in the appendix. In general, practitioners can choose to directly calculate distances in the original space and modality of the data, or can instead compute an embedding over which to calculate a distance. 

\subsection{Point vs. Set Metrics} 
Certain metrics, such as statistical distance metrics, measure the properties of a set while other metrics measure the properties of an individual point (Tab.~\ref{tab:overview}, col. 4). Though point metrics can also be aggregated to describe the set, set metrics cannot evaluate a single point. In design problems where designers need to select one or many `finalists' from a set of designs generated by a DGM, they must do so with point metrics. 

\subsection{Unary vs. Binary Metrics}
Unary metrics, such as hypervolume, generate a score based on a single design or a set of designs (Tab.~\ref{tab:overview}, col. 5). In contrast, binary metrics such as statistical divergence utilize an additional reference set in generating a score. Often, this reference set is the dataset itself.

\subsection{Design Spaces and Performance Spaces}
In design, we often frame a design as a point in some design space, without loss of generality across data modalities. However, we also often care about the functional performance of generated designs. We can similarly frame a design's multi-objective performance as a point in some multi-dimensional performance space. Many metrics can reasonably be evaluated in either space, but some only make sense in one (Tab.~\ref{tab:overview}, col. 6). 

\subsection{Hyperparameters}
Most metrics depend on some parameters, which must be decided before using the metric (Tab.~\ref{tab:overview}, col. 7). This can be viewed as an opportunity for practitioners to carefully consider the parameters they use and to report them clearly when evaluating models. This allows for fair and consistent comparison of models. Additionally, by standardizing parameter values, practitioners can ensure that the evaluation metrics are being used in a fair and unbiased manner. 

\section{Evaluating Statistical Similarity} \label{similarity}
Having touched on some broad classifications of metrics, we begin our detailed discussion with the first of our five main categories: Similarity. In classic machine learning theory and many classic DGM tasks, such as image generation and natural language generation, deep generative models have a single overarching objective: To generate convincing samples that are new, but generally indistinguishable from the dataset. We typically frame this as a distribution-matching problem, i.e. is the distribution over generated samples identical to the distribution over the dataset? Accordingly, the dominant evaluation metrics in both image synthesis (FID~\cite{heusel2017gans}, IS~\cite{salimans2016improved}, KID~\cite{binkowski2018demystifying}, etc.) and natural language generation (ROUGE~\cite{lin2004rouge}, BLEU~\cite{papineni2002bleu}, METEOR~\cite{banerjee2005meteor}, etc.) have focused on similarity. Historically, similarity has also been the central objective of deep generative models in design tasks~\cite{regenwetter2022deep}, since it enforces that generated designs have a general resemblance to the designs used to train the model. For example, if a hypothetical design practitioner trains a model on a dataset of centrifugal pumps, similarity should enforce that their model generates new centrifugal pump designs, rather than reciprocating pumps (or nonsensical designs). However, similarity will not enforce (nor necessarily encourage) that generated pumps are novel, functional, or performant. In this section, we discuss metrics and tools to evaluate similarity and reserve discussion about other objectives for later sections.

\begin{figure}[!htb]
     \centering
     \begin{subfigure}[b]{0.64\textwidth}
         \centering
         \includegraphics[width=\textwidth]{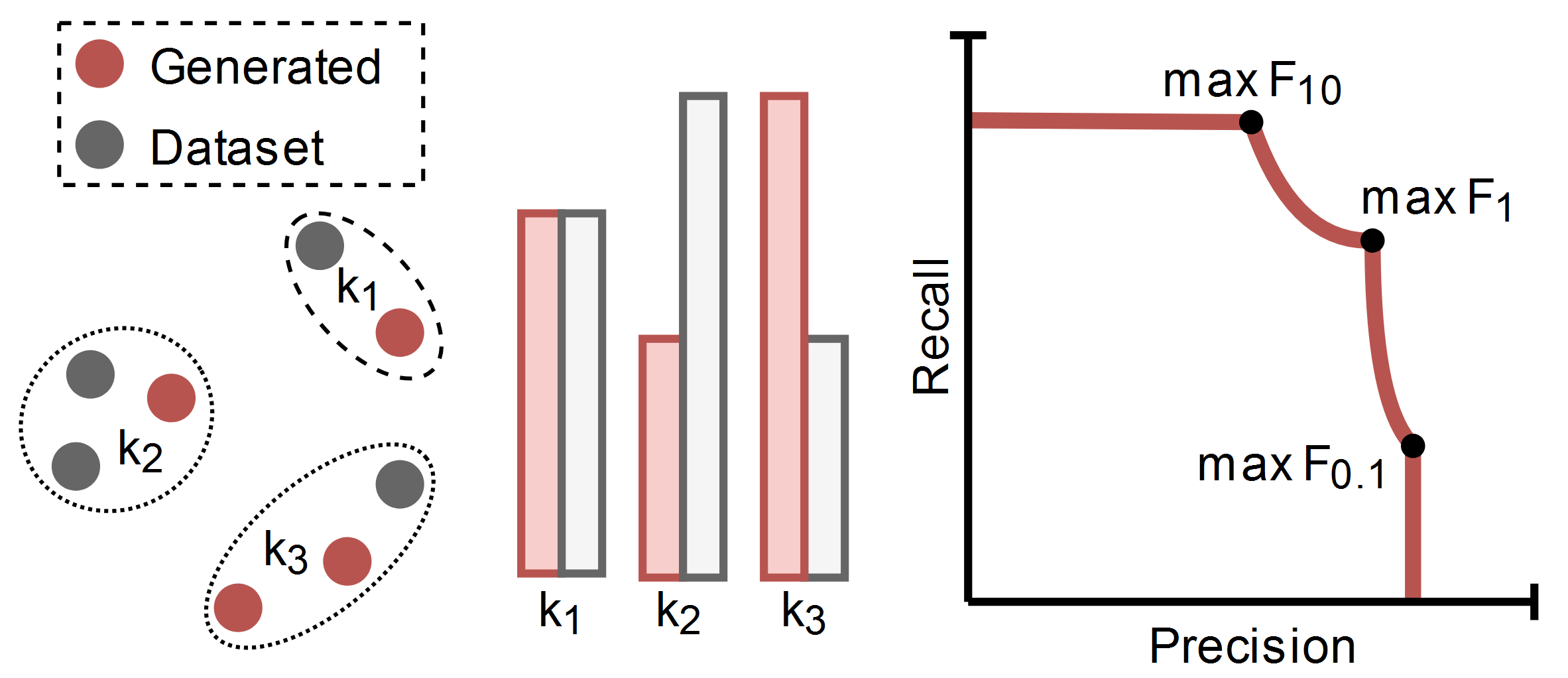}
         \caption{Precision-Recall Curves}
         \label{fig:PR}
     \end{subfigure}
     \begin{subfigure}[b]{0.32\textwidth}
         \centering
         \includegraphics[width=\textwidth]{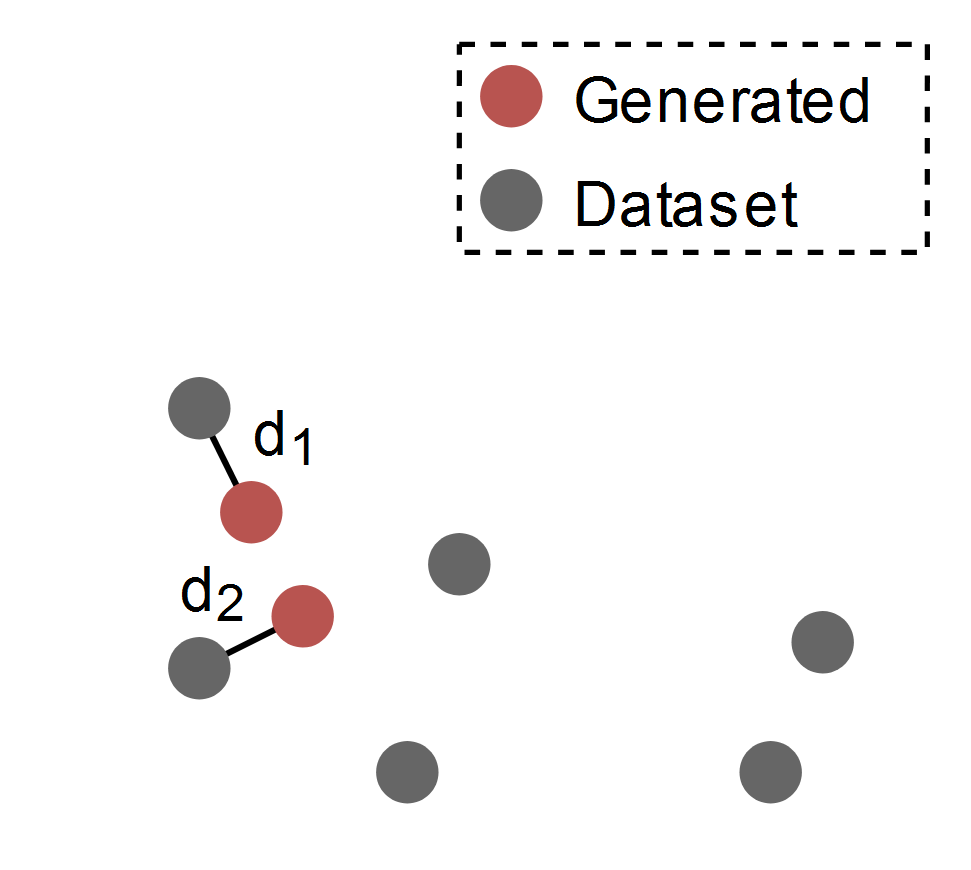}
         \caption{Nearest Datapoint}
         \label{fig:NDS}
     \end{subfigure}
     
     \begin{subfigure}[b]{0.32\textwidth}
         \centering
         \includegraphics[width=\textwidth]{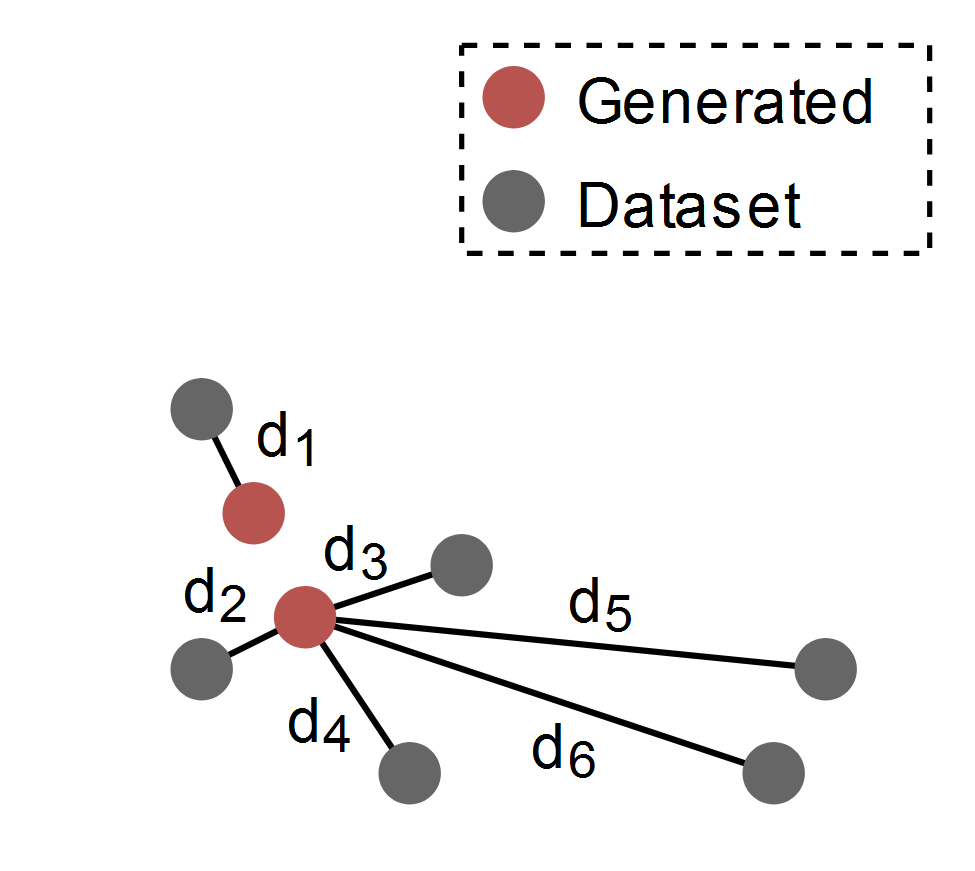}
         \caption{Nearest Generated Sample}
         \label{fig:NGS}
     \end{subfigure}
     \begin{subfigure}[b]{0.32\textwidth}
         \centering
         \includegraphics[width=\textwidth]{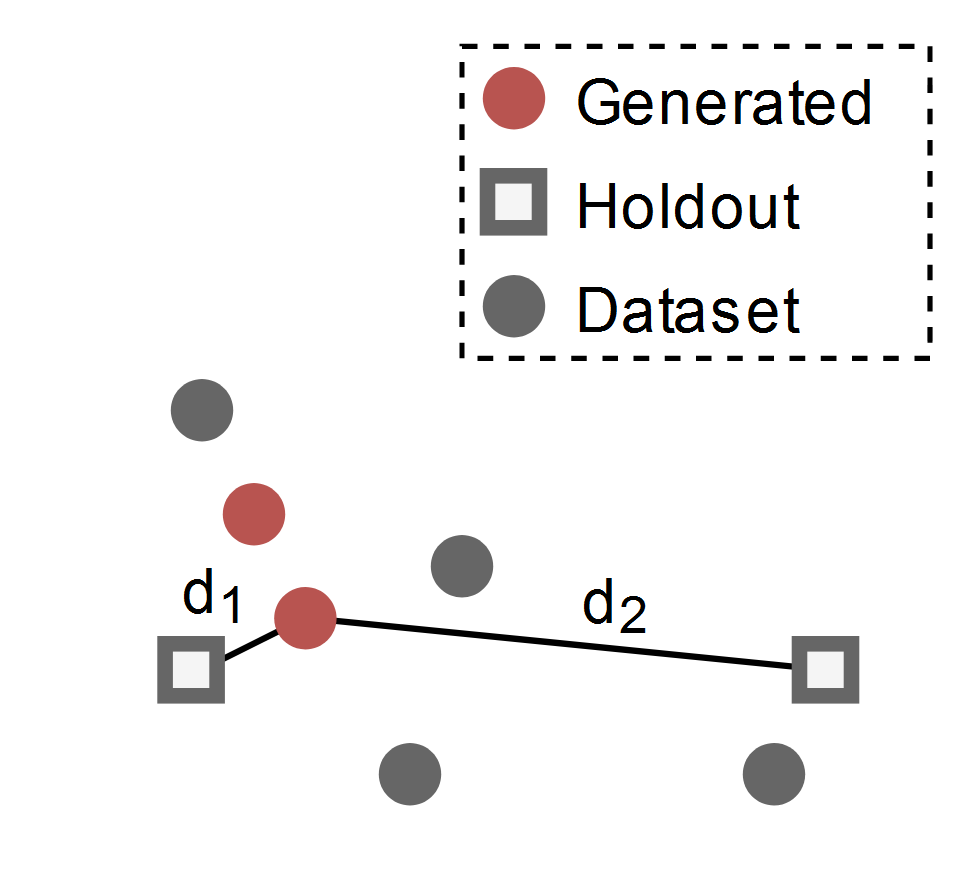}
         \caption{Rediscovery}
         \label{fig:Red}
     \end{subfigure}

    \caption{Illustrations of select similarity-related metrics.}
    \label{fig:similarity}
\end{figure}
\paragraph{Statistical Divergence/Distance Metrics}
Quantifying the discrepancy between two random variables is one of the central themes in machine learning, particularly in ML-based generative modeling~\cite{kingma2013auto, goodfellow2014generative, arjovsky2017wasserstein, arjovsky2017towards, ho2020denoising, dhariwal2021diffusion}. Within the field of deep generative modeling, the two classes of statistical discrepancy measures have been popular, namely $\phi$-divergences and Integral Probability Metrics (IPMs). 
Let $\mathbb{P}$ and $\mathbb{Q}$ be two probability distributions on a measurable space $M$, such that $\mathbb{P} << \mathbb{Q}$. Then, $\phi-$divergence is defined as,
\begin{align}
    D_\phi(\mathbb{P},\mathbb{Q}) = \int_M \phi \Big(\frac{d\mathbb{P}}{d\mathbb{Q}}\Big )d\mathbb{Q}.
\end{align}
Here, $\phi: \mathbb{R}^+ \mapsto \mathbb R$ is a convex function such that $\phi(1)=0$. A popular example of this class is the Kullbeck-Leibler (KL) divergence, where $\phi(t) = t\log(t)$.
Similarly, IPMs can be defined as
\begin{align}
    D_\mathcal{F}(\mathbb{P}, \mathbb{Q}) = \sup_{f \in \mathcal{F}} \Big \vert \int_M fd\mathbb{P} - \int_M fd\mathbb{Q} \Big \vert.
\end{align}
A popular member of this class is the Maximum Mean Discrepancy (MMD), where $\mathcal{F}$ is set to be the Reproducing Kernel Hilbert Space (RKHS).
An intuition behind these two classes of metrics is that the first one tries to measure the ratio (where a ratio of one corresponds to identical sets), while the second one measures the distance between two distributions (where a distance of zero corresponds to identical sets).

Naturally, statistical distances are the optimal choice to measure the similarity of the generated sample distribution to the true data-generating distribution. However, for most high-dimensional problems of interest, computation of such statistical similarity measures is intractable as it requires estimation of the densities induced by the distributions. This has led to the development of estimators of these measures~\cite{goldfeld2021sliced, belghazi2018mine, rhodes2020telescoping, sugiyama2012density, gutmann2010noise}. A popular class of plug-in estimators includes pre-training a classifier~\cite{sugiyama2012density, gutmann2010noise, rhodes2020telescoping} to first estimate the log-ratio of the densities and then taking its Monte-Carlo expectation. Another class of estimators relies on neural density estimation~\cite{alsing2019fast}, where individual densities are estimated using highly non-linear bijective functions~\cite{kingma2018glow, kobyzev2020normalizing} and then used to estimate the discrepancy. Similar projection-based approaches also exist for IPMs~\cite{goldfeld2021sliced, deshpande2019max}. However, the efficacy of such estimators is dependent on the modality and the dimensionality of the problem's data. Therefore domain-agnostic estimators of statistical similarity remain largely an open problem.

Consequently, many of the leading similarity metrics are domain-specific, leveraging certain advantages of the domain to calculate. In computer vision, countless domain-specific metrics have seen widespread adoption~\cite{borji2019pros}. Fr\'echet Inception Distance (FID), for example, uses a pre-trained Inception network to calculate vector embeddings for images, assumes a Gaussian distribution over this embedding space, then calculates Wasserstein-2 Distance (Fr\'echet distance under the Gaussian assumption) between the generated and dataset distributions in this embedding space~\cite{heusel2017gans}\footnote{Other popular metrics calculate different IPMs, such as Maximum Mean Discrepancy in Kernel Inception Distance (KID)~\cite{binkowski2018demystifying}. Other variants like Inception Score (IS) use the statistical distance between marginal and true label distributions~\cite{salimans2016improved}.}. In text generation methods, the commonly used perplexity metric is also a close relative of statistical distance. Perplexity is defined as the exponential of cross-entropy, which itself is the entropy of the data distribution plus the KL divergence between the generated distribution and the data distribution. 

FID and perplexity have been empirically found to correlate well with the human perceptual evaluation of image and text realism. Since human perceptual evaluation is relatively uniform in computer vision and natural language, FID and perplexity are among the most commonly used metrics for model evaluation in their respective fields. In other fields, like design, the perception of realism is much more varied. This non-uniformity, alongside other challenges like data modality, may preclude any generalizable design-specific statistical distance metric from rising to prominence. 

Instead, practitioners should evaluate the viability of applying domain-specific methods to design problems on a case-by-case basis. 
For example, when evaluating similarity in a structural topology generation problem, one may consider using image-based metrics like FID, KID, and IS, which utilize a pre-trained image classifier. These pre-trained image classifiers are often trained on ImageNet, a large computer vision dataset~\cite{deng2009imagenet}.
Accordingly, these metrics are highly biased towards ImageNet~\cite{barratt2018note}. For example, even when evaluating models trained on CIFAR-10, a very similar natural image dataset, researchers have found that inception score significantly misrepresented model performance~\cite{rosca2017variational}. If practitioners use FID to evaluate structural topologies (an evaluation choice with precedent~\cite{regenwetter2022deep}), they will likely incur even greater bias\footnote{It is uncertain whether the low-dimensional latent representations extracted from a network trained on animal or food images in ImageNet would contain any useful information about structural topologies.}. Therefore, practitioners must be cautious about adopting pre-trained metrics from non-design domains and carefully evaluate their suitability.

In general, statistical distance metrics are excellent tools to evaluate similarity and ensure that generated designs are similar to the training data, but are often challenging to estimate in high-dimensional problems. When design data is similar to natural image datasets or natural text corpora, off-the-shelf variants of statistical divergence methods can be effectively used. However, when reliably estimating statistical distance is infeasible or when practitioners desire more nuance in evaluating similarity, they can instead turn to a variety of other methods, which we present in the following subsections. 

\subsection{Decoupling `Realism' and `Coverage'}
Researchers often point to a key shortcoming of statistical distance metrics, namely their inability to decouple two separate ideas in distribution matching: `realism' and `coverage.' Realism is the idea that generated samples should resemble the dataset\footnote{The term `realism' can be misleading since it implies that the dataset reflects reality (i.e., covers the entire space of `realistic' data), which is typically untrue.}. Coverage is the idea that the entire spread of the dataset should be represented by generated samples. While any model that achieves perfect (zero) statistical distance must achieve both perfect realism and coverage, an imperfect model can suffer from an unknown balance of imperfect realism or coverage, which is difficult to diagnose with only a single score. To combat this, methods like precision-recall curves analyze generative models with an entire tradeoff front between these two factors. 

\paragraph{Precision-Recall Curves(Fig.~\ref{fig:PR})}
Precision-recall curves are borrowed concepts from supervised classification but have been adapted as metrics for generative models. The concept was originally proposed by Lucic~\etal~\cite{lucic2018gans} and extended by Sajjadi~\etal~\cite{sajjadi2018assessing}. In Sajjadi~\etal's framework, generated data and original data are pooled, then clustered. A precision-recall (PR) curve is then calculated by comparing the proportion of generated versus original data within each discrete cluster over a sweep of a weighting parameter. Other methods have also been proposed to generate PR curves from arbitrary distributions, bypassing the need for discrete binning of the data~\cite{simon2019revisiting}. From the curve, summary scores like Area-Under-the-Curve (AUC) and a maximum $F_1$ score can also be derived. We refer the reader to~\cite{sajjadi2018assessing} for mathematical reasoning and visual examples behind the PR curves for generative models.

\subsection{Similarity of Generated Data to Dataset (`Realism')} \label{realism}
Although precision-recall curves demonstrate how a DGM tends to balance realism and coverage, in some cases, we may care more for one or the other. When practitioners particularly desire that generated samples resemble the dataset, they can turn to one of several tools to evaluate realism, independent of coverage. 

\paragraph{Precision} \label{precision}
Precision for generative models captures the fraction of generated designs falling within the support of the dataset. Extracting singular precision values from PR curves is challenging since it's unclear which value to select, or how to average values. Instead, Sajjadi~\etal~\cite{sajjadi2018assessing} propose using the maximum $F_\beta$ score for some $\beta<<1$ over all precision-recall pairs in the PR curve as a proxy for precision. 

\paragraph{Nearest Datapoint~(Figure~\ref{fig:NDS})}
For a simple estimate of `realism,' practitioners can calculate the distance to the nearest datapoint for every generated sample. Despite its simplicity, the score is an effective method to capture the local realism of individual samples (or the `realism' performance of a DGM when averaged). This metric is particularly important for applications of DGMs in data augmentation, as it demonstrates that generated synthetic designs (samples) are similar to a dataset of existing designs (datapoints).

\subsection{Dataset Coverage}
In some design applications, there might be a need to focus more on design space coverage to ensure that all key modalities of the design space are reflected in generated designs. In such cases, practitioners can instead estimate recall or use other metrics to quantify coverage. 

\paragraph{Recall} \label{recall}
Like precision, recall is a metric discussed in Sajjadi~\etal~\cite{sajjadi2018assessing}, measuring the proportion of datapoints in the support of the generated distribution. Practitioners can select some maximum $F_\beta$ score for some $\beta>>1$ to estimate recall. 

\paragraph{Nearest Generated Sample~(Figure~\ref{fig:NGS})}
For a simple approach to capture dataset coverage, practitioners can calculate the distance to the nearest generated sample for every original datapoint. This score can be averaged over the dataset to evaluate a set of generated samples. However, this score depends on the size of the generated sample set, necessitating the selection of a set size tuning parameter for standardization.

\paragraph{Rediscovery~(Figure~\ref{fig:Red})}
Rediscovery is an evaluation technique that evaluates an algorithm's ability to rediscover datapoints that were withheld during training. Rediscovery is commonly used in the molecule synthesis domain~\cite{bilodeau2022generative}, though we believe it is widely applicable (and valuable) across design disciplines. In a discrete setting, the rediscovery rate can be calculated as the exact proportion of withheld designs rediscovered in a generated set~\cite{shah2022survey}. To accommodate other data modalities, practitioners can relax the score to instead calculate the distance from a withheld datapoint to the nearest sample in a generated sample set. Effectively, this performs a nearest generated sample evaluation over the set of withheld designs~\footnote{We also note that other metrics can also be evaluated on a ``test set,'' though we feel that evaluating coverage on a holdout set is a particularly insightful choice.}. Though it requires both a tuning parameter (holdout size) and the foresight to remove a split of the dataset before training, rediscovery is an elegant extension of pure `coverage' metrics that further quantifies the simple generalization capabilities of the model. 

\subsection{Evaluating Effective use of Generated Data in Downstream Tasks}
A common approach to measuring the similarity of a generated sample set to the training dataset is to check whether the generated set serves as an effective stand-in for some auxiliary task. If a practitioner is generating design data with a specific downstream task in mind, it may be viable to directly evaluate generated samples on this task as a metric. In other scenarios, artificially constructed tasks can serve as an effective method to evaluate generated samples. 
\paragraph{Machine Learning Efficacy}
Auxiliary machine learning tasks, such as classification, are often used to evaluate generated sample sets in a process known as Machine Learning (ML) efficacy testing. Often, a supervised machine learning model is trained on the generated dataset and then tested on the original data, though several formulations of ML efficacy have been proposed, as in~\cite{ravuri2019classification, xu2019modeling}. 

\subsection{Demonstration of Statistical Similarity Metrics}\label{simdemo}

\begin{figure}[!htb]
     \centering

     % \begin{subfigure}[b]{0.26\textwidth}
     %     \centering
     %     \includegraphics[width=\textwidth]{Images/SDORig.png}
     %     \caption{Dataset}
     %     \label{fig:SDORig}
     % \end{subfigure}

     \begin{subfigure}[b]{0.4\textwidth}
         \centering
         \includegraphics[width=\textwidth]{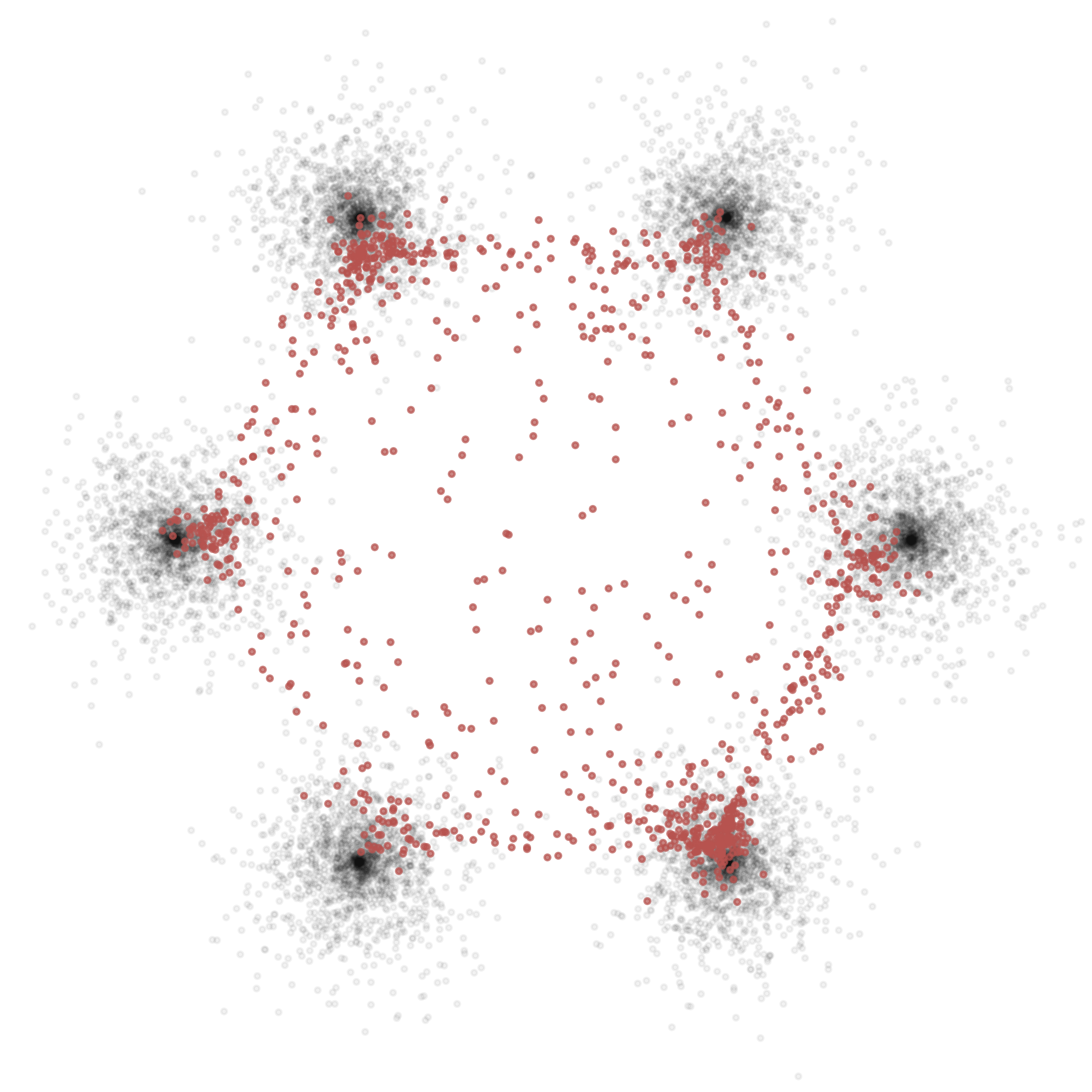}
         \caption{VAE}
         \label{fig:SDVAE}
     \end{subfigure}
     \begin{subfigure}[b]{0.4\textwidth}
         \centering
         \includegraphics[width=\textwidth]{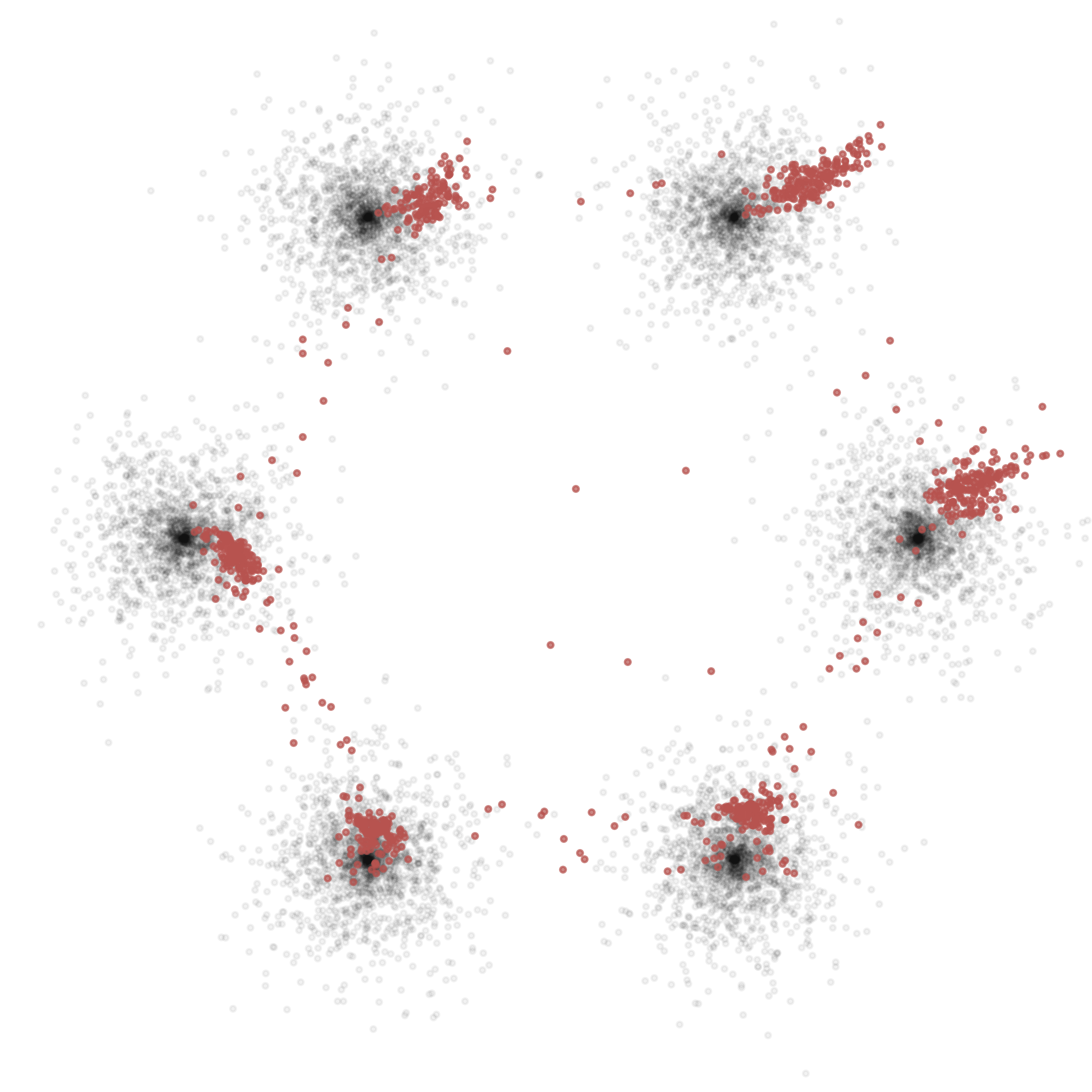}
         \caption{GAN}
         \label{fig:SDGAN}
     \end{subfigure}

    \caption{Distributions generated by a Variational Autoencoder and Generative Adversarial Network (red) overlaid over training data (gray). The GAN dominates in accuracy-related metrics while the VAE outperforms in coverage. }
    \label{fig:SD}
\end{figure}

\begin{table}[!htb]
\centering
\caption{Distribution-matching scores for models in Figure~\ref{fig:SD}. The GAN dominates in accuracy-related metrics while the VAE outperforms in coverage. Point metrics are averaged over the generated set. Bold is better.}
\begin{tabular}{ccc}
\toprule
Metrics & VAE & GAN \\
\midrule
Nearest Datapoint & 0.043 & \textbf{0.018} \\
Nearest Generated Sample & \textbf{0.090} & 0.100 \\
Rediscovery & \textbf{0.093} & 0.099 \\
Precision-Recall Curve F1 & \textbf{0.475} & 0.386 \\
Precision-Recall Curve F10 & 0.907 & \textbf{0.934} \\
Precision-Recall Curve F0.1 & \textbf{0.804} & 0.753 \\
Precision-Recall Curve AUC & \textbf{0.468} & 0.379 \\
Maximum Mean Discrepancy & 0.045 & \textbf{0.013} \\
Machine Learning Efficacy & 0.675 & \textbf{0.765}\\
\bottomrule
\end{tabular}%
\label{tab:SD}
\end{table}

Thus far, we have introduced a variety of metrics to quantify a model's ability to generate distributions of designs that match the training dataset. To showcase the use of these metrics, we evaluate two classic generative models, a Variational Autoencoder (VAE) and a Generative Adversarial Network (GAN)\footnote{The goal of this example is to compare any two DGMs; hence, the state-of-art models and strong instantiations were intentionally not chosen as they performed very well on a two-dimensional case.} using select metrics on a synthetic dataset which challenges models to learn six non-overlapping data modes (Figure~\ref{fig:SD}, Table~\ref{tab:SD}). Details on model architecture, metrics settings, and training are included in the appendix. Individual distribution-matching metrics, such as MMD, indicate that the GAN is the stronger performer among the two models. However, when looking at an array of metrics, one can note that the GAN is the stronger performer in accuracy-based metrics such as F10 and nearest datapoint. In contrast, the VAE outperforms the GAN in many coverage-based metrics such as nearest generated sample, rediscovery, and F0.1. In overall distribution-matching metrics, results are mixed, with the GAN outperforming in MMD and falling behind in F1 and AUC. The GAN also performed significantly better in machine learning efficacy on this dataset. 
Compared to a single metric, this suite presents a more nuanced picture, and the final model selection could vary based on the end goal. If coverage is more important, then the VAE may be preferred, while the GAN may be preferred for realism\footnote{We note that these results need not generalize to other datasets or architectures.}. Since even the best coverage of the design space does not guarantee the novelty or diversity of generated designs, we next introduce design exploration metrics. 

\section{Evaluating Design Exploration}~\label{sec:Novelty}
Design exploration can be an important consideration in many fields, such as product design, architecture, and engineering, where new and innovative solutions are often sought after. A model that successfully explores the design space will typically generate diverse sets of novel designs. Diversity often goes hand-in-hand with the generalizability of a model and diversity-aware DGMs have even been shown to avoid common generalizability pitfalls such as mode collapse~\cite{chen2021mopadgan, chen2021padgan}. Design novelty refers to the degree to which a design is new or unique compared to existing designs. Achieving novelty is particularly difficult for data-driven generative models since it is somewhat contrary to its default objective of learning and mimicking a dataset. In design methodology literature, novelty is typically considered a point metric, whereas diversity (or variety) is typically considered a set metric. In this section, we discuss several metrics with which to quantify the novelty of generated samples and the overall diversity of generated sample sets. 

\begin{figure}[!htb]
     \centering
     \begin{subfigure}[b]{0.32\textwidth}
         \centering
         \includegraphics[width=\textwidth]{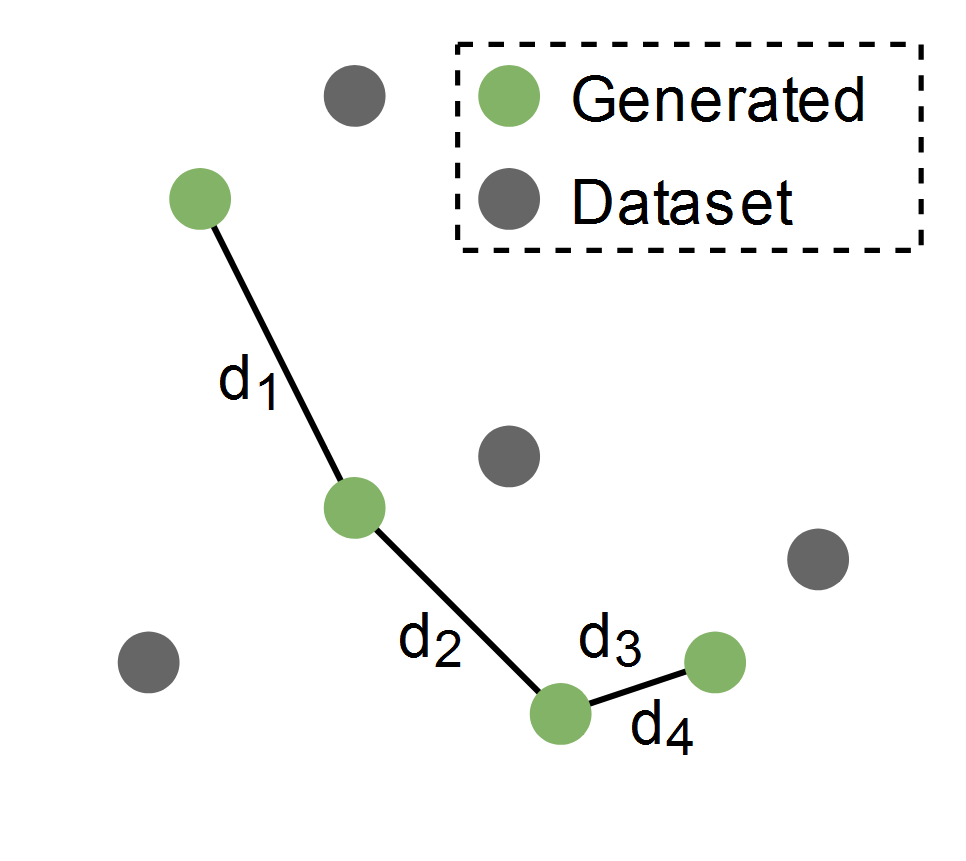}
         \caption{Inter-Sample Distance}
         \label{fig:ISD}
     \end{subfigure}
     \begin{subfigure}[b]{0.32\textwidth}
         \centering
         \includegraphics[width=\textwidth]{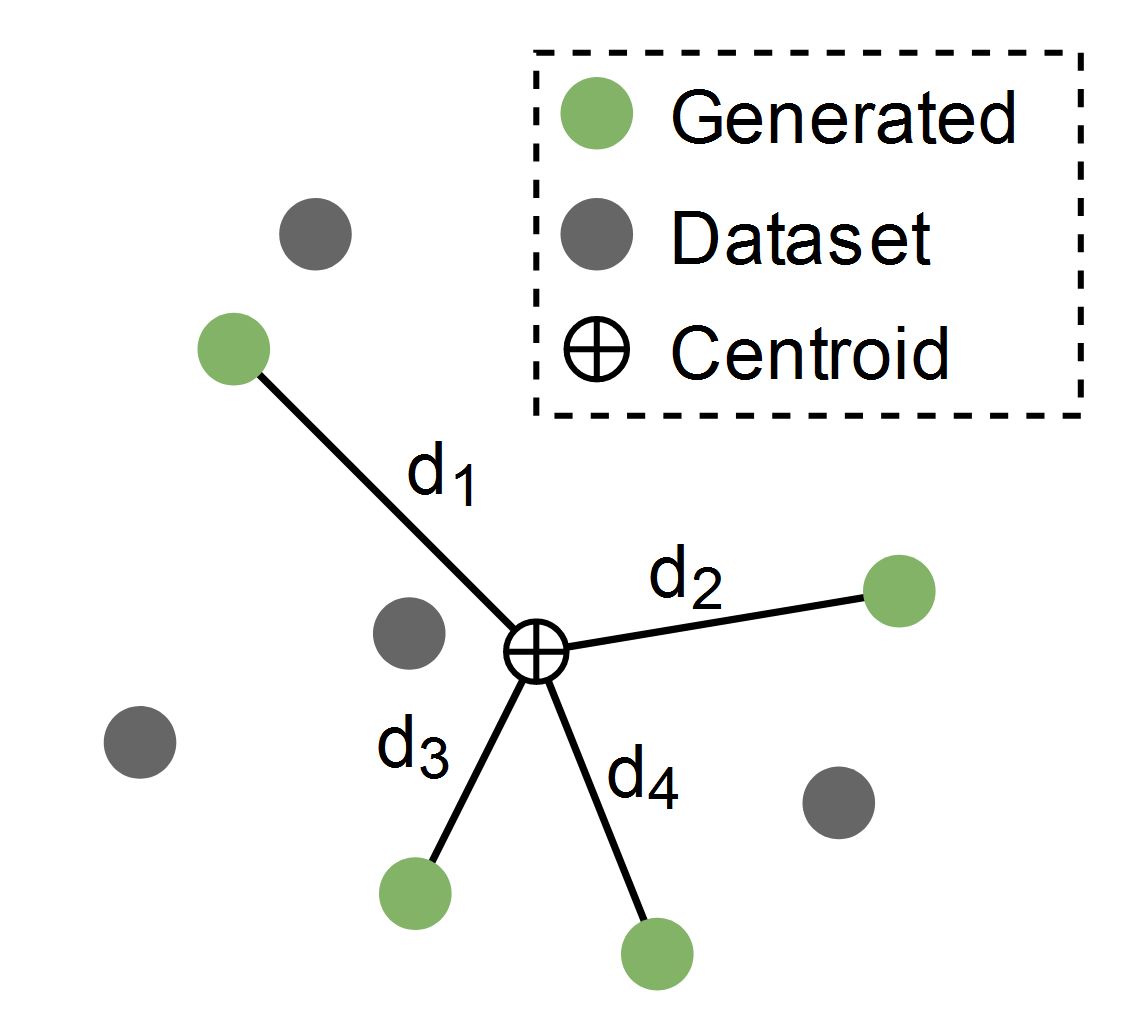}
         \caption{Distance to Centroid}
         \label{fig:DTC}
     \end{subfigure}
     \begin{subfigure}[b]{0.32\textwidth}
         \centering
         \includegraphics[width=\textwidth]{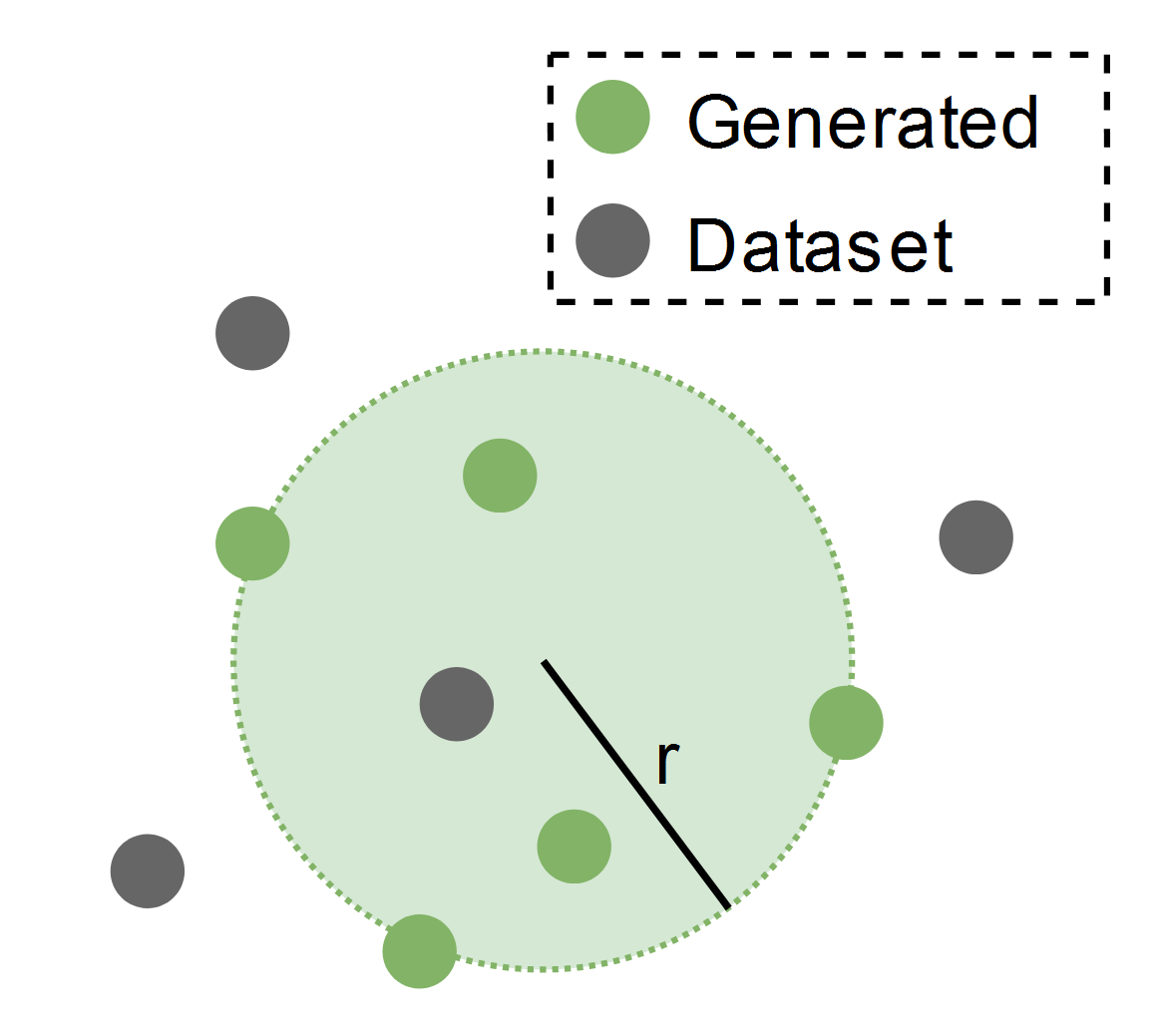}
         \caption{Sm. Enclosing Hypersphere}
         \label{fig:SEH}
     \end{subfigure}
     
     \begin{subfigure}[b]{0.32\textwidth}
         \centering
         \includegraphics[width=\textwidth]{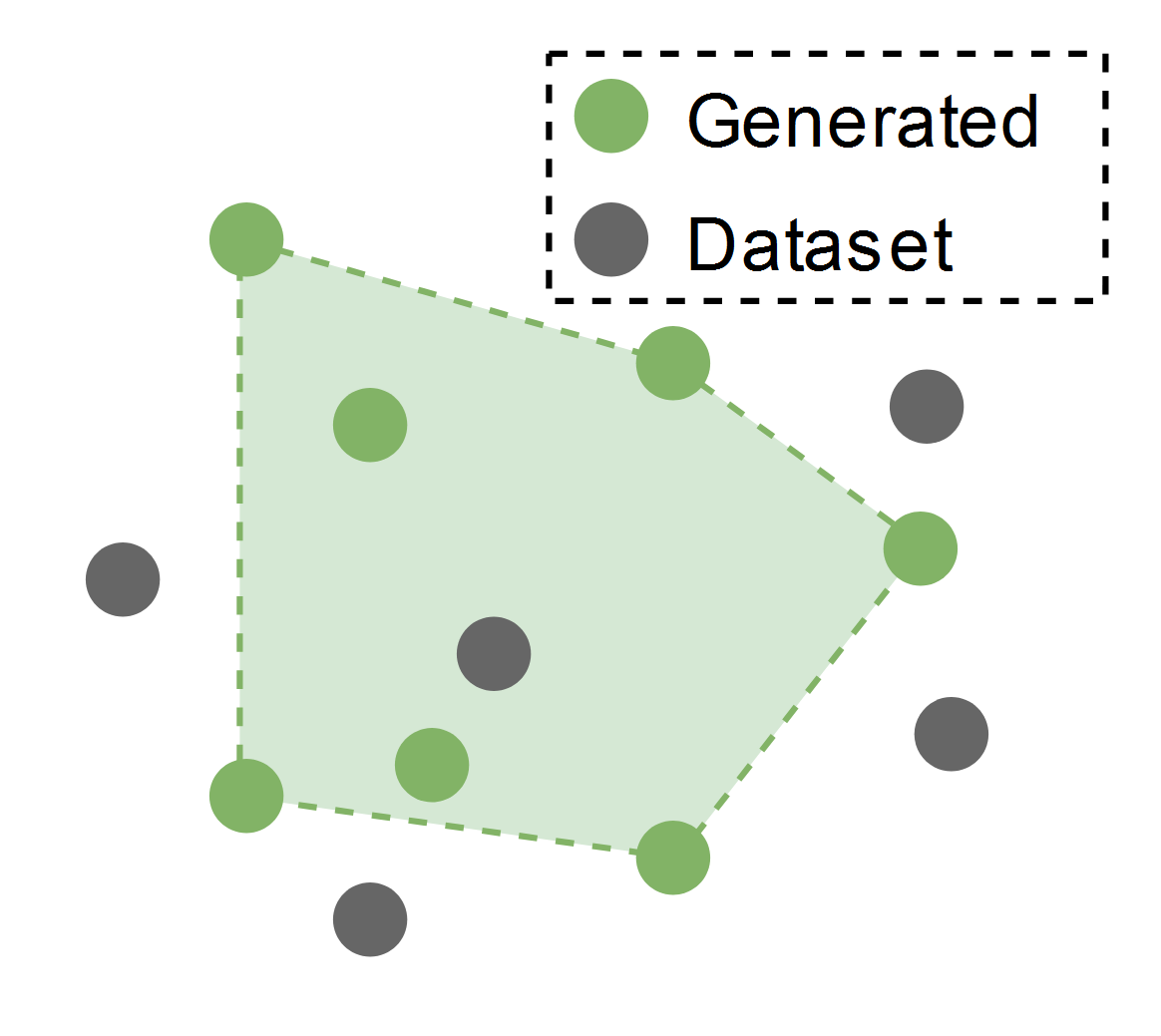}
         \caption{Convex Hull}
         \label{fig:CH}
     \end{subfigure}
     \begin{subfigure}[b]{0.64\textwidth}
         \centering
         \includegraphics[width=\textwidth]{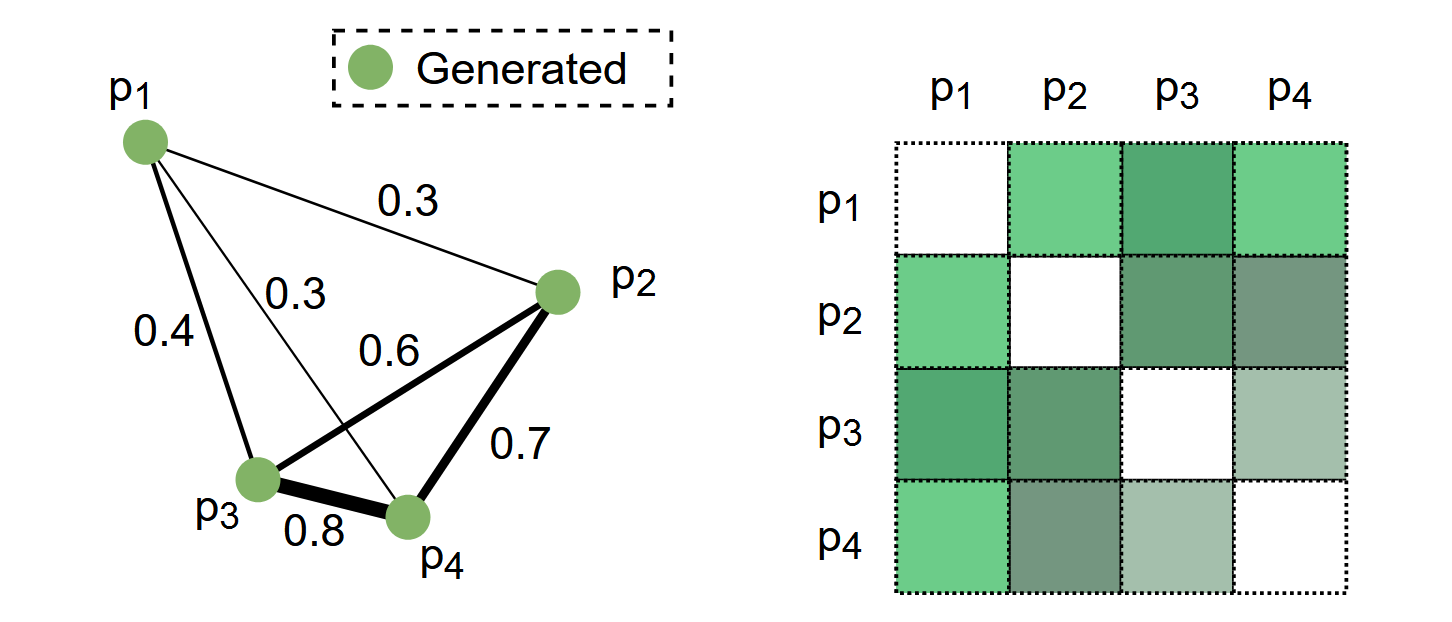}
         \caption{DPP Diversity Score}
         \label{fig:DPP}
     \end{subfigure}

    \caption{Illustrations of select diversity and novelty metrics.}
    \label{fig:novelty}
\end{figure}
\subsection{Novelty}
Novelty refers to the aspect of something being new, original, or unique. 
The design methodology community has a rich body of work classifying and defining novelty. A common distinction is `psychological novelty' (P-novelty) versus `historical novelty' (H-novelty). A design is considered P-novel if the idea is new for the person who generated it. In contrast, a design is H-novel if an idea has never appeared in history before~\cite{boden2009computer}. If we treat our generative model as the `designer' and consider the dataset as the set of designs that the designer has seen, we argue that the P-novelty can be estimated for a DGM. However, since we can't assume that our dataset is comprised of every design ever conceptualized, estimating H-novelty is challenging, even if we use the dataset as a reference distribution. Next, we provide a few methods to estimate novelty.

\paragraph{Nearest Datapoint~(Figure~\ref{fig:NDS})}
The nearest datapoint metric (previously introduced as a similarity metric in Section~\ref{realism}) could be used as a P-novelty metric, measuring how far from the nearest datapoint a generated sample is. Whereas the metric was best minimized as a metric for accuracy, when used to evaluate novelty, a larger score is preferable. Nearest datapoint is a very insightful metric when checking for data copying, where the DGM will overfit and memorize individual datapoints to replicate while sampling. One limitation of the metric is its large sensitivity to individual datapoints. Nevertheless, the nearest datapoint metric has been used in design literature~\cite{chen2021padgan} to demonstrate the capability of deep generative models to create novel designs. 

\paragraph{Inter-Sample Distance~(Figure~\ref{fig:ISD})}
Another simple metric to estimate the P-novelty of a generated sample is the distance to the nearest other generated sample. In this simple form, the metric is a strong assessment of how unique a sample is in its local neighborhood. However, the metric has also been relaxed to use the distance to the $n^{th}$ nearest neighbor~\cite{lehman2011abandoning}.  

\paragraph{Distance to Centroid~(Figure~\ref{fig:DTC})}
Instead of using the distance to the nearest samples, practitioners can instead quantify the novelty of a generated sample using the distance from the sample to a single point which summarizes a set of generated samples, such as the centroid or the geometric median. As Brown \& Mueller~\cite{brown2019quantifying} note, the choice of centroid or median can yield very different results. Though simple and cheap to compute, the metric may carry implicit assumptions about convexity. For example, in a torus-like distribution, the centroid may be novel. Mueller \& Ochsendorf~\cite{mueller2015combining} propose a variant to adapt the metric into a diversity score using the maximum distance to the centroid or median. We discuss diversity in more detail in the next subsection.

\subsection{Diversity}
Design diversity is closely related to the concept of ``entropy'' in information theory and refers to the variety or range of different solutions or designs that are generated for a given problem. In the context of design problems, it can refer to the degree to which a set of solutions to a problem encompasses different styles, forms, or variations. 
While novelty is a point metric, diversity measures a property of a set of designs, though many averaged novelty metrics may often closely correlate with diversity.

Diversity can be broken down into two components: uniformity and spread. Uniformity measures the relative distance between designs. Spread, also known as `extent' in multi-objective optimization literature, measures the range of designs within the generated distribution~\cite{Riquelme2015performance}. Imagine the design space as a balloon with small balls inside it. The spread can be thought of as the diameter of the balloon. However, even with a fixed diameter, the balloon may have different uniformity diversity scores, for example, if all the balls are evenly distributed versus if most of the balls are stuck in one corner. Many diversity metrics, such as entropy, combine uniformity and spread into a single value. Below, we discuss a few of these diversity metrics\footnote{Averaged novelty metrics are also often more focused on either spread or uniformity. For example, averaged inter-sample distance and averaged nearest datapoint are often strong measures of uniformity, whereas the averaged distance to centroid is more focused on spread.}.

\paragraph{Entropy}
Entropy scores can be used as a metric for design diversity when evaluated on a set of generated designs. 
Some popular metrics include the Shannon entropy index, Herfindahl–Hirschman Index (HHI)~\cite{ahmed2019measuring}, Gini-Simpson index, and inverse Simpson index~\cite{brown2019quantifying}. When practitioners have discrete data, they can directly calculate the entropy. When working with continuous representations, however, practitioners must estimate entropy from samples, a well-studied statistical problem~\cite{kozachenko1987sample}. Entropy captures both uniformity and spread, and its properties are well-grounded in mathematics and information theory. 

\paragraph{Smallest Enclosing Hypersphere~(Figure~\ref{fig:SEH})}
The smallest enclosing hypersphere metric is a purely spread-focused metric that identifies the hypervolume of the smallest hypersphere that encloses all generated samples. Originally proposed for novelty measurement~\cite{pavoine2005measuring}, the calculation is nontrivial for high-dimensional data and is often approximated to reduce cost~\cite{brown2019quantifying}. Smallest enclosing hypersphere is highly sensitive to relative scaling between parameters. It also makes a convexity assumption and is sensitive to outliers. 

\paragraph{Convex Hull~(Figure~\ref{fig:CH})}
The convex hull is another spread-focused metric defined as the smallest convex set that includes a set of generated samples. The total hypervolume enclosed within the generated set's convex hull is a common metric for diversity~\cite{podani2009convex}. Brown \& Mueller~\cite{brown2019quantifying} found the convex hull to agree with human assessments of novelty in small low-dimensional design problems, compared to competing metrics. Like the smallest enclosing hypersphere, the convex hull makes a convexity assumption and is often sensitive to outliers. 

\paragraph{DPP Diversity Score~(Figure~\ref{fig:DPP})}
Determinantal Point Processes (DPP) can be used in conjunction with distance metrics to evaluate a diversity score of a generated sample set. DPPs calculate a score based on the eigenvalues of a matrix constructed using distances between points from a generated sample set~\cite{kulesza2012determinantal}. 
The determinant operation in linear algebra computes the volume of a parallelepiped formed by vectors. 
When evaluating the diversity of a set of items using a DPP, one essentially looks at the volume in the feature space that these items span. 
Like entropy, DPP diversity captures both uniformity and spread. The benefit of DPP is its ease of calculation for high-dimensional data, as it only requires a positive-semidefinite kernel as an input. However, it is sensitive to design duplicates, as the determinant collapses to zero when any of the eigenvalues is zero. We caution that DPP diversity is highly nonlinear and small relative changes in score may imply sizeable changes in diversity.

\subsection{Demonstration of Design Exploration Metrics}\label{novdemo}
In this section, we introduced a variety of diversity and novelty-related metrics. To demonstrate their performance, we use the same representative synthetic data problem introduced in Figure~\ref{fig:SD}. Scores are shown in Table~\ref{tab:DN}. The VAE dominates in average novelty metrics (nearest datapoint and inter-sample distance), owing largely to the fact that the generated samples are more spaced apart and span regions of the space outside of the dataset. However, the GAN outperforms the VAE in diversity metrics focused on spread (convex hull and distance to centroid) because samples generated by the GAN span a larger convex space. This illustrates a shortcoming with certain diversity metrics where the potentially diverse samples in the center of the space generated by the VAE do not contribute to the score. The DPP diversity score, which takes into account both uniformity and spread, favors the VAE when using a Euclidean radial basis function (RBF) kernel. 

From this two-dimensional example, we would like to highlight two points. First, average novelty metrics are easily confused with diversity metrics, but as this example illustrates, they do not always agree with diversity scores. Second, practitioners need to be aware of convexity assumptions and consider both uniformity and spread when evaluating a model's diversity. 

\begin{table}[!htb]
\centering
\caption{Diversity and novelty scores for generated distributions from Figure~\ref{fig:SD}. The VAE dominates in average novelty, while the GAN dominates in spread-focused metrics. Point metrics are averaged over the generated set. Bold is better.}
\begin{tabular}{ccc}
\toprule
Metrics & VAE & GAN \\
\midrule
Nearest Datapoint & \textbf{0.043} & 0.018 \\
Inter-Sample Distance & \textbf{0.030} & 0.019 \\
Convex Hull & 5.651 & \textbf{7.173} \\
DPP Diversity & \textbf{14.398} & 14.688 \\
Distance to Centroid & 1.120 & \textbf{1.391} \\
\bottomrule
\end{tabular}%

\label{tab:DN}
\end{table}

\section{Evaluating Design Constraints} \label{Constraints}
Many design problems have constraints, which are limitations placed on the possible designs within a given problem or task. These constraints can be physical, such as geometric constraints, or functional, such as performance or cost requirements. They are used to guide the design process and ensure that the final solution meets certain requirements and is feasible to implement. Constraints are commonly driven by materials or manufacturing limitations, industry standards, or nonnegotiable safety requirements. 
Unlike performance targets, which we discuss in Section~\ref{targets}, we \textit{must} satisfy constraints for the design to be valid. Constraint satisfaction metrics can roughly be sorted in a strict order of preference depending on what constraint information is provided (closed-form, black-box, dataset, etc.). In this section, we propose several techniques for quantifying constraint satisfaction in DGMs depending on the constraint information available.

\begin{figure}[!htb]
    \centering
    \begin{subfigure}[b]{0.32\textwidth}
     \centering
     \includegraphics[width=\textwidth]{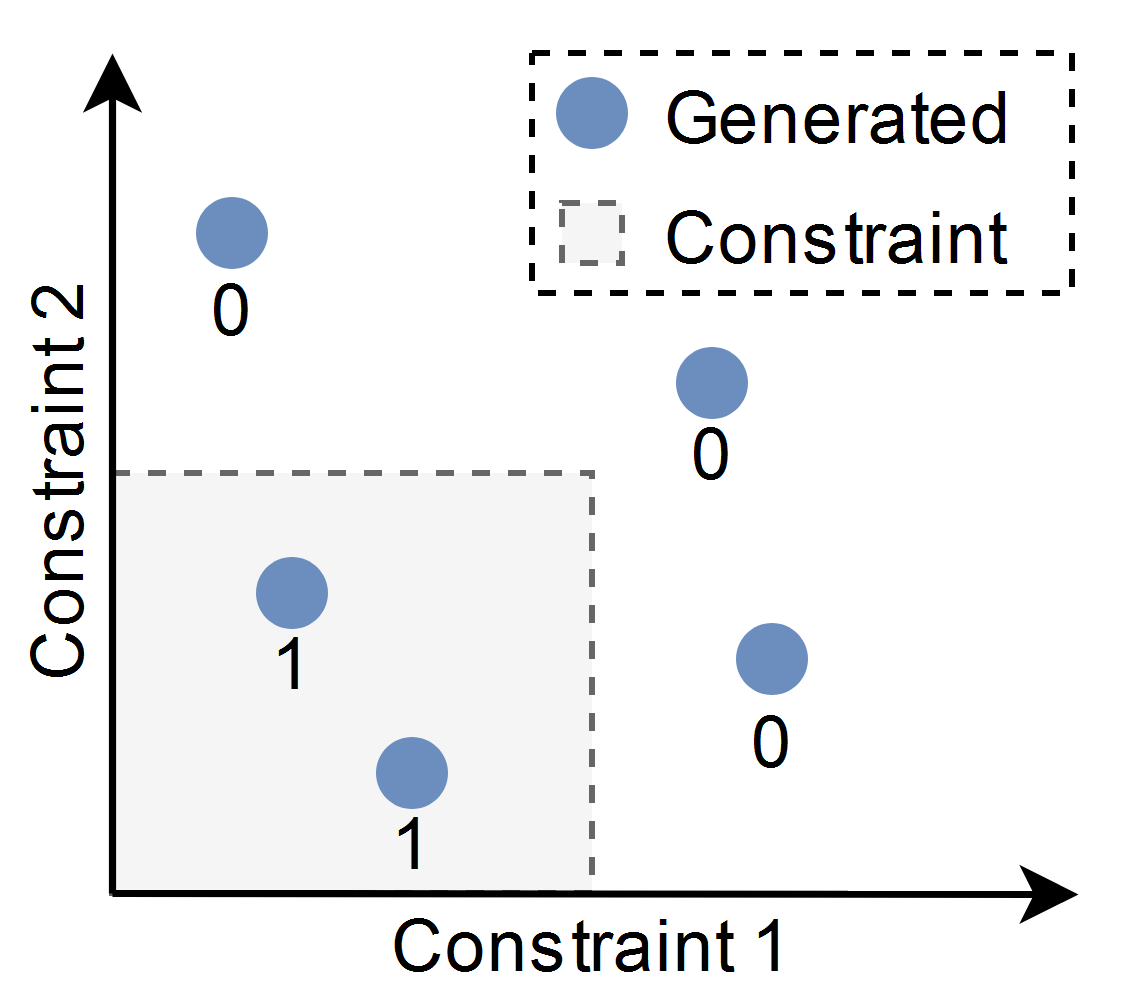}
     \caption{Constraint Satisfaction}
     \label{fig:CS}
    \end{subfigure}
    \begin{subfigure}[b]{0.32\textwidth}
     \centering
     \includegraphics[width=\textwidth]{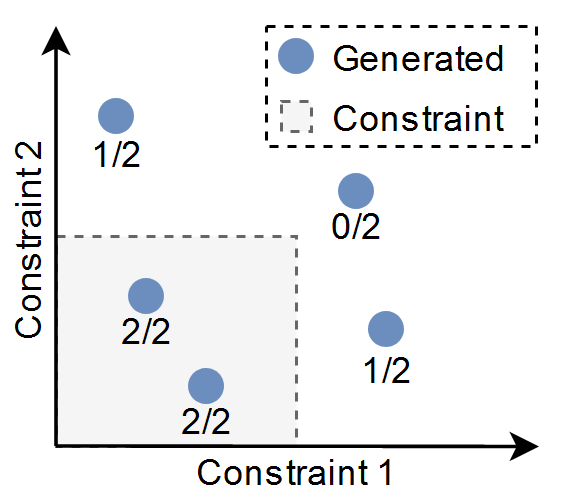}
     \caption{Constraint Satisfaction Rate}
     \label{fig:CSR}
    \end{subfigure}
    \begin{subfigure}[b]{0.32\textwidth}
     \centering
     \includegraphics[width=\textwidth]{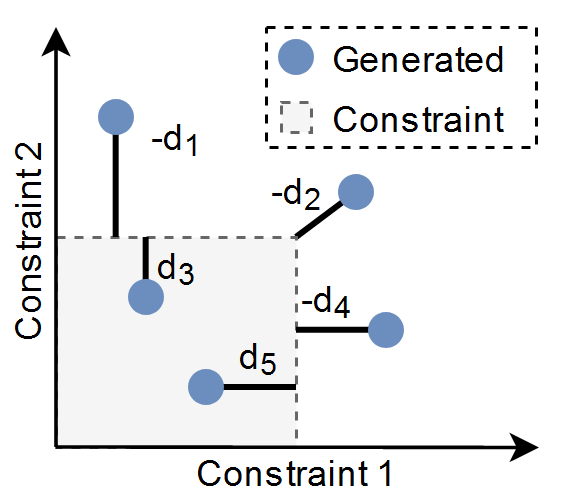}
     \caption{Signed Dist. to Bound.}
     \label{fig:SDC}
    \end{subfigure}
    \begin{subfigure}[b]{0.32\textwidth}
     \centering
     \includegraphics[width=\textwidth]{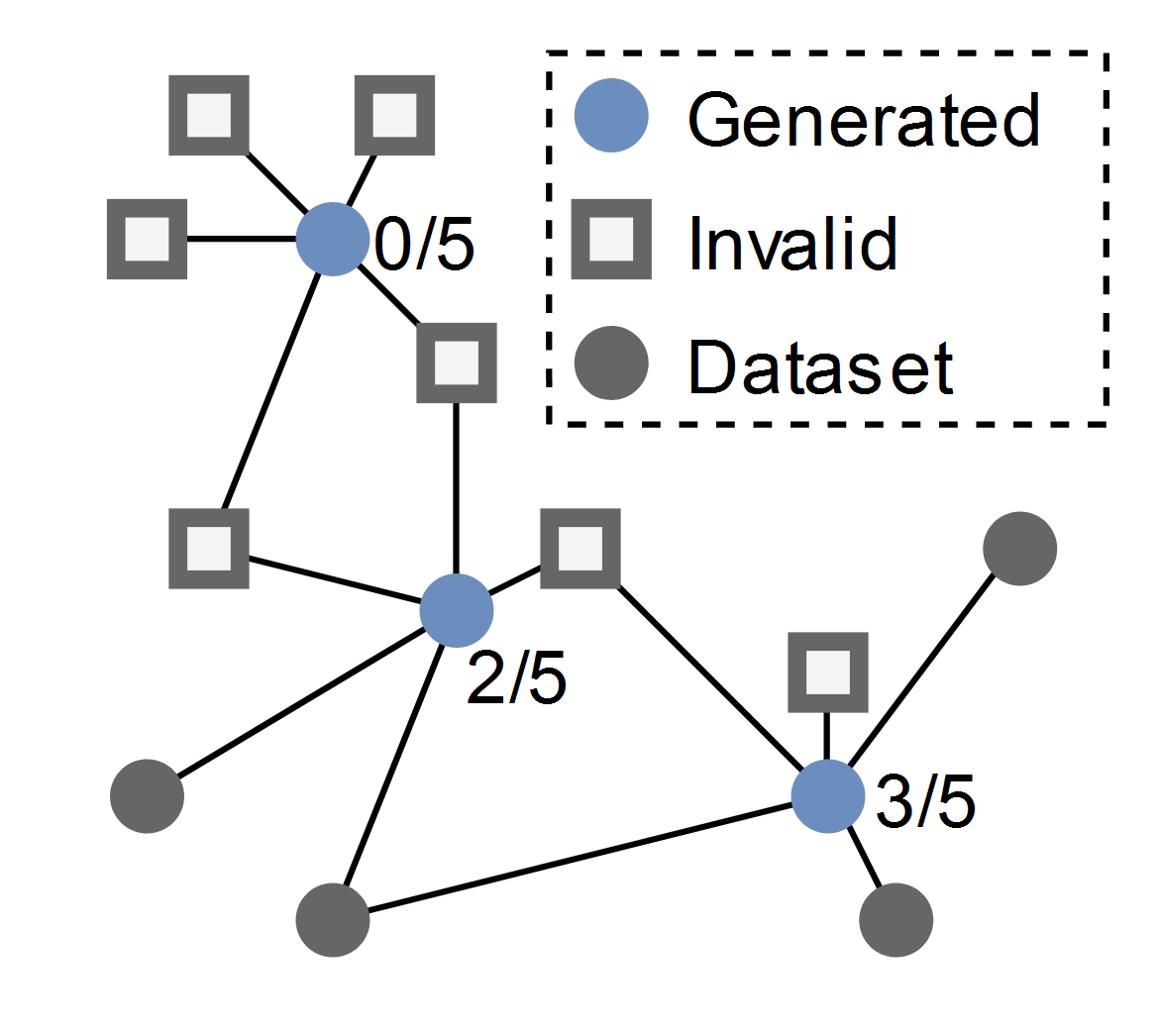}
     \caption{Neighbor Validity Fraction}
     \label{fig:NVF}
    \end{subfigure}
    \begin{subfigure}[b]{0.32\textwidth}
     \centering
     \includegraphics[width=\textwidth]{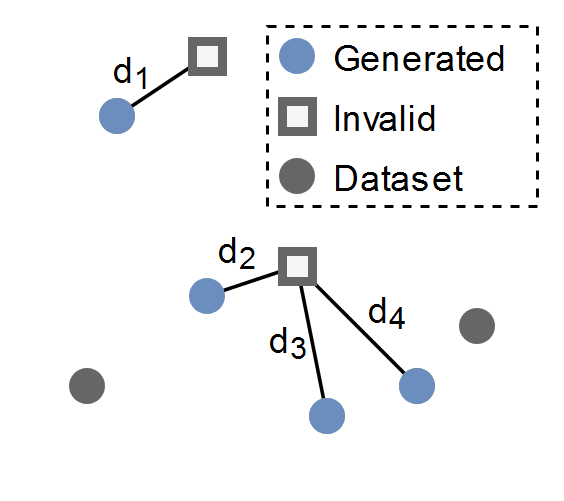}
     \caption{Nearest Invalid Datapoint}
     \label{fig:NIS}
    \end{subfigure}

    \caption{Illustrations of select constraint satisfaction metrics.}
    \label{fig:Constraints}
\end{figure}

\subsection{Leveraging Constraint Violation Tests}
In many design problems, there exist known methods (using analytical equations, rules, simulations, or physics knowledge) to test whether a design is valid with respect to each of the problem constraints. Since closed-form constraint definitions are rare in design problems, practitioners may instead have to turn to such methods to evaluate constraint satisfaction. Below, we list a few methods that leverage such constraint violation tests, when available.

\paragraph{Constraint Satisfaction~(Figure~\ref{fig:CS})}~\label{sec:scs} 
Provided that practitioners have access to some black-box constraint satisfaction test, a simple binary constraint satisfaction value (i.e., does a generated sample simultaneously meet all constraints?) suffices as a simple metric. When averaged over a generated sample set, this can effectively serve as an indicator for the proportion of generated samples that are valid. In practice, more versatile scores are often more informative, especially in problems with multiple constraints.

\paragraph{Constraint Satisfaction Rate~(Figure~\ref{fig:CSR})}
A variant of the simple constraint satisfaction score is the constraint satisfaction rate, which quantifies the proportion of all constraints met by a single generated sample. If different constraints have different priority weightings, this score can be weighted by the priority of the various constraints. 

\subsection{Leveraging Mathematically-Defined Constraint Boundaries}
Though rare, practitioners may sometimes have access to a closed-form mathematically-defined constraint boundary. They can usually then calculate distances from generated design samples to the constraint boundary, which can be highly informative. 

\paragraph{Signed Distance to Constraint Boundary:~(Figure~\ref{fig:SDC})}
The distance to constraint boundary metric is particularly informative as a signed distance field (SDF), with generated samples satisfying the constraints having positive distances and samples violating the constraints having negative distances. Indicating by what margin samples are satisfying or violating constraints can be significantly more informative than a simple binary criterion or proportion of constraints met. 

\subsection{Metrics that Leverage Datasets of Invalid Designs}
Unfortunately, it is common not to have even a black-box constraint evaluator or methods allowing the direct estimation of the distance to the constraint boundary. In such cases, practitioners may have access to or may be able to procedurally generate a large collection of infeasible or invalid designs. These datasets of constraint-violating (invalid) designs provide a tool to approximate the constraint adherence of generated designs. 

\paragraph{Predicted Constraint Satisfaction}
% \subsection{Leveraging Datasets of Constraint-Violating Designs}
When practitioners have access to a reference set of constraint-violating datapoints, they can use a classifier to predict the constraint satisfaction of their generated samples. This classifier can be something complex like a neural network, or something simple and robust like k-nearest neighbors. In the case of k-nearest neighbors, the fraction of neighbors that are valid provides a likelihood that a generated sample satisfies constraints, which serves as the score, as shown in Figure~\ref{fig:NVF}. 

\paragraph{Nearest Invalid Datapoint~(Figure~\ref{fig:NIS})}
When practitioners have access to a reference set of constraint-violating datapoints, they can calculate the distance from each generated sample to the nearest known invalid datapoint. The underlying assumption is that samples near constraint-violating datapoints are also likely to be constraint-violating. This gives a rough approximation of the distance to the constraint boundary. 

\subsection{Demonstration of Design Constraint Metrics}\label{constdemo}
\begin{figure}[!htb]
     \centering
     \begin{subfigure}[b]{0.24\textwidth}
         \centering
         \includegraphics[width=\textwidth]{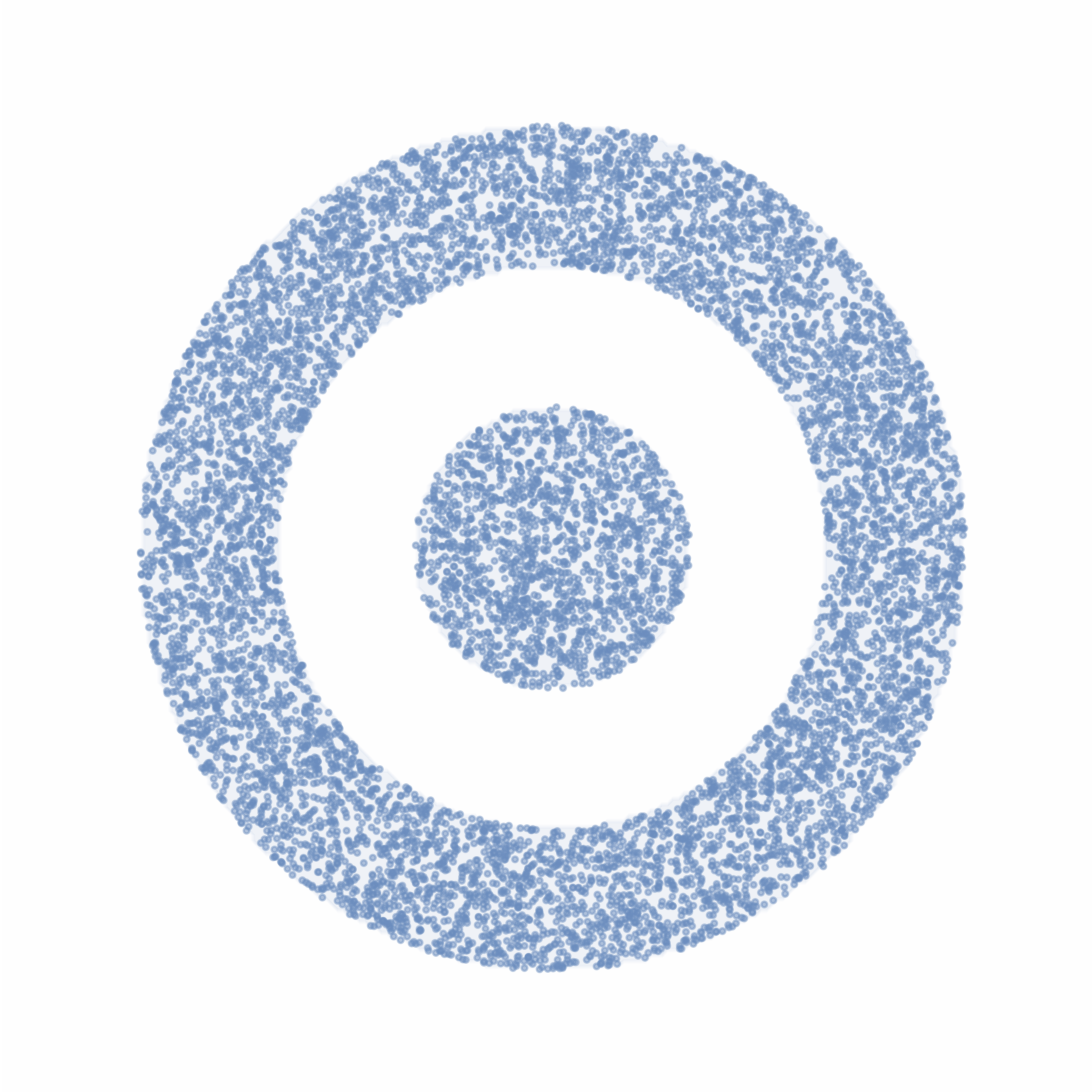}
         \caption{Valid Datapoints}
         \label{fig:CDATA}
     \end{subfigure}
     \begin{subfigure}[b]{0.24\textwidth}
         \centering
         \includegraphics[width=\textwidth]{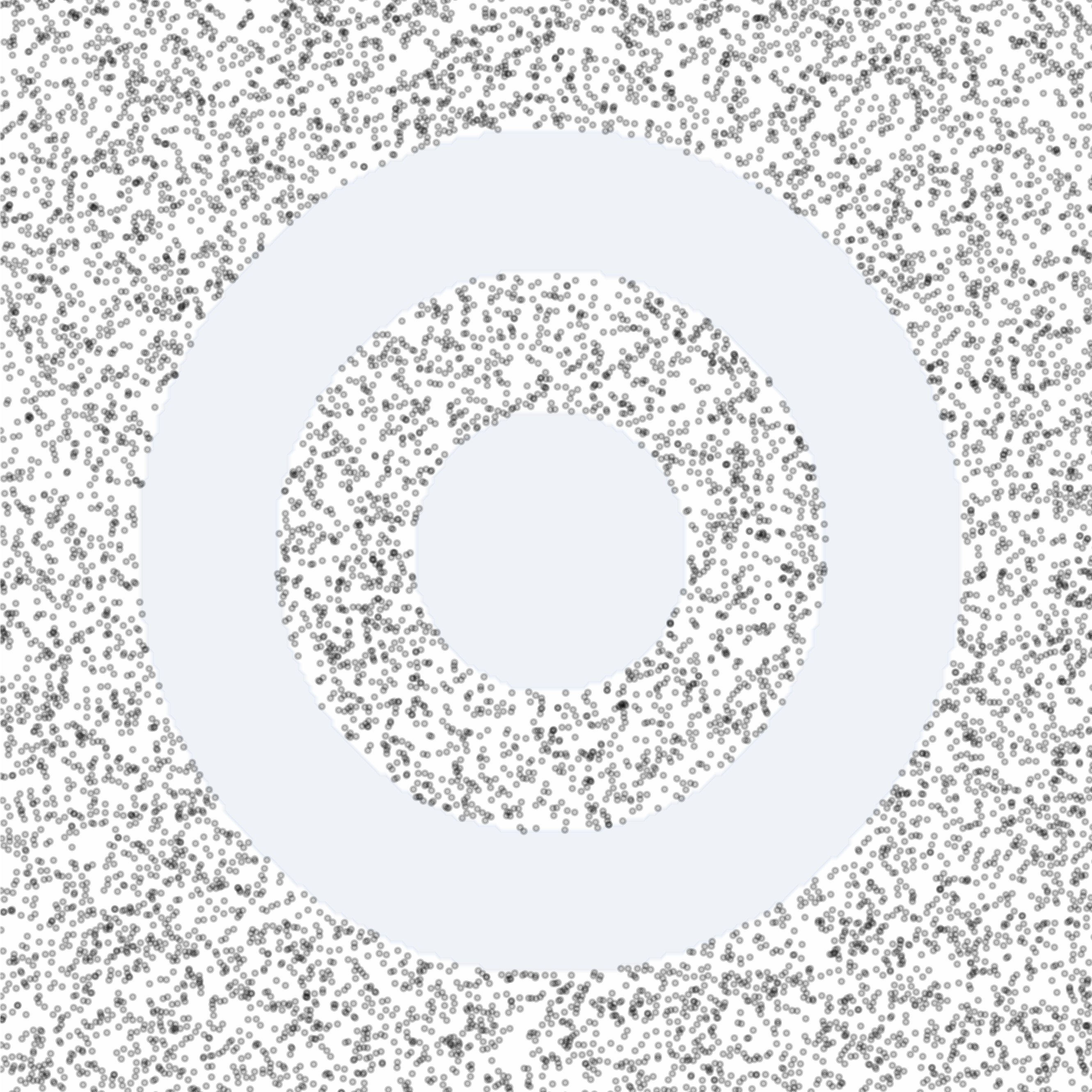}
         \caption{Invalid Datapoints}
         \label{fig:CINV}
     \end{subfigure}
     \begin{subfigure}[b]{0.24\textwidth}
         \centering
         \includegraphics[width=\textwidth]{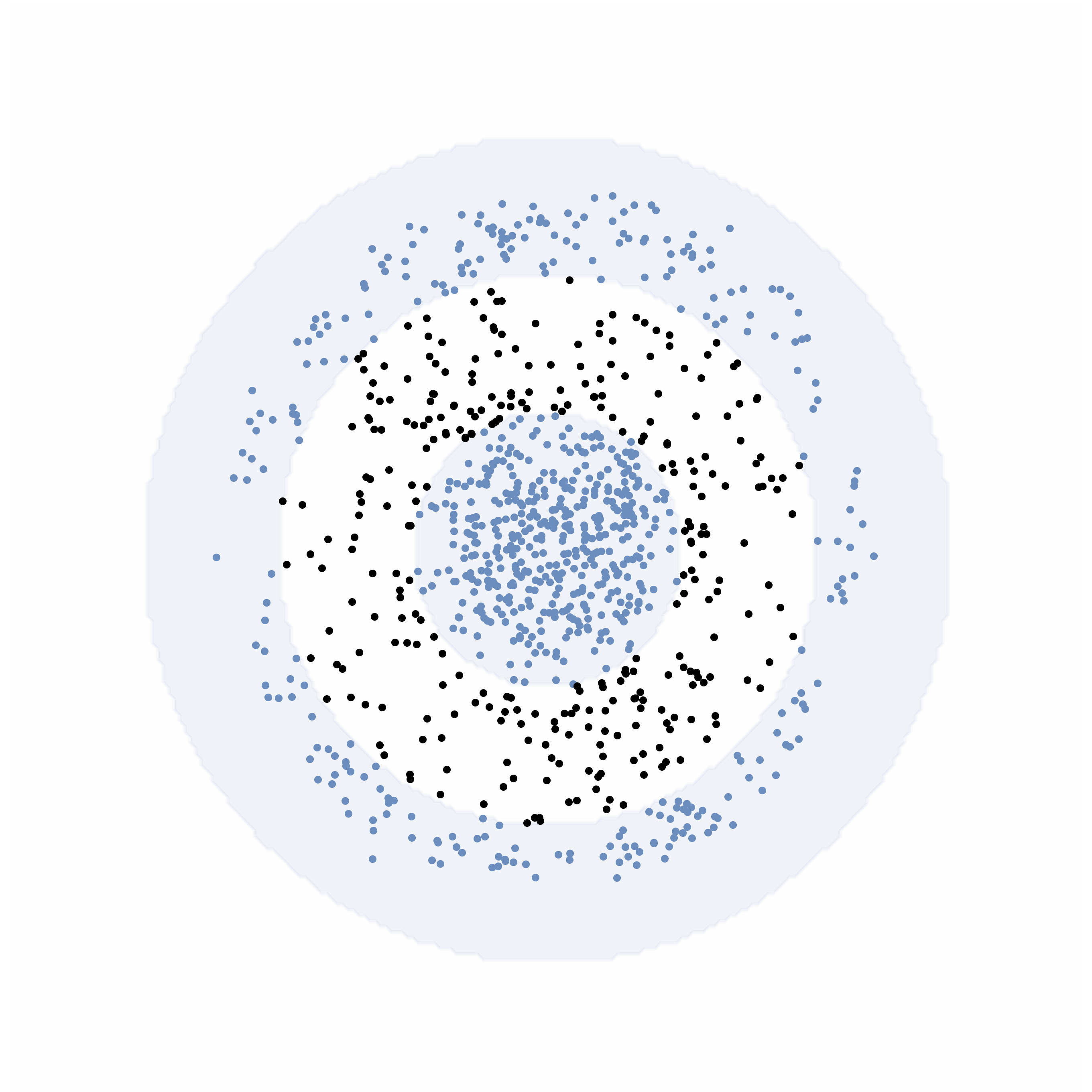}
         \caption{VAE}
         \label{fig:CVAE}
     \end{subfigure}
     \begin{subfigure}[b]{0.24\textwidth}
         \centering
         \includegraphics[width=\textwidth]{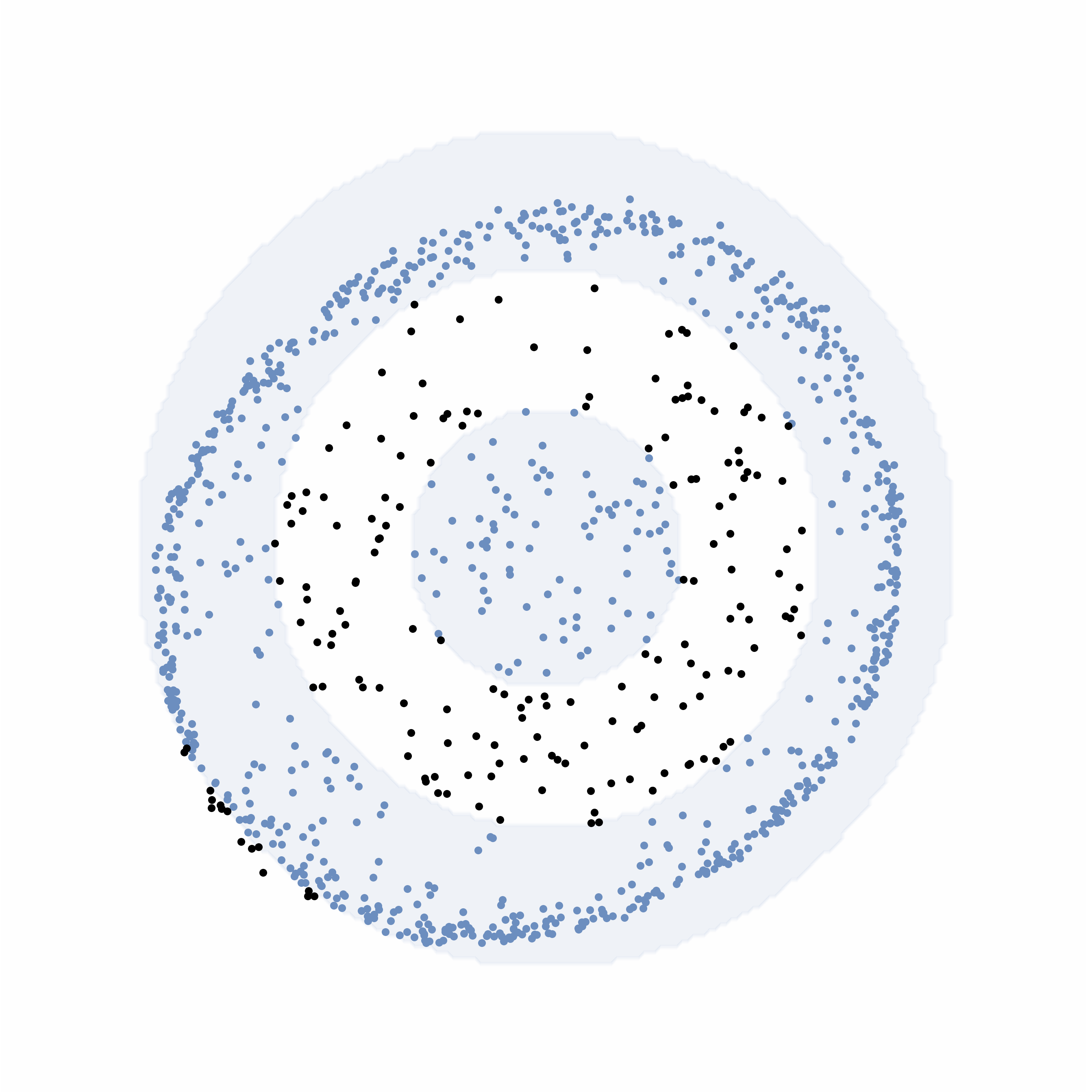}
         \caption{GAN}
         \label{fig:CGAN}
     \end{subfigure}

    \caption{Distributions generated by a Variational Autoencoder and Generative Adversarial Network. Both the VAE and GAN struggle to observe the non-convex constraint dividing the two valid regions of the design space. Generated samples that violate constraints are shown in black, while valid generated samples are shown in blue. The GAN demonstrates generally higher constraint satisfaction performance.}
    \label{fig:C}
\end{figure}

Having presented a variety of metrics for evaluating constraint satisfaction, we again showcase these metrics on a simple two-dimensional problem. To demonstrate the constraint adherence task, we test a GAN and a VAE on a concentric ring problem that we created for this task (Figure~\ref{fig:CDATA}). To support the nearest invalid datapoint metric, we also include a dataset of invalid datapoints, shown in Figure~\ref{fig:CINV}, though these invalid datapoints are not used during training\footnote{Since closed-form constraints are given, there would be little reason to use dataset-based metrics like predicted constraint satisfaction in practice.}. Both the GAN and the VAE struggle to avoid the invalid area of the design space, as seen by points overlapping with the infeasible regions in Figures~\ref{fig:CVAE} and~\ref{fig:CGAN}. However, as shown in Table~\ref{tab:C}, the GAN outperforms the VAE in every metric, indicating that it is better suited to generate feasible designs in this problem. With this, we conclude our discussion about constraint adherence metrics and move on to consider metrics that evaluate design performance.

\begin{table}[!htb]
\centering
\caption{Constraint adherence scores averaged over the generated distributions in Figure~\ref{fig:C}. The GAN demonstrates generally higher constraint satisfaction performance. Point metrics are averaged over the generated set. Bold is better.}
\begin{tabular}{ccc}
\toprule
Metrics & VAE & GAN \\
\midrule
% Reference: MMD & XXXX & \textbf{XXXX} \\
% \midrule
Constraint Satisfaction & 0.713 & \textbf{0.833} \\
Constraint Satisfaction Rate & 0.857 & \textbf{0.917} \\
Predicted Constraint Satisfaction & 0.698 & \textbf{0.823} \\
Nearest Invalid Datapoint & 0.17 & \textbf{0.173}\\
\bottomrule
\end{tabular}%
\label{tab:C}
\end{table}

\section{Evaluating Design Quality or Performance}
In image synthesis problems like human face generation, `quality' and `realism' are almost synonymous -- the more realistic the generated images, the higher their `quality.' In design, quality is typically not associated with similarity to existing designs. Instead, the quality of designs is often governed by an associated set of functional performance characteristics, often modeled as a mapping from the design space to some performance space. In Fig.~\ref{fig:twobikes}, we gave an anecdote of two bike frames with drastically different functional performances, despite being visually and parametrically similar.
Characteristics of quality in design may include factors such as cost, weight, efficiency, etc., and are highly problem dependent. 

The metrics presented in this section typically require a method to evaluate functional performance for generated designs. Sometimes, a trained predictive model can suffice for this task, but increases the possibility of inaccurate and biased scores. We would like to note that a generated design's innate performance attributes as calculated by an evaluator or predictor can also serve as simple but effective metrics for the design and the generator, by extension. 

\begin{figure}[!htb]
     \centering
     \hfill
     \begin{subfigure}[b]{0.32\textwidth}
         \centering
         \includegraphics[width=\textwidth]{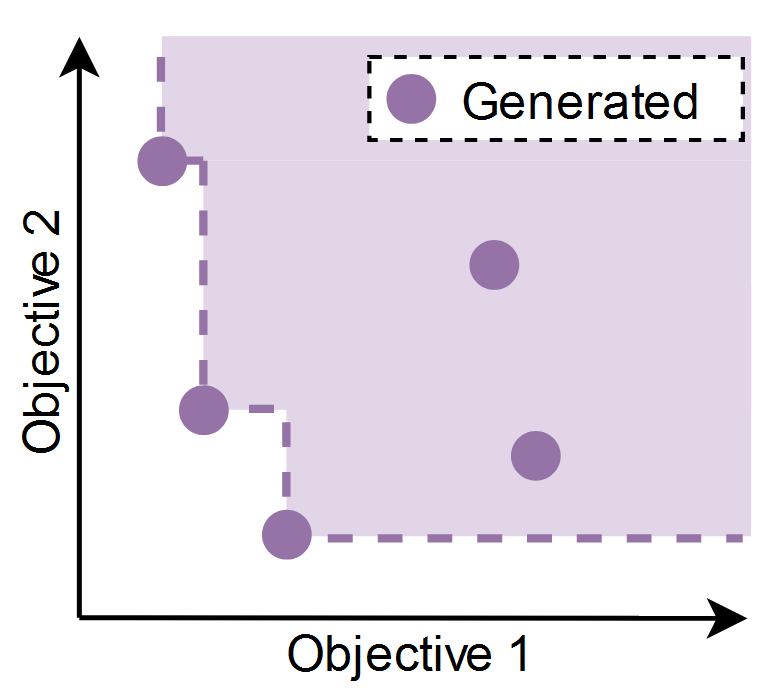}
         \caption{Hypervolume}
         \label{fig:HV}
     \end{subfigure}
     \hfill
     \begin{subfigure}[b]{0.32\textwidth}
         \centering
         \includegraphics[width=\textwidth]{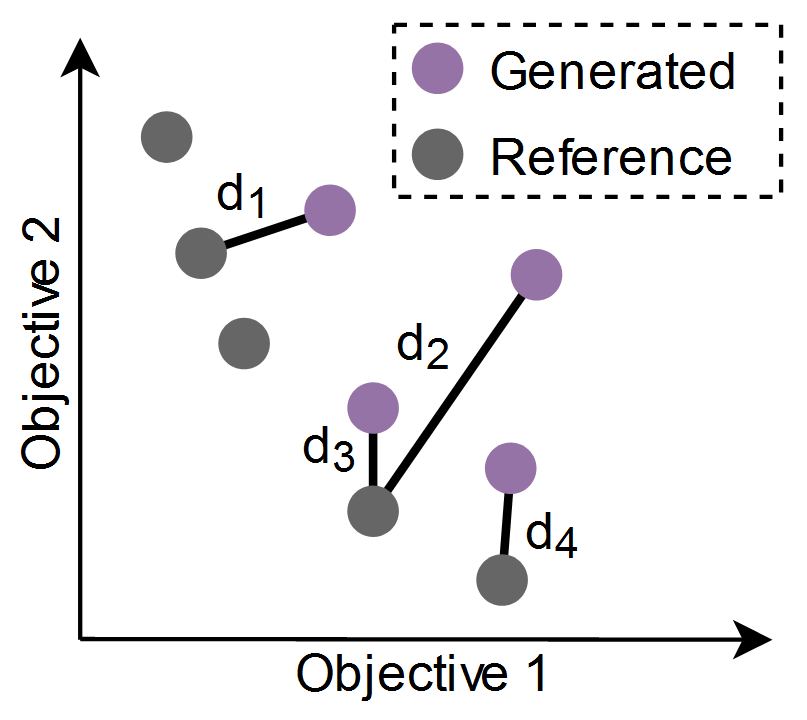}
         \caption{Generational Distance}
         \label{fig:GD}
     \end{subfigure}
     \hfill
     \begin{subfigure}[b]{0.32\textwidth}
         \centering
         \includegraphics[width=\textwidth]{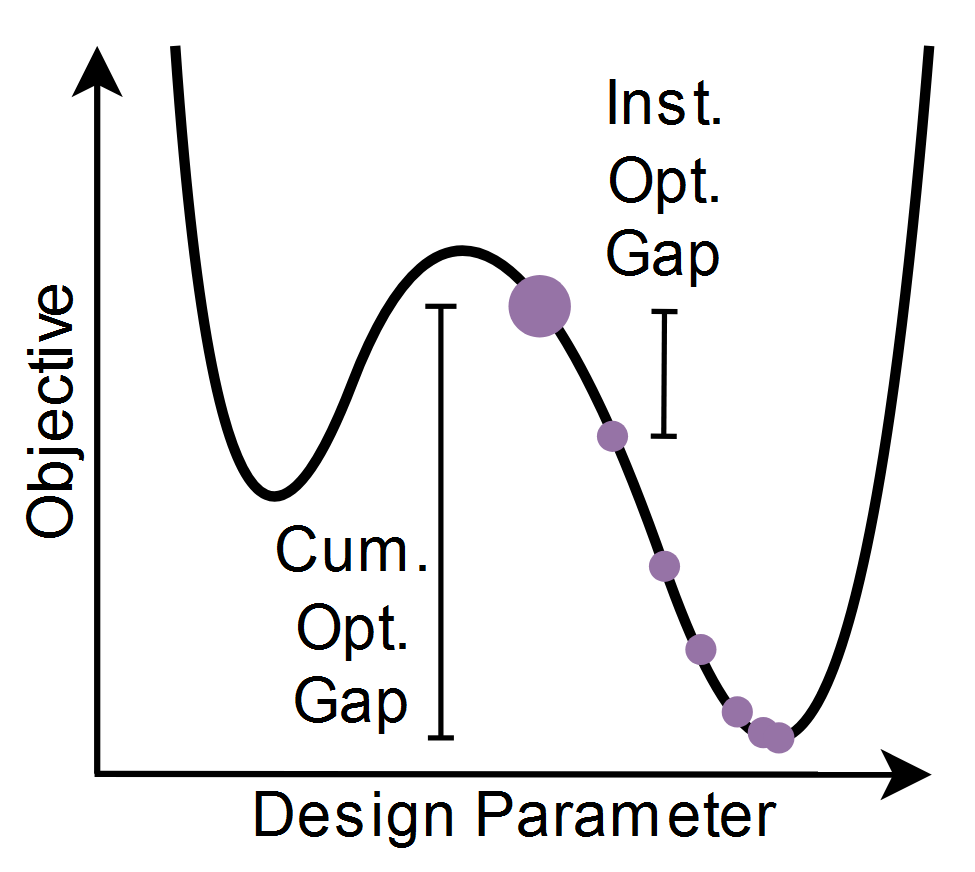}
         \caption{Optimality Gap}
         \label{fig:OG}
     \end{subfigure}
     \hfill

     \hfill
     \begin{subfigure}[b]{0.32\textwidth}
         \centering
         \includegraphics[width=\textwidth]{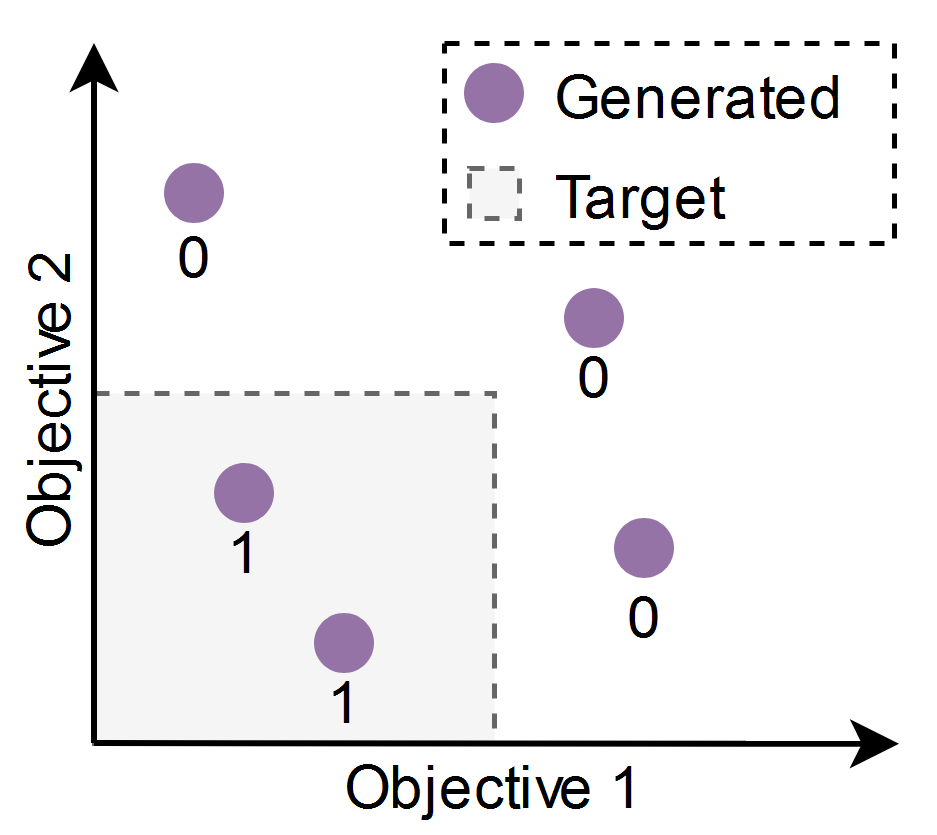}
         \caption{Target Achievement}
         \label{fig:TA}
     \end{subfigure}
     \hfill
     \begin{subfigure}[b]{0.32\textwidth}
         \centering
         \includegraphics[width=\textwidth]{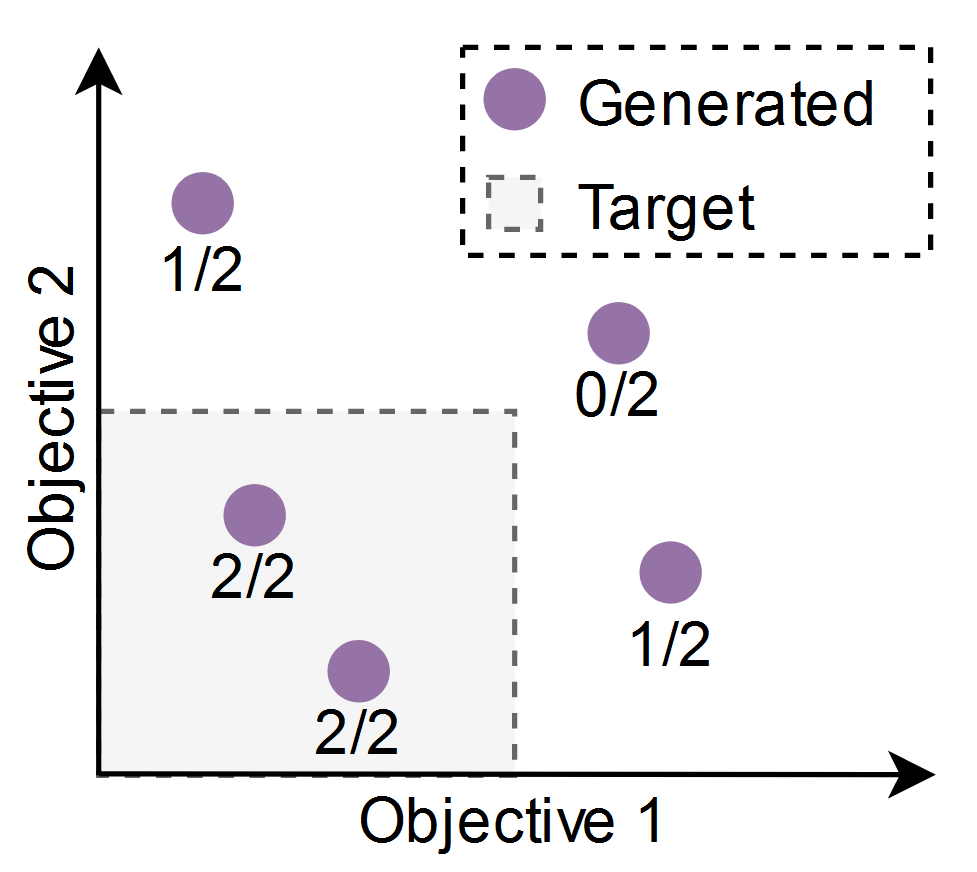}
         \caption{Target Achievement Rate}
         \label{fig:TAR}
     \end{subfigure}
     \hfill
     \begin{subfigure}[b]{0.32\textwidth}
         \centering
         \includegraphics[width=\textwidth]{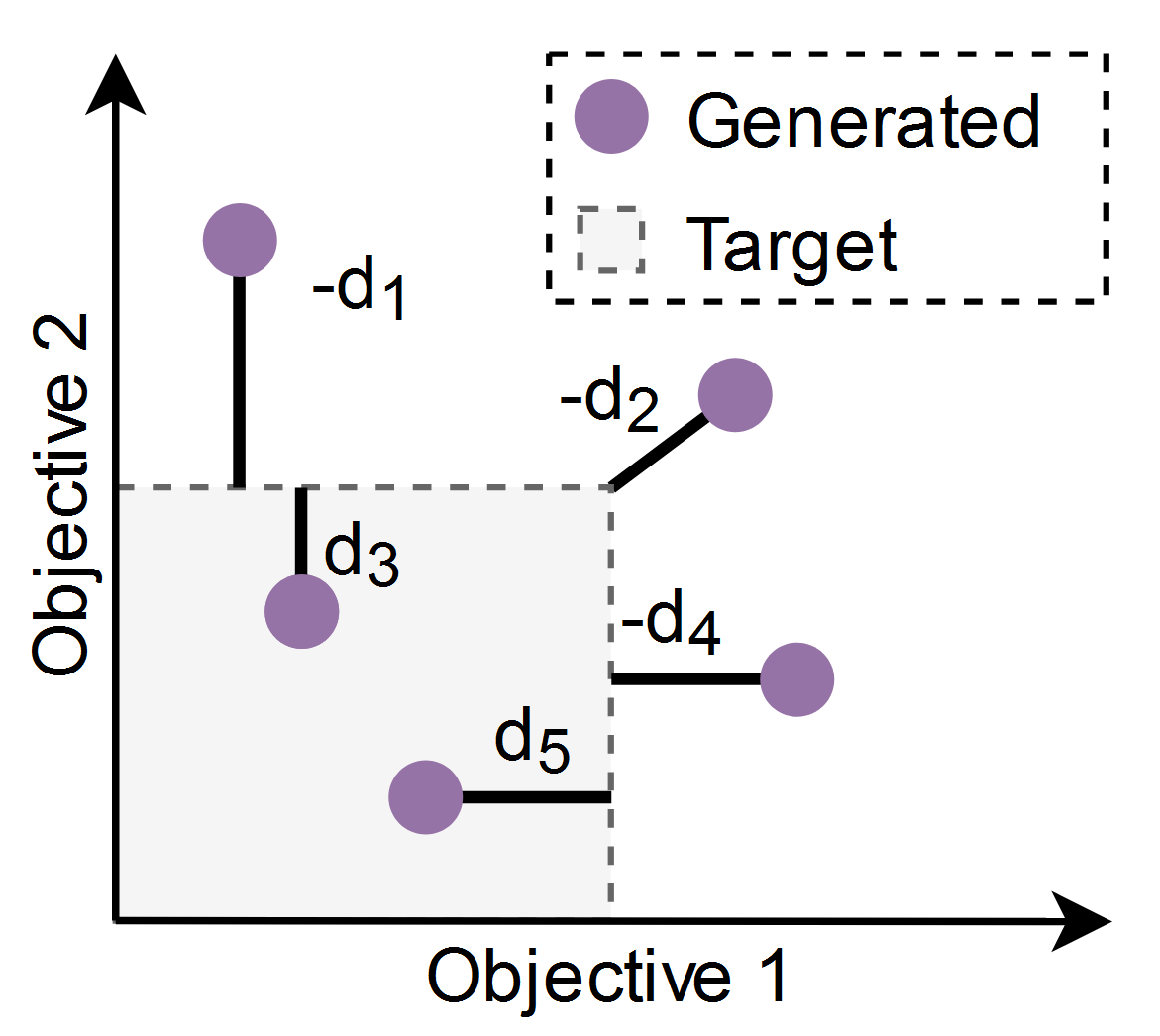}
         \caption{Signed Distance to Target}
         \label{fig:SDT}
     \end{subfigure}
     \hfill
     
     \begin{subfigure}[b]{0.64\textwidth}
         \centering
         \includegraphics[width=\textwidth]{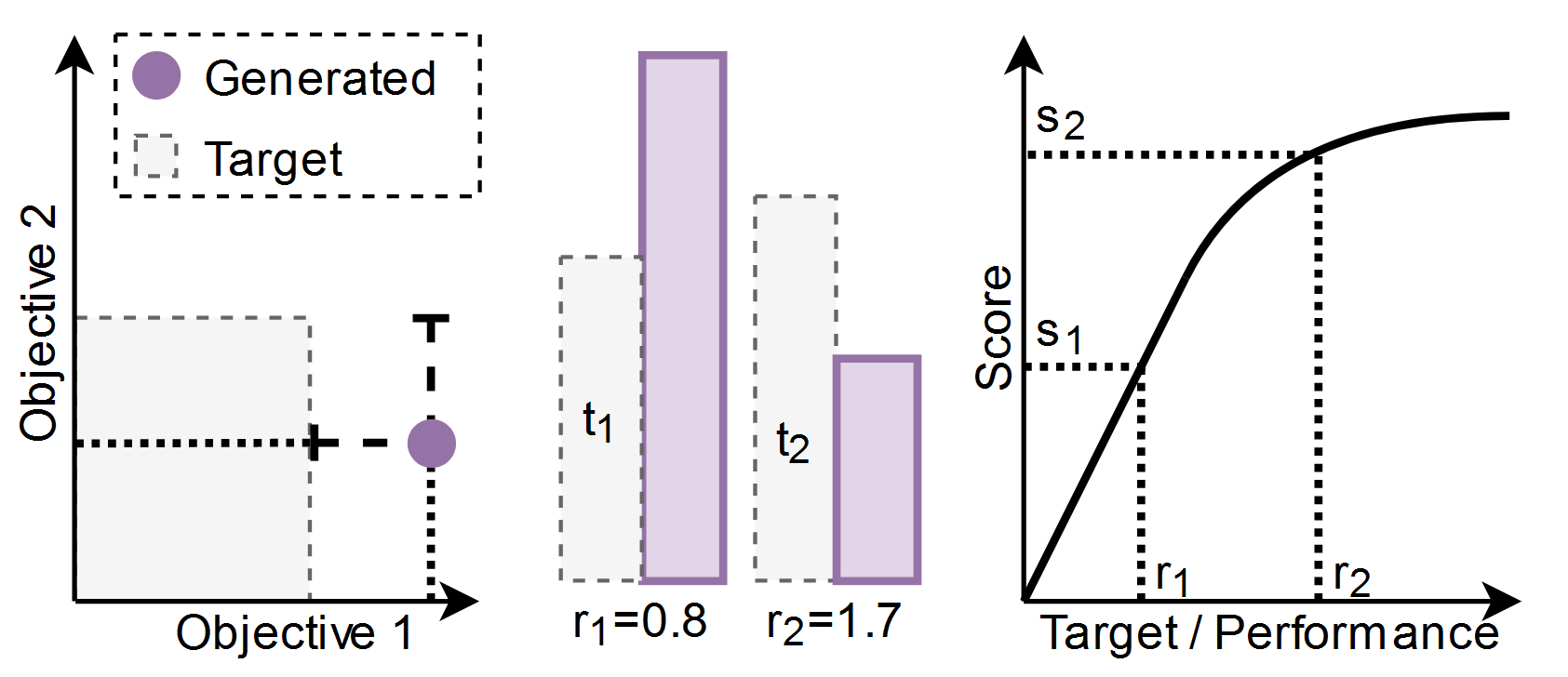}
         \caption{Design Target Achievement Index}
         \label{fig:DTAI}
     \end{subfigure}
     
    \caption{Illustrations of select design quality and target achievement metrics.}
    \label{fig:performance}
\end{figure}

\subsection{Performance Optimality}
In many engineering problems, designers want to find a distributed set of solutions that are performant across multiple objectives. This is especially common in multi-objective optimization problems, where designers often seek to identify a Pareto-optimal set of points. Pareto-optimal designs have the property that performance improvement in any objective must come at the expense of performance in some other objective. This means that Pareto-optimal designs cannot be `dominated' by any other design, i.e., there exists no design that has superior performance in every objective. Identifying a strong approximation set for the Pareto-optimal front can be extremely valuable in design problems because it effectively provides an optimal design given any arbitrary choice of objective priorities by the designer. If practitioners seek to generate a diverse set of near-optimal designs, they may seek to quantify how close a set of generated designs is to the true Pareto-optimal front. Below, we discuss a few metrics that could be adopted by design researchers to quantify performance optimality and refer readers to more detailed reviews, such as~\cite{Riquelme2015performance} for other metrics.

\paragraph{Hypervolume Metric~(Figure~\ref{fig:HV})}
The hypervolume metric, often simply referred to as `hypervolume,' is a staple metric in the multi-objective optimization field~\cite{Riquelme2015performance}, which estimates the proximity of a set of generated samples to the (often unknown) Pareto-optimal front. In simple terms, the metric calculates the hypervolume comprised of all points that are dominated by some point in the generated set but simultaneously dominate some fixed reference point. This leading optimization metric has seen previous use in DGM evaluation, such as in ~\cite{chen2021mopadgan} to compare the performance of different GAN models for airfoil synthesis.

\paragraph{Generational Distance~(Figure~\ref{fig:GD})}
When a set of `optimal' reference designs is known, generational distance can be used to measure design optimality. Generational distance, another staple of the multi-objective optimization community~\cite{Riquelme2015performance}, measures the distance from a generated sample to the nearest point on an `optimal' reference set~\cite{van1999multiobjective}. This reference set may not consist of truly optimal designs but is typically taken as a reliable approximation for a true Pareto-optimal design set. Generational distance is not widely adopted as a metric for deep generative models but can serve as an excellent metric for evaluating the quality of generated designs in cases when an optimality frontier is known or could be constructed from the training data using non-dominated ranking methods. 

\paragraph{Optimality Gap~(Figure~\ref{fig:OG})}
When working on well-defined problems, practitioners may be able to perform gradient-based optimization on their problems. In these cases, they can estimate the distance to an optimal design using `optimality gap' metrics. The distance from the generated design to the local minimum discovered by the optimizer is often referred to as the `cumulative optimality gap' or simply the `optimality gap.' A variant, the instantaneous optimality gap, has also been proposed to measure the distance to the modified design after the first step of gradient descent~\cite{chen2022inverse}. 

\subsection{Target Achievement} \label{targets}
While generating an entire set of Pareto-optimal designs can be helpful when exact design goals are not yet decided, practitioners may also need to apply generative models to design problems where performance targets are specified. These types of problems necessitate a suite of metrics that quantify a model's ability to achieve performance targets. We would like to clarify that, in contrast to hard design constraints, performance targets are intended to be negotiable. They are also different from soft constraints, as exceeding a target by a larger margin is often desirable, which is not the case with constraints.
While many of the metrics for constraint satisfaction can be modified to quantify target achievement, the handling of design targets can call for more nuanced metrics which reflect their flexibility.

\paragraph{Target Achievement Scores} \label{sec:TAsimple}
Reformulating several of the previously discussed constraint satisfaction metrics, we can define analogs for the target achievement case. Target achievement~(Figure~\ref{fig:TA}), much like the constraint satisfaction score, measures whether a design simultaneously meets all performance targets across objectives. However, since simultaneously achieving all the targets in a given problem may be difficult, more nuanced scores are typically more informative in quantifying proximity to the target. Target achievement rate~(Figure~\ref{fig:TAR}) is analogous to the constraint satisfaction rate, quantifying the weighted proportion of multi-objective design targets met by a design. 
When practitioners have a well-defined target criterion, they can calculate a signed distance to target~(Figure~\ref{fig:SDT}), indicating the degree to which a design is outperforming or underperforming the set of multi-objective design targets. 

\paragraph{Design Target Achievement Index~(Figure~\ref{fig:DTAI})}
In problems with particularly nuanced design goals and targets, practitioners may want an even more flexible metric than the signed distance to the target. Regenwetter et. al.~\cite{regenwetter2022design} proposed the design target achievement index (DTAI), which considers the relative importance of targets and the value of continued optimization beyond the target. The key idea is to aggregate weighted soft penalties when a target in any of the objectives is not achieved and combine them with decaying rewards when the target is exceeded. DTAI is also bounded and differentiable, making it viable as a training loss in generative design problems where design performance values are provided in the dataset.

\subsection{Demonstration of Design Quality Metrics}\label{perfdemo}
\begin{figure}[!htb]
     \centering
     % \hfill
     \begin{subfigure}[b]{0.32\textwidth}
         \centering
         \includegraphics[width=\textwidth]{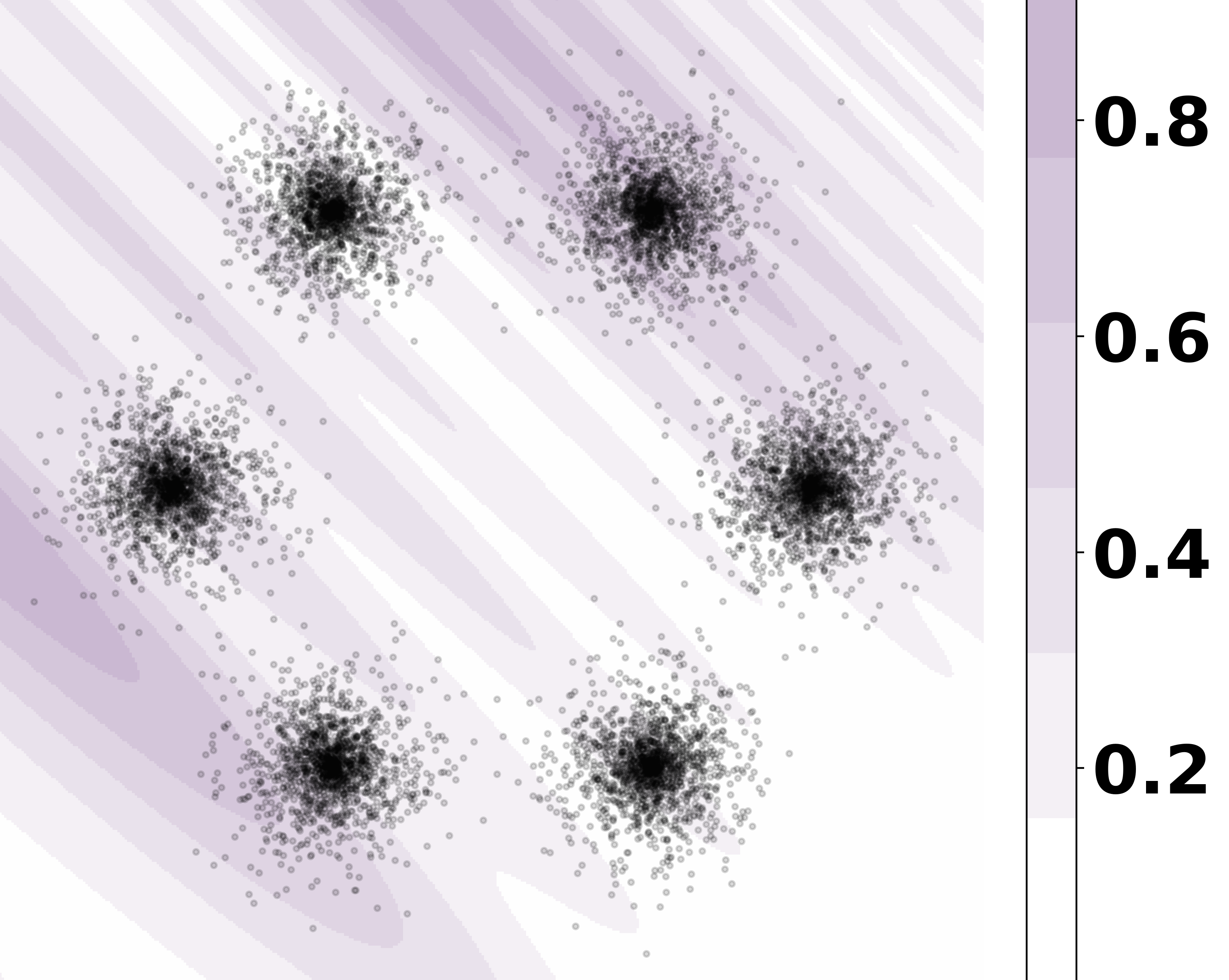}
         \caption{Objective 1}
         \label{fig:PO1}
     \end{subfigure}
     \hfill
     \begin{subfigure}[b]{0.32\textwidth}
         \centering
         \includegraphics[width=\textwidth]{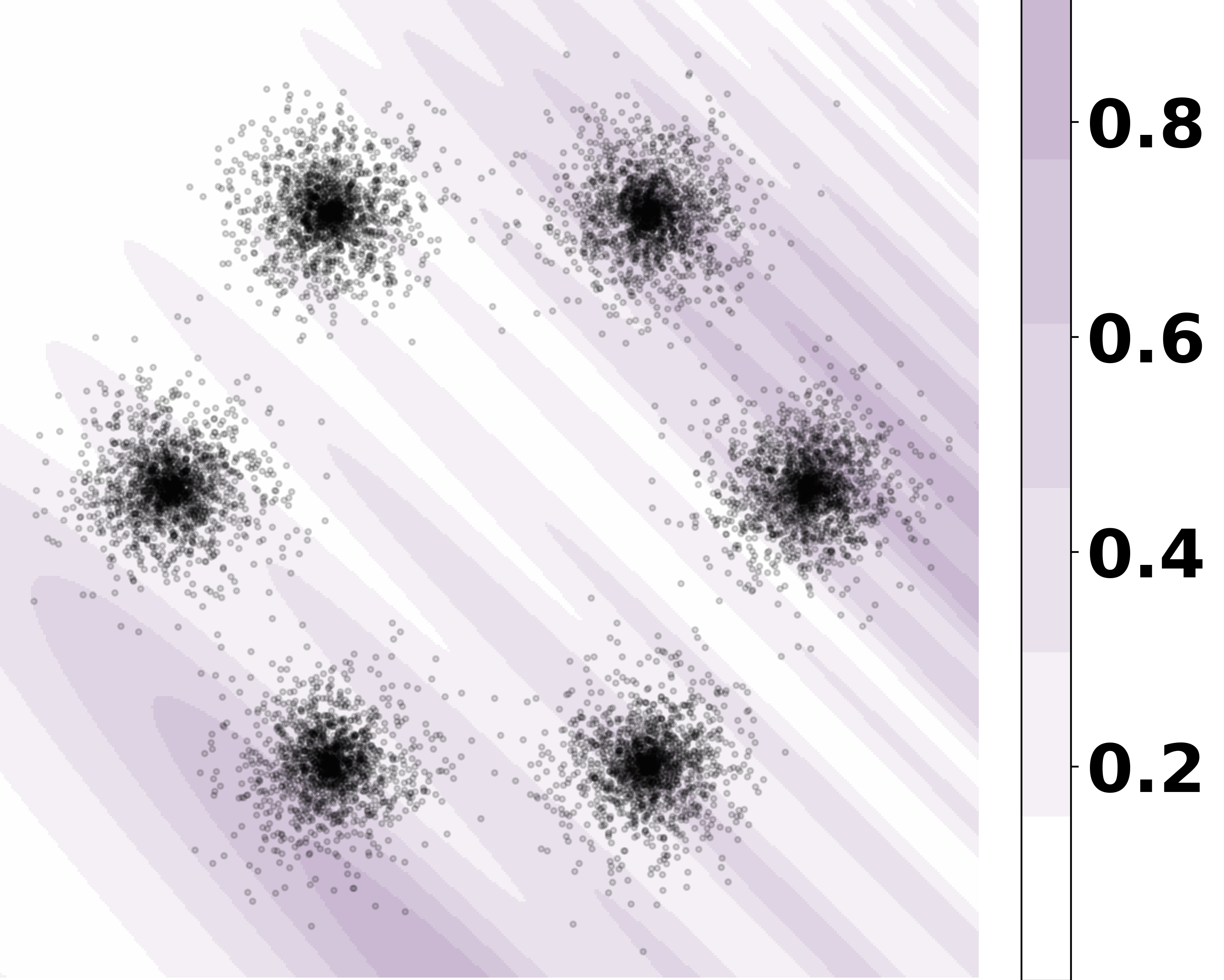}
         \caption{Objective 2}
         \label{fig:PO2}
     \end{subfigure}
     \hfill
     \begin{subfigure}[b]{0.32\textwidth}
         \centering
         \includegraphics[width=\textwidth]{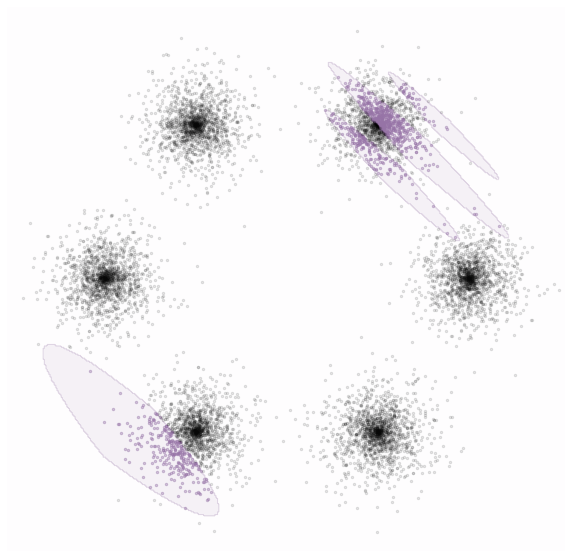}
         \caption{Target-achieving region}
         \label{fig:PT}
     \end{subfigure}

     \begin{subfigure}[b]{0.32\textwidth}
         \centering
         \includegraphics[width=\textwidth]{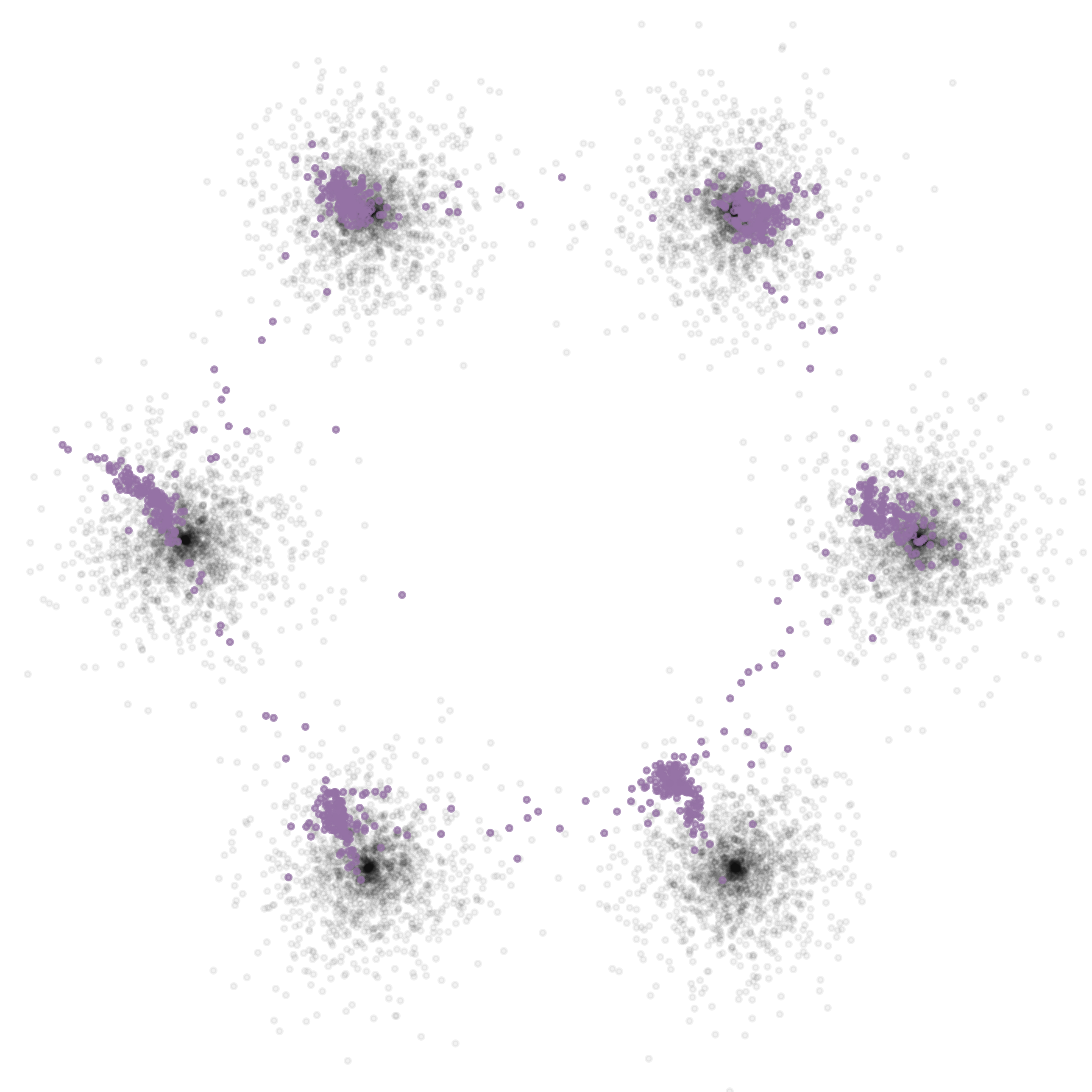}
         \caption{GAN}
         \label{fig:PGAN}
     \end{subfigure}
     \begin{subfigure}[b]{0.32\textwidth}
         \centering
         \includegraphics[width=\textwidth]{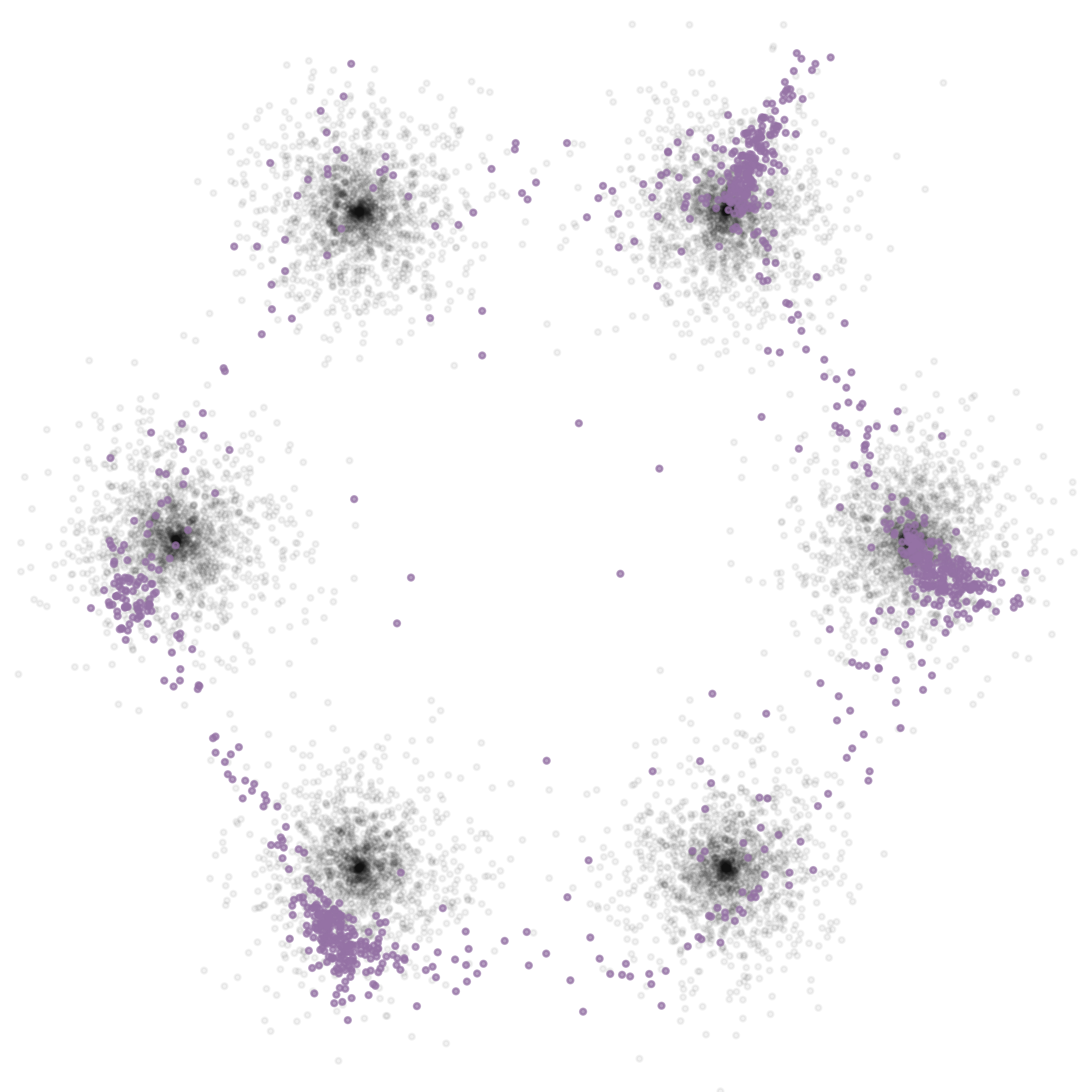}
         \caption{MO-PaDGAN}
         \label{fig:PMOPADGAN}
     \end{subfigure}
    \caption{Visual demonstration of a Generative Adversarial Network and Multi-Objective Performance-augmented Diverse GAN on the two-dimensional KNO1 objective. The MO-PaDGAN, which considers functional performance samples predominantly from higher-performing modes.}
    \label{fig:P}
\end{figure}

Having introduced a variety of performance and target achievement metrics for DGMs in design, we once again present a two-dimensional example demonstrating the use of these metrics. We select the KNO1 test problem as the objective function~\cite{knowles2006parego} with a restrictive minimum performance target of $(0.5, 0.5)$ and add it to the previously discussed two-dimensional example (Figure~\ref{fig:P}, Table~\ref{tab:P}). We test a standard GAN, which ignores performance, and a Multi-Objective Performance-augmented Diverse GAN (MO-PaDGAN)~\cite{chen2021mopadgan}, which attempts to generate a diverse set of high-performing designs. Model and metric parameters are included in the appendix. The MO-PaDGAN largely ignores the two low-performance modes and outperforms the standard GAN in every performance metric tested. It achieves a stronger Pareto-optimal set as indicated by the hypervolume metric. It generates designs that are, in general, closer to the known Pareto-optimal reference set, as indicated by the generational distance metric. Its designs achieve a greater fraction of the performance targets, lie closer to the target boundary, and adhere more closely to the target overall, as indicated by the weighted target achievement rate, signed distance to target, and design target achievement index, respectively. These metrics match our intuition and provide a way to quantify differences between different models: A standard GAN underperforms in all metrics, as it only maximizes distribution similarity, without any consideration for other factors such as objectives and targets. Note that the GAN would dominate the MO-PaDGAN in almost every statistical similarity metric. In design problems, higher functional performance often comes at the expense of statistical similarity. 

\begin{table}[!htb]
\centering
\caption{Performance and target achievement scores for distributions in Figure~\ref{fig:P}.The MO-PaDGAN, which considers functional performance, samples predominantly from higher-performing modes. Point metrics are averaged over the generated set. Bold is better.}
\begin{tabular}{ccc}
\toprule
Metrics & GAN & MO-PaDGAN \\
\midrule
Hypervolume               & 0.460  & \textbf{0.571}     \\
Generational Distance     & 0.215  & \textbf{0.315}     \\
Weighted Target Ach. Rate & 0.162  & \textbf{0.430}     \\
Signed Distance to Target & -0.275 & \textbf{-0.154}    \\
Design Target Ach. Index  & 0.324  & \textbf{0.433} \\
\bottomrule
\end{tabular}%

\label{tab:P}
\end{table}

\section{Evaluating Conditioning Requirements}

In DGMs, conditioning refers to the process of incorporating additional information, such as labels or attributes, into the model when generating new data. For example, in image generation, a deep generative model can be conditioned on class labels, such as `dog' or `cat', in order to generate images of specific animals, rather than randomly picking cats or dogs (or training separate generators for each class of image). Conditional DGMs are often more robust and data-efficient than training many individual DGMs since they allow designers to reuse a single model for many variants of a particular design problem.  

Conditional variants of many classic generative models have been proposed~\cite{mirza2014conditional, chen2016infogan, sohn2015learning}, but are most commonly applied to class-conditional problems. In design, many tasks necessitate continuous conditioning, and specialized models have been proposed~\cite{pcdgan, rangegan}. Conditioning information varies from problem to problem. It is commonly used to encode constraints, functional performance targets, or parameters used to distinguish one version of a problem from another. For example, DGMs for structural topologies are typically conditioned on boundary conditions (normally thought of as constraints), volume fraction (sometimes thought of as a functional performance attribute), and load locations (used to distinguish different loading problems)~\cite{nie2021topologygan, maze2022topodiff}.

\begin{figure}[!htb]
    \centering
    \begin{subfigure}[b]{0.32\textwidth}
     \centering
     \includegraphics[width=\textwidth]{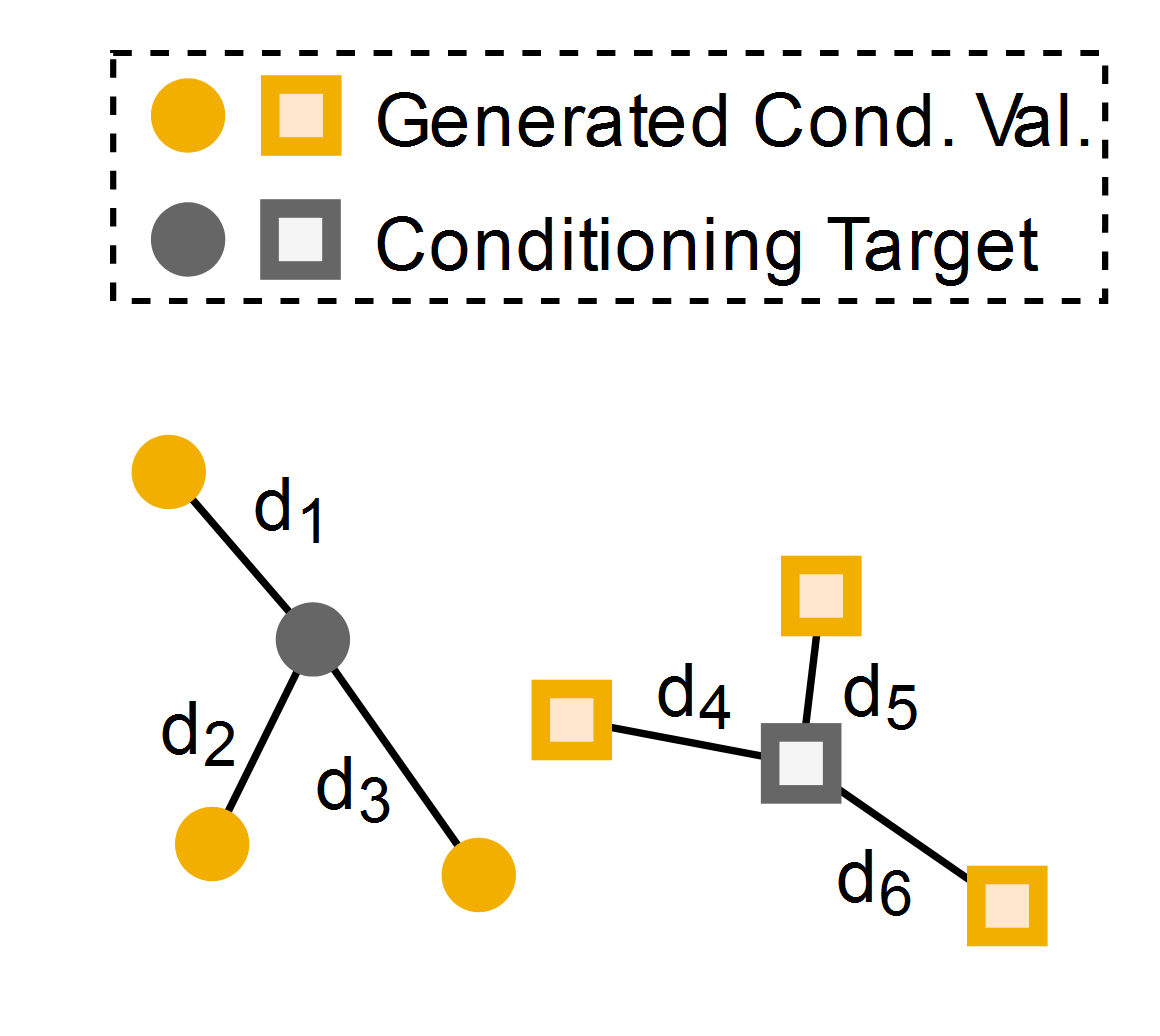}
     \caption{Conditioning Adherence}
     \label{fig:CA}
    \end{subfigure}
    \begin{subfigure}[b]{0.32\textwidth}
     \centering
     \includegraphics[width=\textwidth]{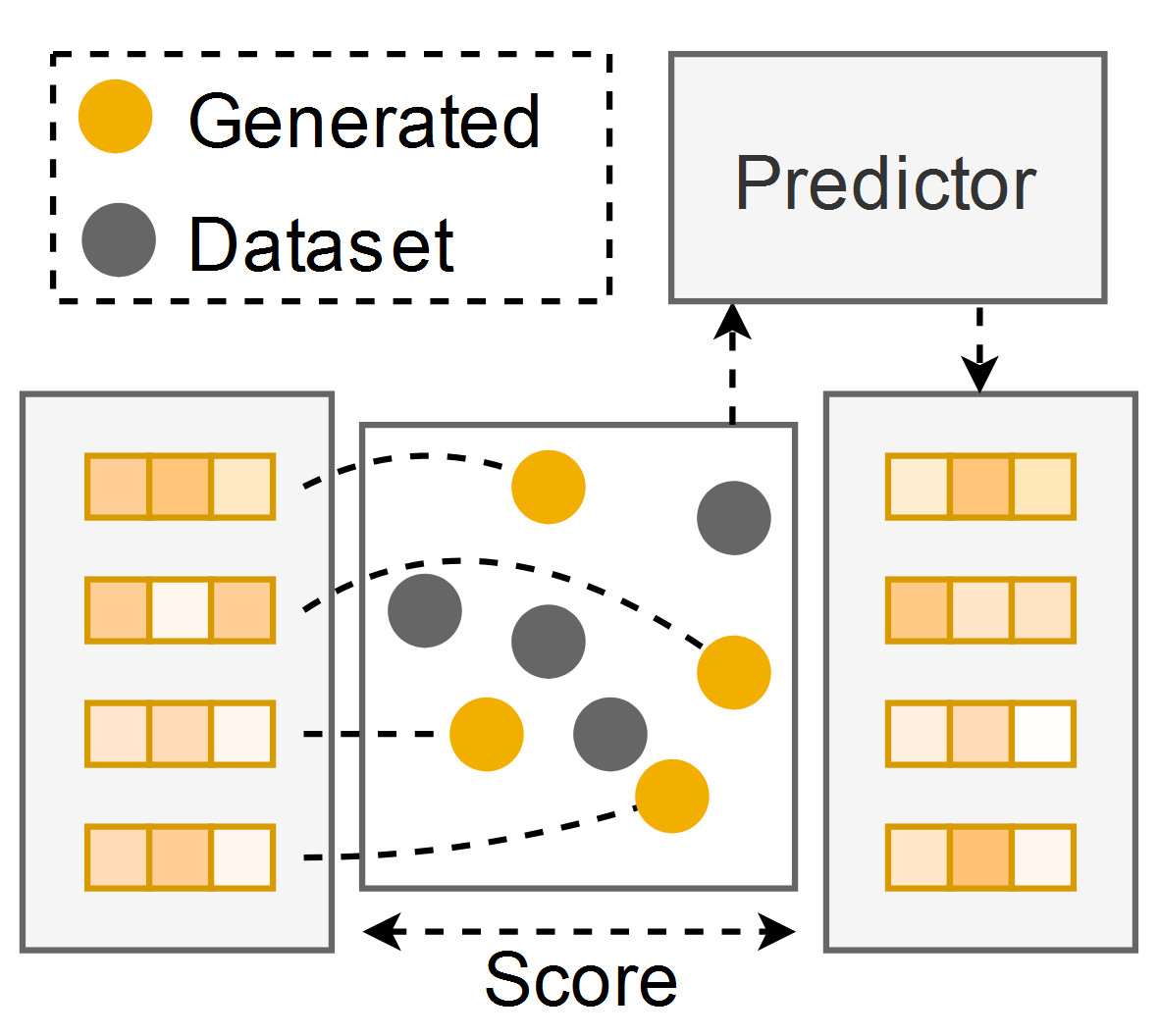}
     \caption{Conditioning Reconstruction}
     \label{fig:CR}
    \end{subfigure}
    \begin{subfigure}[b]{0.32\textwidth}
     \centering
     \includegraphics[width=\textwidth]{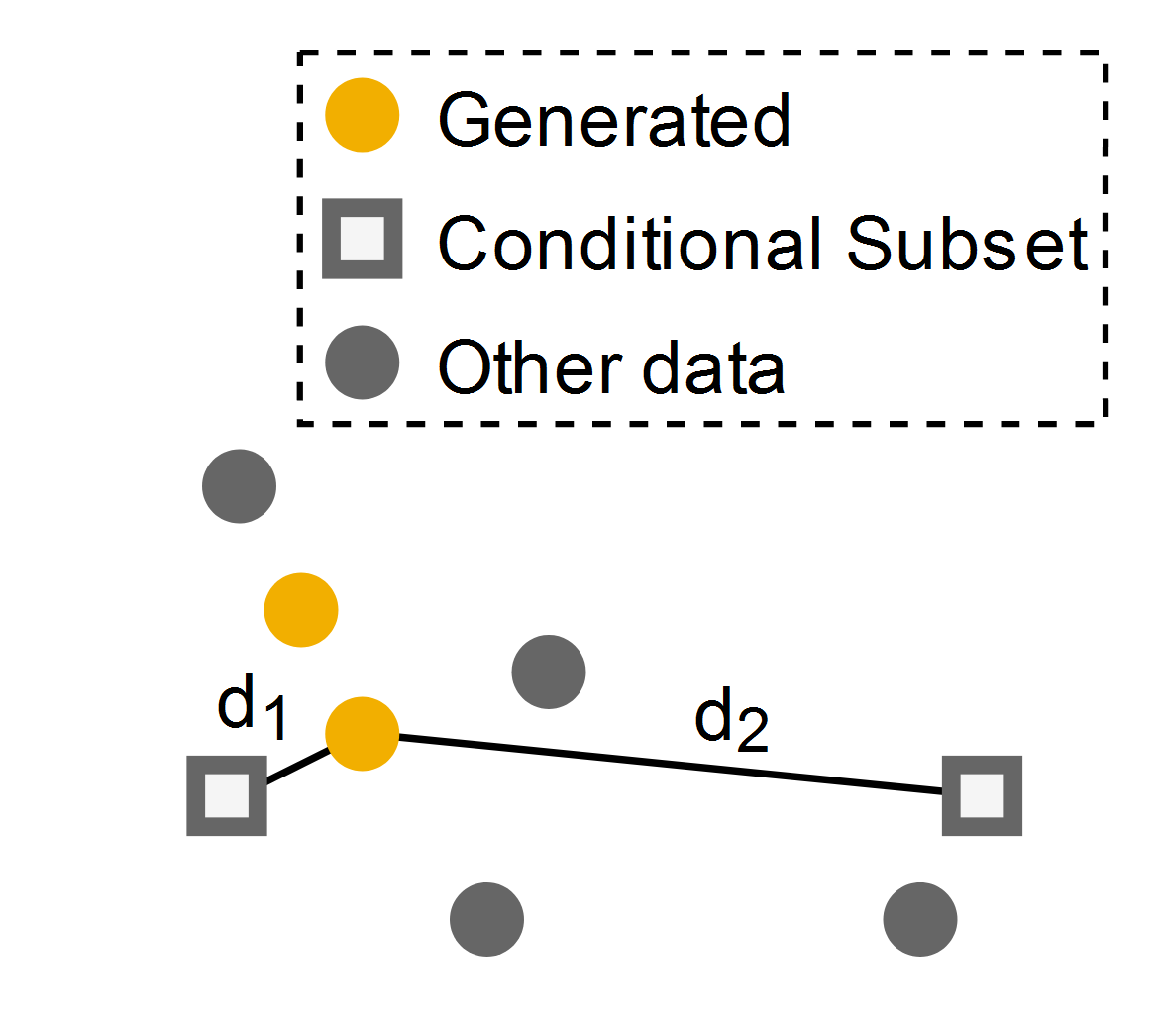}
     \caption{Cond. Nearest Gen. Sample}
     \label{fig:CNGS}
    \end{subfigure}
     
    \caption{Illustrations of select conditioning metrics.}
    \label{fig:Conditioning}
\end{figure}

\subsection{Conditioning Adherence~(Figure~\ref{fig:CA})} \label{sec:CA}
When working with conditional models, practitioners may want to quantify the degree to which their model respects the information on which it is conditioned. In some problems, practitioners are conditioning on information that they can directly calculate for generated samples. For example, if conditioning a structural topology DGM on volume fraction, the volume fraction of generated samples can be calculated as the fraction of filled pixels to total pixels. In this case, the difference in conditioned versus actual volume fraction serves as a conditioning adherence metric. Broadly speaking, the distance between the condition and the recalculated condition is a viable approach to quantify conditioning adherence. When exact conditioning information cannot be calculated, it can instead be estimated using a predictor (Conditioning reconstruction, (Figure~\ref{fig:CR})). Since the reconstruction loss serves as the score, the accuracy of the predictor directly impacts the metric. 

\subsection{Adapting Unary Metrics to Conditional Problems}
Many of the metrics presented in previous sections must be adapted for conditional problems. We first discuss strategies to adapt unary metrics to conditional settings and discuss the more challenging task of adapting binary metrics in the following section. Unary metrics must be calculated and averaged over numerous conditions. Naturally, we wouldn't compare Model A's conditional bike-generating ability with Model B's scooter-generating ability. We instead must report each model's average performance over an identical and comprehensive set of conditions or report scores for each condition of interest. For continuous conditioning problems, this may mean averaging over many conditions sampled from some repeatable parametric statistical distribution.

\subsection{Adapting Binary Metrics to Conditional Problems}
In unconditional problems, binary metrics typically compare the generated distribution to a reference distribution (often the dataset). When we instead have a conditionally-generated distribution, we can compare against a conditional reference or a marginal reference, approaches with their own strengths, which we discuss below. In either case, much like for unary metrics, we typically average scores over a sweep of conditions to yield an overall condition-agnostic score, or simply report metrics under several individual conditions. 

\paragraph{Comparing Generated Distribution to a Conditional Reference} 
Intuitively and most commonly, we would compare the conditionally-generated distribution to a conditional reference\footnote{This is the standard formulation for class-conditional variants of popular metrics like class-conditional FID and IS~\cite{benny2021evaluation}.}. Classically, this is the subset of the dataset that adheres to the prescribed condition\footnote{When the reference set is not the dataset, the concept of a conditional subset may not make sense. For example, the reference distribution in generational distance is a Pareto-optimal set. The conditional subset of this reference may have significant gaps and may no longer be a strong reference set, defeating the purpose of the metric.}. We have illustrated the conditional nearest generated sample metric in Figure~\ref{fig:CNGS}, but the principle applies broadly to many binary metrics. 

Identifying a conditional data subset is often nontrivial in continuous conditioning problems since we can't select a discrete subset of the dataset which exactly respects the condition. Instead, we can approximate by selecting a subset of the dataset whose conditions are proximal to the condition used to create the generated distribution. However, this necessitates calculating some distance in the conditioning space, such as the vicinal loss defined in literature~\cite{pcdgan, ding2020ccgan}. Calculating these vicinity-based values may again be nontrivial in problems with high-dimensional or multimodal conditioning information. 

\paragraph{Comparing Generated Distributions to a Marginal Reference}
Alternatively, we can compare the conditionally-generated distribution to the full marginal reference distribution. This may serve as a next-best option when calculating a conditional reference is intractable. However, comparing conditionally-generated distributions to a marginal reference may actually be uniquely illustrative for certain metrics such as nearest datapoint. Say, for example, that we are conditionally generating lightweight bikes using a DGM. The dataset contains only heavy mountain bikes and light road bikes. Therefore, the conditional subset contains only light road bikes. If our DGM generates light mountain bikes, it will score well if compared to the full (marginal) dataset, which includes mountain bikes, and poorly if compared to the conditional subset of only road bikes. On one hand, discovering adaptations of designs to meet the specified condition may be innovative and desirable. On the other hand, lightweight mountain bikes are `unrealistic', according to the data. The two variants of nearest datapoint provide different insights in the conditional setting. 

\subsection{Demonstration of Conditioning Metrics} \label{conddemo}

\begin{figure}[!htb]
     \centering
     \begin{subfigure}[b]{0.32\textwidth}
         \centering
         \includegraphics[width=\textwidth]{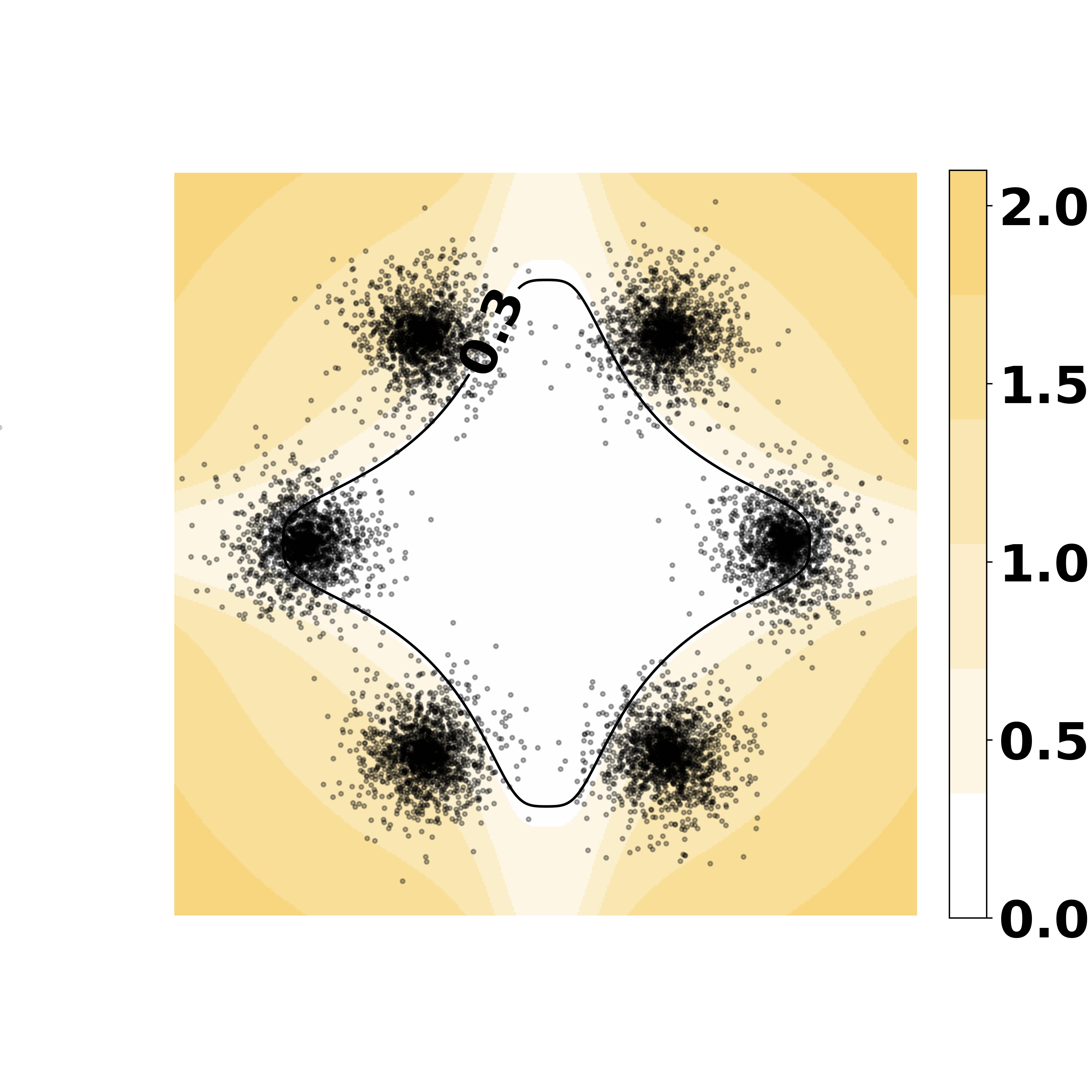}
         \caption{Original data}
         \label{fig:CO1}
     \end{subfigure}
     \begin{subfigure}[b]{0.32\textwidth}
         \centering
         \includegraphics[width=\textwidth]{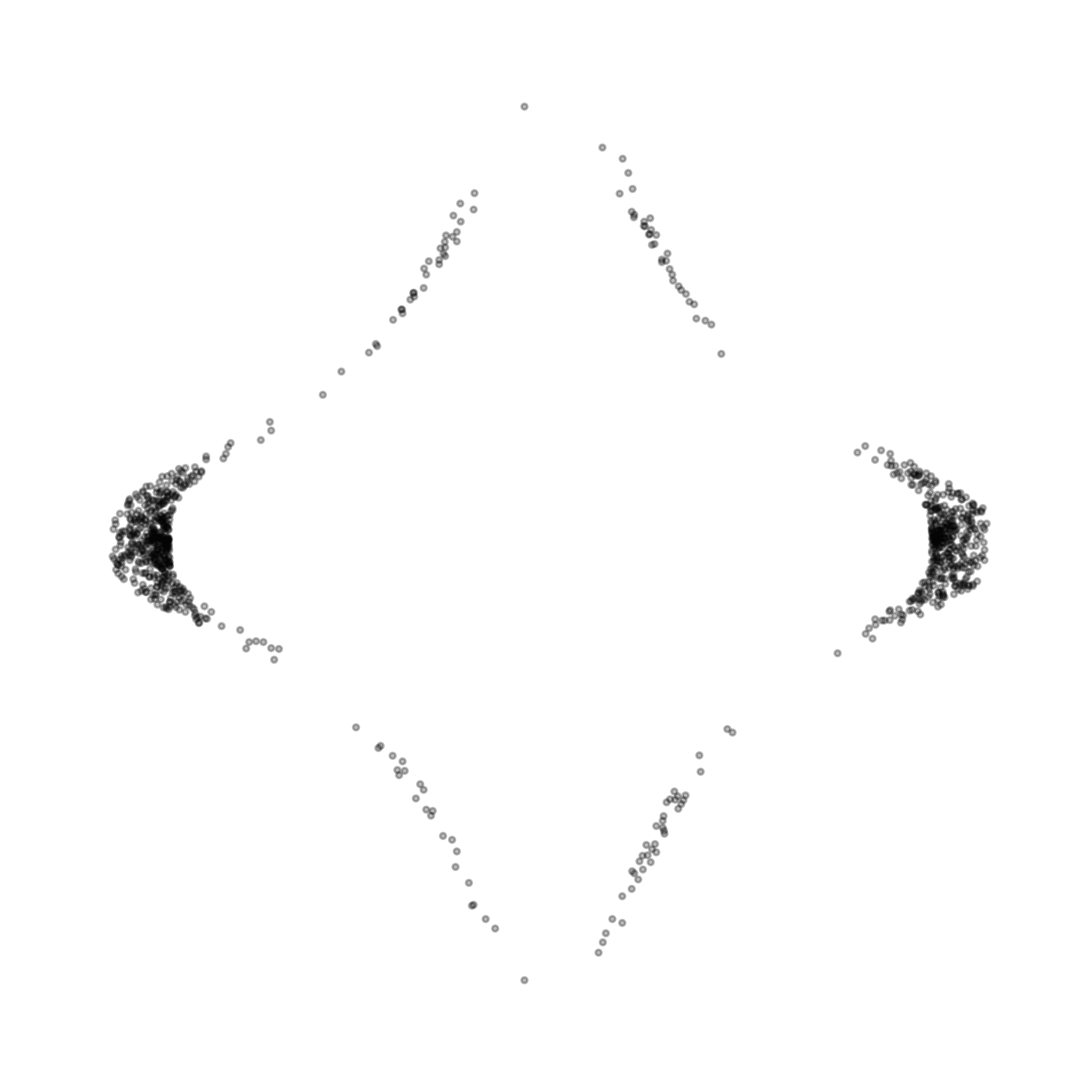}
         \caption{Conditional Prior}
         \label{fig:CO2}
     \end{subfigure}
     \begin{subfigure}[b]{0.32\textwidth}
         \centering
         \includegraphics[width=\textwidth]{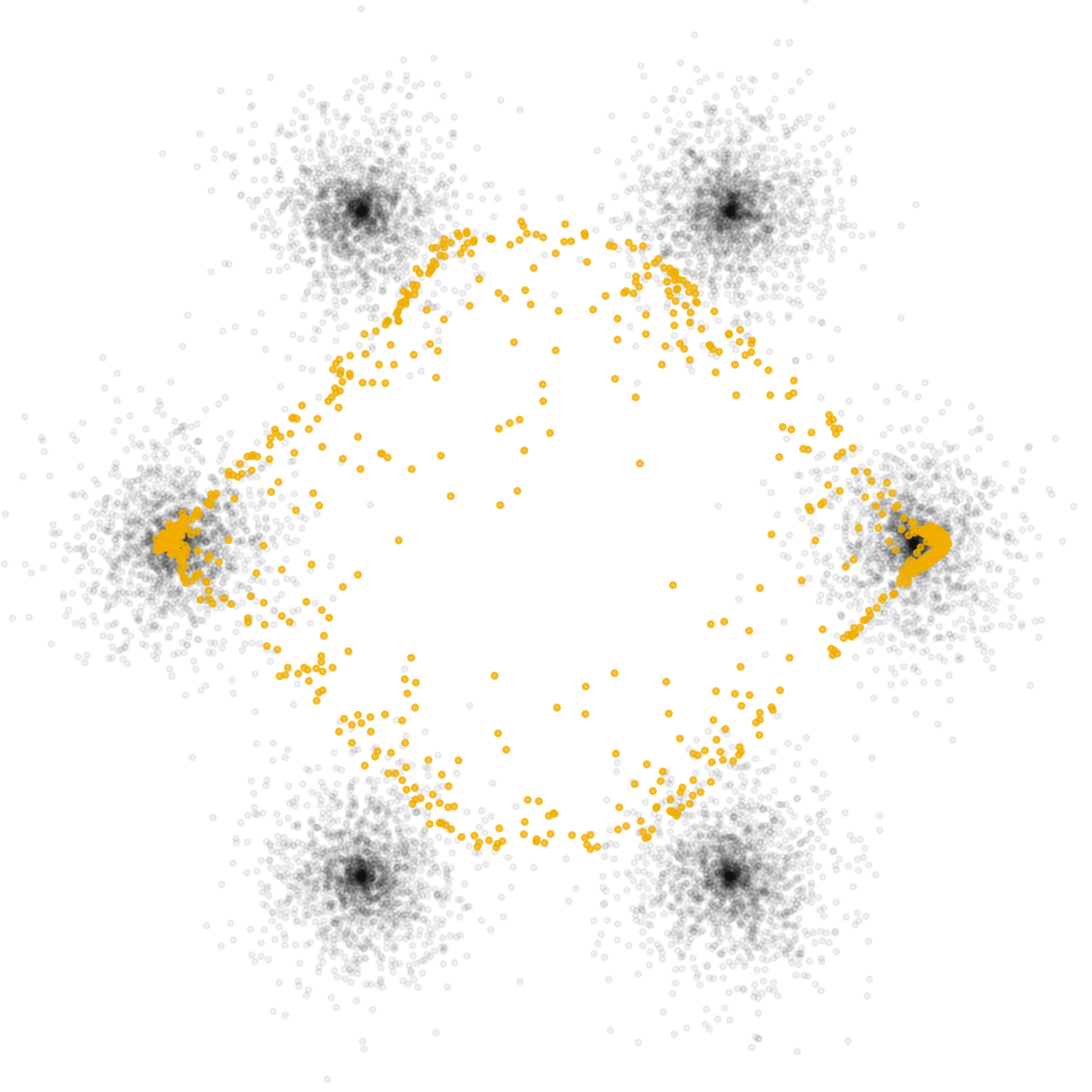}
         \caption{cVAE vs. marginal prior}
         \label{fig:CO3}
     \end{subfigure}

     \begin{subfigure}[b]{0.32\textwidth}
         \centering
         \includegraphics[width=\textwidth]{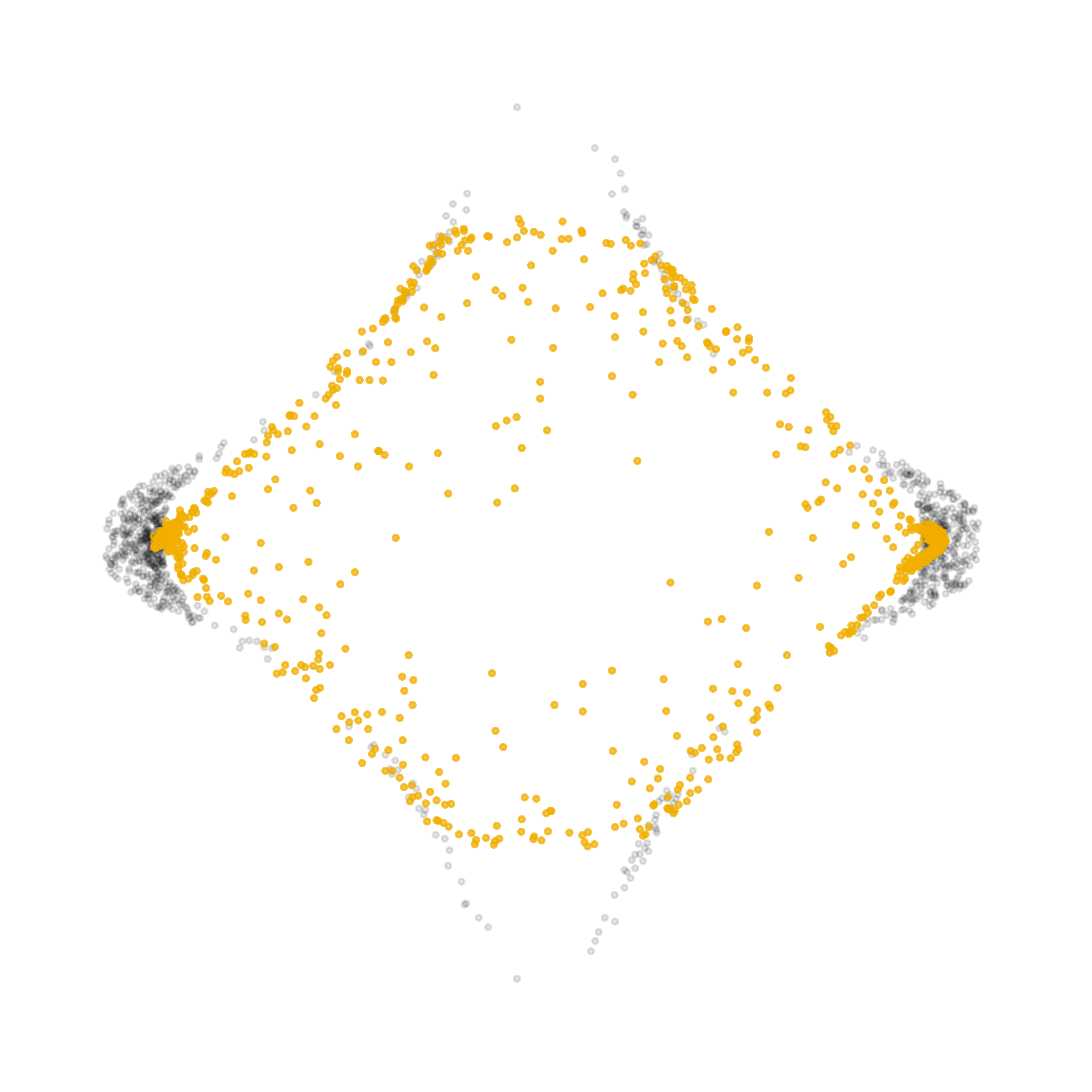}
         \caption{cVAE vs. conditional prior}
         \label{fig:CO4}
     \end{subfigure}
     \begin{subfigure}[b]{0.32\textwidth}
         \centering
         \includegraphics[width=\textwidth]{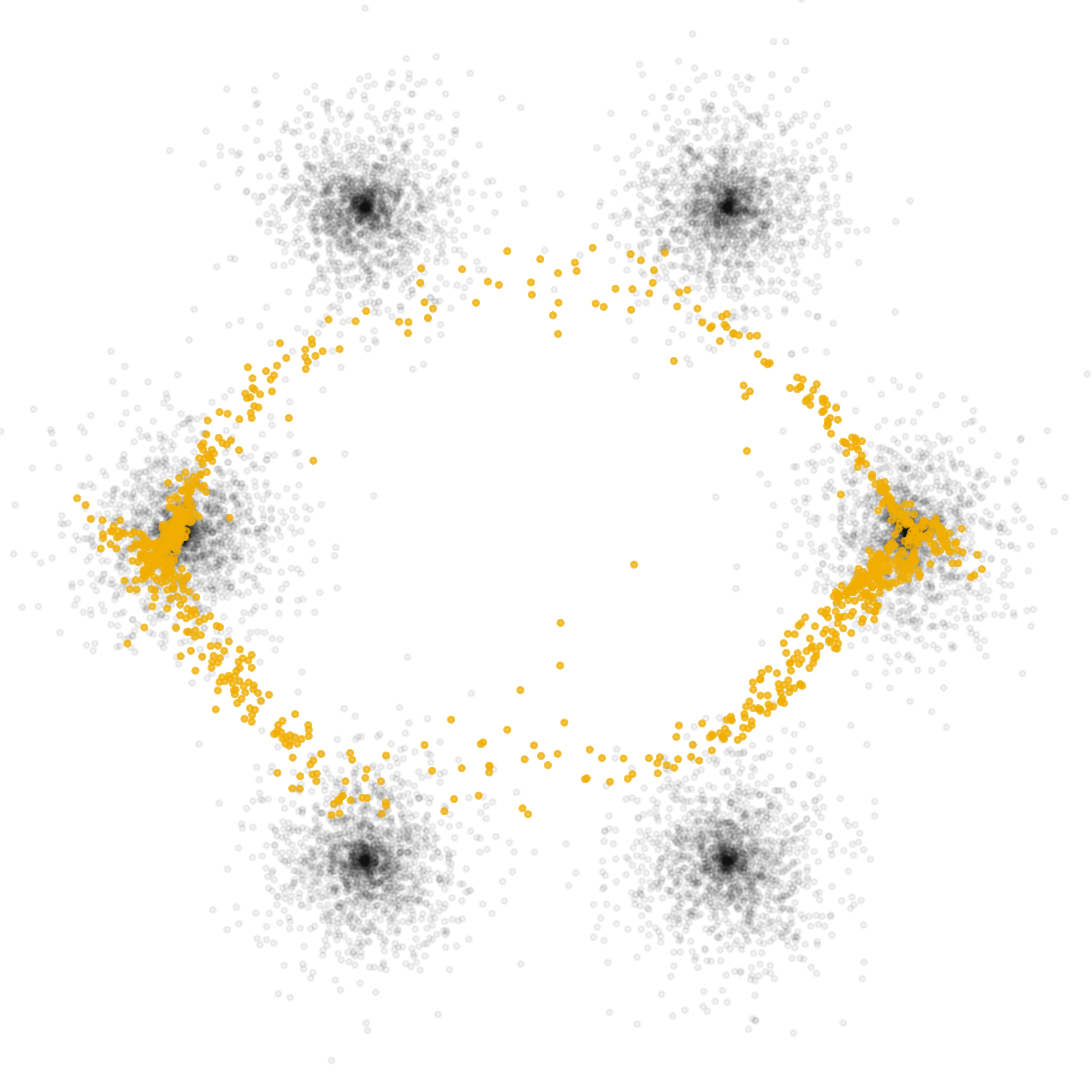}
         \caption{cGAN vs. marginal prior}
         \label{fig:CO5}
     \end{subfigure}
     \begin{subfigure}[b]{0.32\textwidth}
         \centering
         \includegraphics[width=\textwidth]{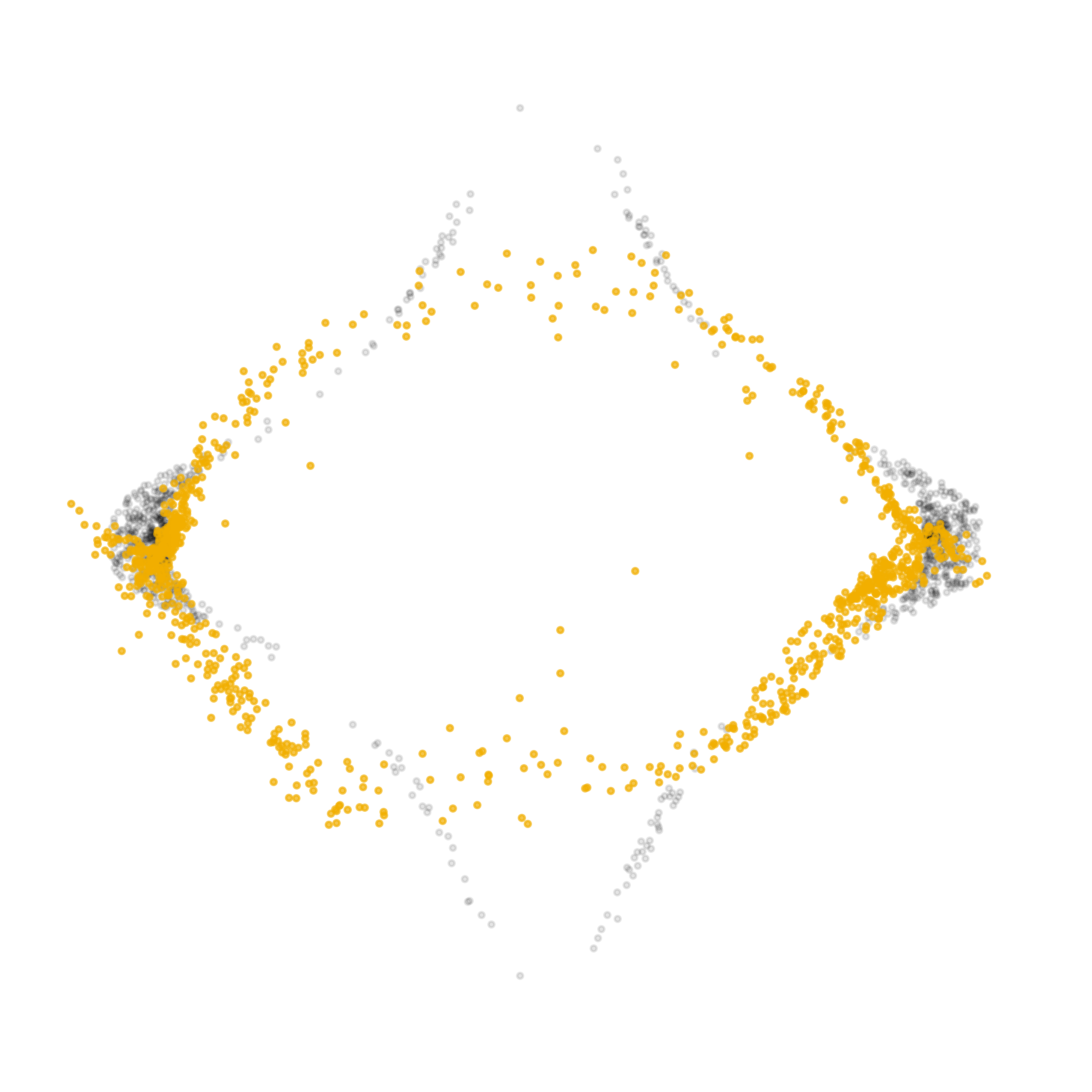}
         \caption{cGAN vs. conditional prior}
         \label{fig:CO6}
     \end{subfigure}

    \caption{Visual demonstration of a conditional Variational Autoencoder and conditional GAN on a continuous conditioning problem. Generated distributions are overlaid over both the original dataset (marginal prior), as well as the nearest 10\% of the data to the condition (conditional prior). The cVAE better matches the marginal prior, while the cGAN better matches the conditional prior.}
    \label{fig:Cond}.  
\end{figure}
To demonstrate the conditioning of generative models, we construct a fairly challenging continuous conditioning problem on the previously-used dataset. Each of the datapoints is labeled according to a highly nonlinear conditioning function. For simplicity, we only examine a single condition value of exactly 0.3, though we would typically care about and average performance across a variety of condition values. We train a conditional VAE and a conditional GAN (Figure~\ref{fig:Cond}, Table~\ref{tab:Cond}). Details about the model architecture, training, and metric settings are included in the appendix. We take the conditional prior to be the 10\% of the dataset that most closely matches the condition, as shown in~\ref{fig:CO2}. Overall, the cGAN focuses heavily on the leftmost and rightmost modes, struggles to accurately capture other areas of the distribution, and ignores sparse areas. In contrast, the cVAE under-represents the two main modes, but much more faithfully captures the remainder of the distribution. Overall, the cVAE significantly outperforms the GAN in conditioning reconstruction and adherence, indicating that it generated samples that, on average, much more closely match the condition. In both conditional and marginal F10 and F0.1, the cVAE and cGAN repeat trends from their unconditional counterparts in Section~\ref{simdemo}. Due to its focus on the main modes, the GAN significantly outperforms in maximum mean discrepancy when compared to the subset of the dataset that most closely matches the condition. However, when compared to the marginal prior, the cVAE is the clear leader in MMD, generating more samples that lie closer to the four underrepresented modes. As illustrated, binary metrics can tell very different stories when compared to conditional or marginal priors.

\begin{table}[!htb]
\centering
\caption{Performance and target achievement scores for distributions in Figure~\ref{fig:Cond}. The cVAE better matches the marginal prior, while the cGAN better matches the conditional prior on this dataset. Point metrics are averaged over the generated set. Bold is better.}
\begin{tabular}{ccc}
\toprule
Metrics  & cVAE & cGAN \\
 \midrule
Conditioning Reconstruction & \textbf{0.007} & 0.026          \\
Conditioning Adherence      & \textbf{0.008} & 0.026          \\
\midrule
Conditional F10             & \textbf{0.909} & 0.826          \\
Conditional F0.1            & 0.879          & \textbf{0.960} \\
Conditional MMD             & 0.035          & \textbf{0.021} \\
\midrule
Marginal F10                & \textbf{0.949} & 0.933          \\
Marginal F0.1               & 0.447          & \textbf{0.554} \\
Marginal MMD                & \textbf{0.068} & 0.097    \\
\bottomrule
\end{tabular}%
\label{tab:Cond}
\end{table}

% \section{Application Study 1: Diverse Topology Generation with Constraints}
\section{Application Study: Exploratory Bicycle Frame Design with Constraints and Performance Targets}

Throughout the paper, we have applied metrics to simple two-dimensional problems to visually demonstrate how the presented metrics function. However, real design problems are typically higher-dimensional, often have more constraints and objectives, and are generally highly non-convex. To showcase that the discussed metrics are effective in real design problems, we include two case studies, the first of which is presented in this section. In this problem, we want to design novel, diverse, and high-performing bicycle frames that meet a set of ten structural performance targets and adhere to an unknown set of implicit design constraints. Specifically, we seek a DGM that consistently generates designs meeting performance targets and constraints, but generates a wide enough variety to offer a broad selection of design candidates. 

\subsection{Dataset}
The dataset we use~\cite{regenwetter2022framed} features roughly 4000 constraint-satisfying designs and roughly 300 constraint-violating designs. Each frame design is inspired by a real bicycle design~\cite{regenwetter2022biked} and is parameterized over 37 variables. Each constraint-satisfying design is `labeled' with a vector of 10 structural performance values, such as weight, safety factors, and deflections under various loading conditions. This dataset was previously used as a benchmark for target-seeking deep generative models in ~\cite{regenwetter2022design} and we adapt the models tested for our demonstration. We also adopt the objective weights and DTAI parameters from~\cite{regenwetter2022design}. 

\subsection{Models}
We train and test three GAN variants. The first is a `vanilla' GAN, which is blind to design performance, only implicitly observes constraints, and does not promote exploration beyond the convex region of the dataset. The second is a Multi-Objective Performance-augmented Diverse GAN (MO-PaDGAN) which uses a performance-weighted DPP kernel to simultaneously encourage higher performance and greater diversity among generated designs. MO-PaDGAN generally encourages diverse, higher-performing designs, but does not consider specific performance targets or explicit constraint handling. The third is a DTAI-GAN which modifies the MO-PaDGAN to consider performance targets and classifier guidance to avoid constraint-violating designs. In this sense, DTAI-GAN is the only model that `explicitly' considers constraints, while the other models `implicitly' consider constraints by training only on constraint-satisfying designs. 

\subsection{Selecting Metrics}
We would like to evaluate diversity, performance, target achievement, and constraint satisfaction, alongside distribution matching. We discuss how we select metrics for this problem and justify these choices. 
\paragraph{Statistical Similarity} While the goal is not just to mimic existing designs, measuring similarity is important to make sure that we are still generating bicycle frames. We would also ideally like our generated designs to span as much of the design space as possible, thereby representing as many key types of designs as possible. As such, we care about both `realism' and coverage, as well as general similarity. For this, we select nearest datapoint (NDP), nearest generated sample (NGS), and F1 to capture realism, coverage, and similarity, respectively. Since we primarily care about these metrics in the design space, we evaluate all three in the design space, rather than the performance space. 

\paragraph{Design Exploration} As one of the stated objectives of the problem, we want to generate a diverse set of novel designs. This is common in scenarios where a final design will be selected by experts from a diverse set. While we could look to maximize nearest datapoint as a novelty metric, we still want our designs to resemble designs in the dataset. We instead calculate nearest datapoint in the performance space, instead of the design space. Put plainly, while we want generated designs to resemble existing designs, we would like them to have different performance values. Since we also want generated designs to be diverse, we select inter-sample distance and DPP diversity as two diversity metrics of choice, which we evaluate in the design space.

\paragraph{Design Constraints} The FRAMED problem provides a method to evaluate constraint adherence for generated samples, making the constraint satisfaction metric a natural choice. Since closed-form constraint definitions are not available, more informative choices such as signed distance to constraint boundary are not feasible in this problem. 

\paragraph{Design Quality and Target Achievement} 
We would like to evaluate the overall optimality of generated designs but lack both a Pareto-optimal reference set to use generational distance~\footnote{While we could infer an approximate set from the dataset, we have no basis for assuming that dataset designs are near-optimal, in part due to a randomization step in the FRAMED dataset generation pipeline~\cite{regenwetter2022framed}.} and a differentiable solver to use optimality gap. Instead, we select hypervolume. To capture target achievement, we use a weighted target achievement rate as a simple metric, but we can also use signed distance to target since we have access to simple closed-form target criteria. Finally, we use the design target achievement index (DTAI) to evaluate overall target achievement performance. 

\subsection{Results}
\begin{figure}[!htb]
    \centering
    \includegraphics[width=0.81\textwidth]{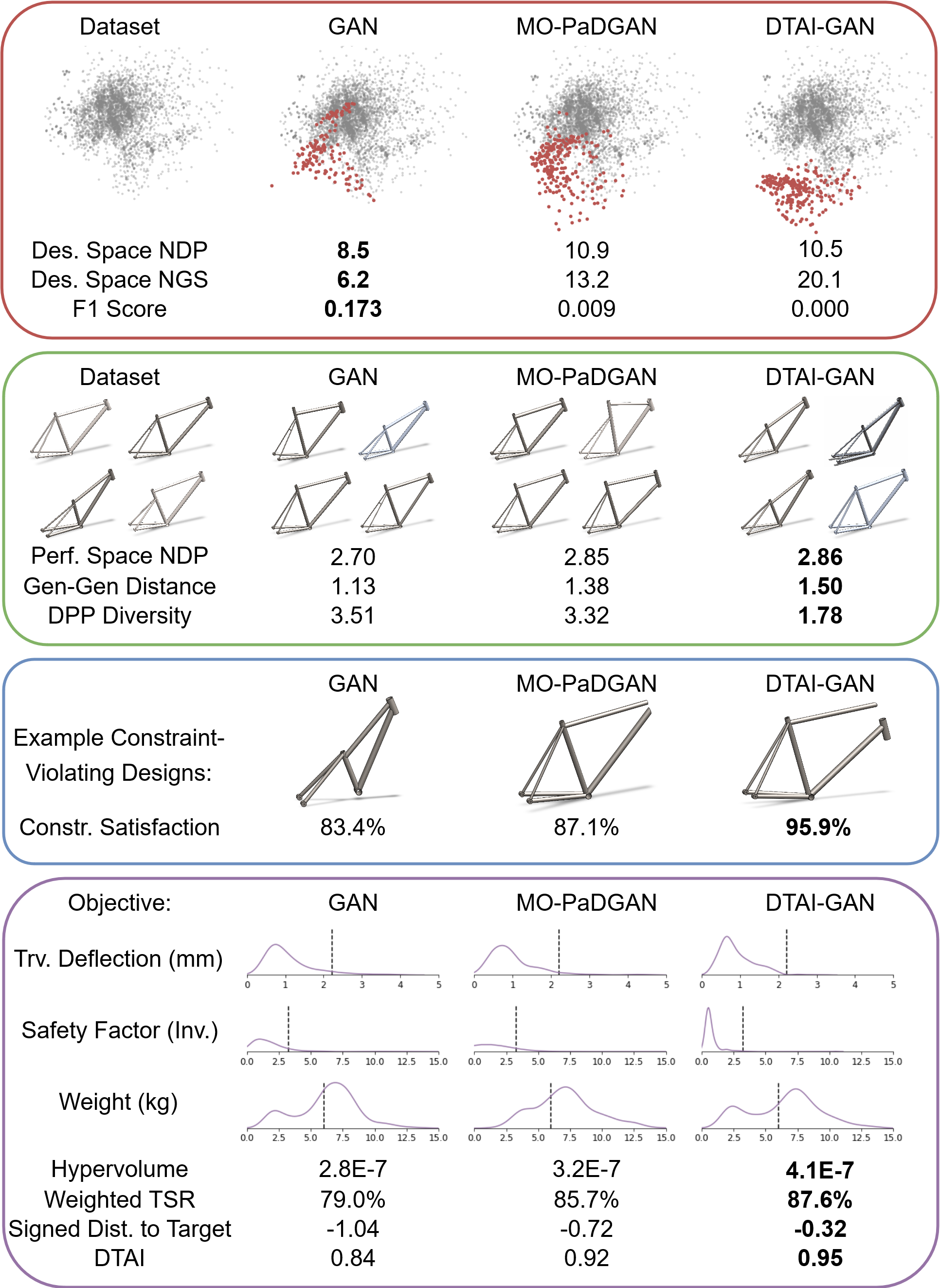}
    \caption{Evaluation of three models on the FRAMED dataset. The first box, focused on similarity shows a 2-D Principal Component Analysis embedding of generated designs overlaid over the dataset. The second box, focused on diversity shows select generated bike frames. The third box, focused on constraint satisfaction shows select invalid bike frames. The last box, focused on functional performance shows kernel density plots of structural performance scores of generated bikes (design target shown as the dotted line; smaller is better).}
    \label{fig:framed}
\end{figure}

We demonstrate the performance of the three DGMs and their scores on selected metrics in Figure~\ref{fig:framed}. 

\paragraph{Statistical Similarity} To assist with visualization, generated designs are projected onto a 2D space using Principal Component Analysis. Across the board, the vanilla GAN achieved superior distribution-matching performance to the GAN variants focused on design performance. In contrast, MO-PaDGAN and especially DTAI-GAN identified higher-performing or target-satisfying regions of the design space and consistently generated many designs away from the original design distribution, scoring lower on all distribution matching scores. Compared to MO-PaDGAN, DTAI-GAN's distributions were further from the dataset and resulted in lower F1 and nearest generated sample, but had a slight edge in nearest datapoint since there were a handful of dataset samples near DTAI-GAN's generated distribution. Though the distribution-matching scores indicate that the performance-aware models deviated significantly from the dataset, we don't necessarily find this concerning since we primarily care about finding novel high-performing designs. 

\paragraph{Design Exploration}
The DTAI-GAN scored highest in performance space novelty, generating many designs whose performance differs greatly from those in the dataset. These models also achieved much higher diversity, exploring more of the space and generating more varied samples. 

\paragraph{Design Constraints}
Since we don't have access to closed-form constraint tests, we are not able to evaluate distance to constraint boundary or other more insightful constraint-related metrics. Instead, we simply know that, as expected, the constraint-aware DTAI-GAN is a significantly stronger performer at constraint satisfaction than the two models that only implicitly considered constraints. 

\paragraph{Design Quality and Target Achievement}
Across all performance and target achievement metrics, DTAI-GAN scores the highest, and the vanilla GAN scores the lowest. Kernel density plots for three of the ten performance objectives are shown in the last panel of~\ref{fig:framed}, and in each, DTAI-GAN's distribution is most favorable. Interestingly, the DTAI-GAN's average target success rate was barely higher than MO-PaDGAN's. However, its average signed distance to the target was much higher, indicating that DTAI-GAN's designs that exceeded the target did so more drastically and designs that missed the target did so by a smaller margin. 

\subsection{Analysis}
The metrics demonstrate that the DTAI-GAN achieves its stated goal of generating a diverse and novel set of designs that achieve performance targets and satisfy constraints. The enhanced functional performance and constraint satisfaction of generated designs detracted from the DTAI-GAN's ability to match the training dataset. However, this behavior was largely expected and encouraged to discover higher-performing regions of the design space than were previously represented in the dataset. We now move on to our second case study, exploring optimal topology generation. 

\section{Application Study: Optimal Topology Generation using Diffusion Models} \label{sec:TO}
In this section, we demonstrate the appropriate selection of metrics for structural topology generation problems. Recent work~\cite{nie2021topologygan, behzadi2022gantl, maze2022topodiff, giannone2023diffusing} in the field has focused on training DGMs to circumvent the reliable but slower Topology Optimization (TO) solvers, such as Solid Isotropic Material with Penalization (SIMP)~\cite{bendsoe1988generating, rozvany1992generalized}. These DGMs train on a dataset of topologies generated by SIMP, typically taking the volume fraction and loading information as conditioning. Then, for various loading cases and volume fractions, they generate topologies that they predict SIMP would generate. Typically, the single functional performance objective is to minimize compliance of the generated topologies. 

\subsection{Metric Selection}
We first discuss existing metrics commonly used for evaluating DGMs in optimal topology generation, then propose several additional metrics that may be valuable. 
\paragraph{Existing Metrics}
Maz\'e \& Ahmed~\cite{maze2022topodiff} propose a set of four evaluation metrics for DGMs in optimal topology generation which have been adopted in later works~\cite{giannone2023diffusing}: Compliance error, volume fraction error, load violation, and floating material. We break down what these metrics mean, and which metrics from this paper they correspond to: 
\begin{enumerate}
    \item Compliance Error: Compliance error describes the percent difference between a generated topology's compliance and the compliance of a SIMP-generated topology under the same loading condition. This metric is a variant of the signed distance to target metric (Sec.~\ref{sec:TAsimple}). Compliance is treated as a functional performance objective in TO. For each conditional input, a target compliance value is selected to be the compliance of the topology generated by SIMP. The compliance error is then simply given as the normalized signed distance to this target. 
    \item Volume Fraction Error: Volume fraction error quantifies the percent error between a generated topology's volume fraction and the target volume fraction given to the generator. This metric is conditioning adherence (Sec.~\ref{sec:CA}) since volume fraction is provided as a conditional input to the model and calculated for generated topologies. 
    \item Load Violation and Floating Material: Load violation and floating material quantify whether the generated topology has material at the point of load application, and disconnected material, respectively. These are constraint satisfaction scores (Sec.~\ref{sec:scs}).
\end{enumerate}
All in all, this is a strong set of evaluation metrics capturing functional performance, constraint satisfaction, and conditioning. We note that each of these metrics is a point metric and can be summarized across a generated sample set as practitioners see fit. Papers commonly report mean, median, or proportion above some threshold.
\subsection{Other Metrics}
Though the previously discussed metrics quantify many important facets of DGM performance in optimal topology generation, we feel that several other metrics could also be informative. Specifically, we propose three additional metrics focused on diversity, novelty, and more nuanced constraint satisfaction scoring. 
\begin{enumerate}
    \item Proportion of Floating Material: While load violation is a binary constraint, floating material can be quantified using a scalar measuring the proportion of pixels that are disconnected from the topology. This continuous score may be more informative than a binary metric. Since the constraint boundary is at zero, this proportion is simply the distance to the constraint boundary metric. 
    \item Distance to SIMP:  Quantifying the (design space) novelty of generated topologies relative to their SIMP-generated counterparts can yield insight as to whether the model is memorizing SIMP-generated solutions or learning general structural optimization skills that may yield different results from SIMP. Models that are able to generate novel topologies may be able to seed SIMP to find superior solutions\footnote{A known limitation of SIMP is its susceptibility to local minima caused by nonconvexity.} even if their average compliance error is relatively high (or in rare cases find stronger solutions outright). In contrast, a model that mimics SIMP will seldom be able to find superior topologies even if it has a low compliance error. In many TO datasets, the only datapoint available for each discrete condition is the SIMP-generated solution, meaning that distance to SIMP is simply the conditional nearest datapoint metric. To calculate this score, we need a distance metric that works on topologies. As discussed earlier, embedding models trained on natural images may not offer meaningful distances, so we instead use pixel-wise distance, though Structural Similarity Index Metric~\cite{wang2004image} could be a valid choice.
    \item Topology Diversity: Just as we may like to understand how novel our DGM-generated topologies are from SIMP's topologies, we may seek to quantify how diverse these topologies are with respect to one another. This can tell us whether our model is capable of finding several families of strong solutions, increasing the likelihood that a strong design candidate is present in a large sample. We can use pixel distance to calculate a diversity matrix over a batch of samples generated under a single condition. We can then calculate DPP diversity as a score for the diversity of the generator.
\end{enumerate}

\subsection{Models}
We benchmark two versions of the state-of-the-art DGM for topology optimization, TopoDiff~\cite{maze2022topodiff}. TopoDiff is a Denoising Diffusion Probabilistic Model (DDPM) that trains using a similarity objective to mimic topologies generated by SIMP. Simultaneously, it uses classifier guidance during sampling to `push' otherwise invalid topologies away across constraint boundaries and into feasible regions. TopoDiff is conditioned on 2D stress field images that indicate the loading condition and assist in its training. Classically, these stress fields are calculated using Finite Element Analysis (FEA) to simulate stresses when the loads are applied to a solid material block. However, running FEA to generate stress fields on every input problem is time-consuming, so we train an alternative version of TopoDiff conditioned on kernel-based stress field estimates, as proposed in~\cite{giannone2023diffusing}. 

\subsection{Results}
\begin{figure}[!htb]
    \centering
    \includegraphics[width=0.99\textwidth]{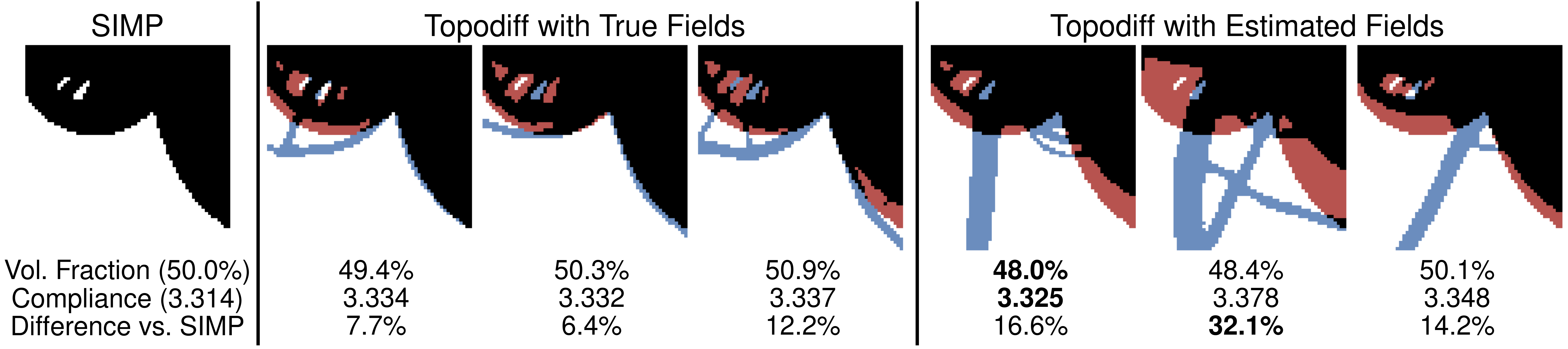}
    \caption{Sample topologies generated by two versions of TopoDiff compared against SIMP-generated topology. Three samples are generated using each model. Additional material is shown in blue while removed material is shown in red (relative to SIMP). SIMP's compliance and volume fraction are shown in parentheses. TopoDiff tends to generate more novel and diverse results when conditioned on kernel estimates of the stress fields. In this test problem, it generates the lightest and least compliant topology, all while being fairly novel relative to SIMP.}
    \label{fig:topology}
\end{figure}

The two variants of TopoDiff are benchmarked on the four original metrics from the paper, as well as the additional three that we propose. Models are tested on the first 300 conditions in TopoDiff's level 1 test data. We diffuse all samples over 100 steps. Unlike~\cite{giannone2023diffusing} and~\cite{maze2022topodiff}, we generate 10 samples per condition per model instead of one to score diversity. This adds additional complexity to the averaging process. Both point and set metrics must both be averaged over conditions but point metrics must additionally be averaged over samples generated for a single condition. The scores are tabulated in Table~\ref{tab:topotable}. The reported averaging technique for point metrics (`mean' or `median') is used to average over samples for a single condition, while mean averaging is used for all metrics (both point and set) over all conditions. Reported results are expected to differ slightly from~\cite{giannone2023diffusing} and~\cite{maze2022topodiff} since different instantiations of models were tested and only a subset of the test set was evaluated. Inference time, as reported in~\cite{giannone2023diffusing}, is also included~\footnote{We do not recalculate inference time and instead use published values from~\cite{giannone2023diffusing}.}. For qualitative examination, Figure~\ref{fig:topology} demonstrates several generated topologies using both approaches, emphasizing the difference between the generated topology and the SIMP-generated solution. 

\begin{table}[!htb]
\caption{Scores of two variants of TopoDiff. While the variant conditioned on true stress fields demonstrates superior target achievement, constraint satisfaction, and conditioning adherence, the variant conditioned on estimated fields exhibits higher novelty and diversity of generated solutions, in addition to being significantly faster during inference.}
\centering
\resizebox{0.8\textwidth}{!}{
\begin{tabular}{ccc}
\toprule
\textbf{Metric} & \begin{tabular}[c]{@{}c@{}}Topodiff -\\ \textbf{FEA Fields}\end{tabular} & \begin{tabular}[c]{@{}c@{}}Topodiff -\\ \textbf{Kernel Fields}\end{tabular}\\
\midrule
\begin{tabular}[c]{@{}c@{}}\textbf{Median Compliance Error} \\ (Distance to Target Boundary)\end{tabular} & \textbf{1.99\%} & 9.61\% \\
\begin{tabular}[c]{@{}c@{}}\textbf{Mean Volume Fraction Error}\\ (Conditioning Adherence)\end{tabular} & \textbf{1.49\%} & 1.69\% \\
\begin{tabular}[c]{@{}c@{}}\textbf{Mean Load Validity} \\ (Binary Constraint Satisfaction 1)\end{tabular} & \textbf{0.00\%} & 0.13\% \\
\begin{tabular}[c]{@{}c@{}}\textbf{Mean Floating Material} \\ (Binary Constraint Satisfaction 2)\end{tabular} & \textbf{6.60}\% & 8.97\% \\
\midrule
\begin{tabular}[c]{@{}c@{}}\textbf{Mean Floating Material Quantity} \\ (Distance to Constraint Boundary 2)\end{tabular} & \textbf{0.022\%} & 0.031\% \\
\begin{tabular}[c]{@{}c@{}}\textbf{Mean Distance to SIMP} \\ (Mean Novelty)\end{tabular} & 11.6\% & \textbf{17.3\%} \\
\begin{tabular}[c]{@{}c@{}}\textbf{Topology Diversity} \\ (Diversity)\end{tabular} & 7.144 & \textbf{5.87}\\
\midrule
\textbf{Mean Inference Time (sec.)} & 5.87 & \textbf{2.35} \\
\bottomrule
\end{tabular}
}
\label{tab:topotable}
\end{table}

\subsection{Discussion}
Since the FEA-generated strain fields are more accurate and informative than the kernel-estimated strain fields, we anticipate (correctly) that the FEA-generated fields yield superior target achievement (lower compliance error), higher constraint achievement (load validity, floating material), and lower volume fraction error (stronger conditioning adherence). Additionally, as the new floating material quantity metric indicates, generated designs are also closer to being constraint-satisfying when TopoDiff is conditioned on FEA fields. 

In addition to vastly improved inference time emphasized in~\cite{giannone2023diffusing}, a notable strength that was not emphasized is the higher novelty versus SIMP-generated results and the higher diversity among generated topologies using the kernel-conditioned TopoDiff variant. Specifically, the solutions generated differ from the SIMP solutions by almost 50\% more pixels than samples from the FEA field-conditioned model. Coupled with the significantly higher diversity, they have a much higher chance of re-seeding SIMP to find a stronger solution. This largely explains the significant performance improvements found by~\cite{giannone2023diffusing}, using generated samples to seed further re-optimization in SIMP. The kernel-estimated fields may generate not only faster samples but better samples for this task compared to the FEA fields. 

Through the selection of new metrics for DGMs in optimal topology generation, we have enabled insightful comparisons between models regarding the novelty and diversity of the samples they generate. These comparisons are highly relevant for many practical tasks, such as re-seeding TO solvers like SIMP to find better local optima. With this, we conclude our second case study and proceed to some final discussion on implications for engineering design. 

\section{Implications for Engineering Design}
Since the introduction of deep generative models in engineering design research, the overwhelming majority of proposed models have been trained exclusively to optimize for statistical similarity~\cite{regenwetter2022deep}. Since most of these models were adopted from natural image synthesis problems, it's no surprise that design researchers have similarly adopted statistical similarity, the commonplace class of metrics in the image synthesis domain. DGMs have shown immense promise in numerous design fields, such as topology optimization, airfoil design, photonic/phononic materials, molecular design, metamaterials, and many more~\cite{regenwetter2022deep}. However, as design researchers continue to adapt and enhance these models for design tasks, a similar need for modifications and new additions to evaluation metrics grows more pressing by the day.

\subsection{The Pitfall of Statistical Similarity}
Early in this paper, we attempted to illustrate why relying solely upon statistical similarity as a design evaluation metric can be misguided. In short, statistical similarity is simply insufficient for evaluating most designs, since it ignores design constraints, novelty, diversity, and performance, which are often primary concerns in a design setting. This limitation has been raised as a primary critique of DGMs in engineering. For example, Woldseth~\etal~\cite{woldseth2022use} demonstrate how DGMs applied to topology optimization fail in very simple cases when neglecting to account for performance or constraints due to the massive performance difference between some statistically similar designs. Furthermore, over-optimizing for statistical similarity can even be counterproductive and restrict the ability of a model to discover truly useful designs that exhibit superior properties to designs in the dataset. 

Since statistical similarity is still an important facet of DGM performance, we do not advocate for universal abstinence from similarity metrics. Instead, we encourage researchers to explore the nuance of similarity and think critically about their choice of metrics. Simultaneously, we hope that design researchers will look beyond statistical similarity to consider the wealth of other metrics at their disposal. 

\subsection{Selecting Appropriate Metrics}
What should practitioners consider beyond similarity? This typically requires a careful examination of the design problem at hand and depends on how practitioners seek to expand upon existing designs in the dataset. Some common goals are a greater level of diversity in design candidates, a set of designs that meet a challenging set of constraints, or a collection of designs that attain superior performance attributes to existing designs. We will briefly discuss several of the popular application areas and which types of metrics may be relevant for these problems. However, the exact choice of metrics often depends on the representation scheme, the use case, the cost of evaluation of metrics, and the DGMs used.

\paragraph{Molecular Design \& Drug Discovery} Molecular design problems typically require that designs adhere to a rigorous set of design constraints based on fundamental chemistry principles. Constraint satisfaction is, therefore, central to molecular design. Nonetheless, diversity and functional performance are typically important as well, as many practitioners seek to explore a wide range of designs that achieve desirable functional properties~\cite{bilodeau2022generative}. As the goal is often ``discovery'' of something useful and new, it is imperative to consider metrics related to novelty and performance. Not surprisingly, the ``rediscovery'' metric was also reported in molecule design literature to assess models.

\paragraph{Topology Optimization, Structural Design, and Metamaterials}
As we highlighted in the case study in Section~\ref{sec:TO}, TO and other related structural design problems tend to feature functional performance goals, constraints, and conditioning, which should each be quantified when evaluating DGMs. For image representation of topologies, structures, or metamaterials, pixel similarity is often used as the loss function. Researchers have shown that adding goals of performance (e.g., compliance) and constraints (e.g., manufacturability) significantly improves the usefulness of the generated designs.
Diversity metrics could also be useful to discover new structures, particularly in problems where DGM-generated topologies are used to warm-start classic topology optimization methods. 

\paragraph{Material and Microstructure Design} 
Generative models are often used to generate datasets of synthetic microstructure images to assist in microstructure characterization and reconstruction tasks~\cite{bostanabad2018computational}. This can, in turn, help establish process-structure-property links, which enable a variety of benefits, including inverse design of processes to attain desired material properties. Since realism is a primary goal in synthetic dataset generation, similarity metrics are typically favored. However, popular metrics such as FID and IS may incur heavy bias due to the domain gap between natural image datasets and microstructure scans. 

\paragraph{Product Design and Inverse Product Design}
Though product design is a fairly abstract categorization, inverse design involves `reverse engineering' a design given a set of desired properties and characteristics.  Since many inverse product design problems feature functional performance goals, both target achievement and general functional performance metrics are usually essential. Additionally, inverse design problems lend themselves to conditional models, making conditioning metrics particularly important. Creativity tends to be an important consideration in product design. It is often defined as a combination of novelty and quality (functional performance) of design. As a result, novelty metrics are used in many of the real-world product design problems tackled by DGMs in the current literature~\cite{yoo2021integrating, oh2019deep, regenwetter2022design}. 
% Constraints are generally present as well and should be evaluated when possible. 

\subsection{Responsible use of Evaluation Metrics}
Having access to a wealth of evaluation metrics provides researchers with a great deal of flexibility in showcasing the performance of their models. However, in such a young field with few established common practices, this flexibility may also cause confusion. For the good of the community, we encourage researchers to leverage as many relevant metrics as possible to showcase both the strengths and weaknesses of their models and provide a more comprehensive picture. Finally, we encourage researchers to be diligent in publishing any hyperparameters or evaluation metric settings and share their training datasets. To properly compare models and establish clear progression in the field, evaluation metric settings and training datasets must be fully transparent to be replicated. 

Finally, though this paper has focused on presenting metrics for the effective evaluation of DGMs, some of the proposed metrics can also be incorporated into the training of DGMs with little to no modification, as with the design target achievement index in~\cite{regenwetter2022design} and DPP diversity metric in~\cite{chen2021mopadgan}. When directly used as training objectives, validating performance using other metrics is important to ensure the model hasn't exploited its objectives. 

\section{Conclusion}
As deep generative models continue to expand their reach in engineering design communities, the increasing need for effective and standardized metrics grows more pronounced. This paper explored evaluation metrics for deep generative models in engineering design and presented a curated set of design-focused metrics to address this growing demand. We presented a wide variety of evaluation metrics focusing on realism, coverage, diversity, novelty, constraint satisfaction, performance, target achievement, and conditioning. We discussed which metrics to select when evaluating different facets of model performance and demonstrated the application of these metrics on easy-to-visualize problems. Finally, we presented a practical application of the methods discussed on a challenging bicycle frame design problem. We detailed how and why we selected appropriate metrics and described the performance of three deep generative models applied to the problem. 

In writing this paper, our overarching goal has been to call attention to the fallacies of statistical similarity metrics in engineering design problems, inspire practitioners to consider the many alternatives that we discuss, and provide them with the requisite knowledge to effectively apply these metrics to their design problems. We publicly release our datasets, models, and implementation code for the metrics used in our case studies at \url{decode.mit.edu/projects/metrics/} to facilitate easy adoption in different domains. We sincerely hope that researchers will use these resources to better understand their models and greatly improve their capabilities in engineering design tasks. 

\section{Acknowledgements}
The authors would like to acknowledge the MIT-IBM Watson AI Lab in supporting this research. They would also like to acknowledge Giorgio Giannone for his assistance with TopoDiff training. 

\bibliographystyle{elsarticle-num} 

\bibliography{bibliography}

\section{Appendix}
\subsection{Other Evaluation Methods and DGM Considerations}\label{sec:misc}
Thus far, we have discussed metrics to evaluate similarity, diversity, constraints, performance, and conditioning. Below, we briefly summarize two more types of metrics that may be important in certain design problems. A detailed, but not necessarily comprehensive list of other considerations for DGMs in design is included after this discussion. 

\subsubsection{Latent Disentanglement} 
Many DGMs synthesize designs by taking randomized inputs from some (usually multi-dimensional) latent variable space. Many design researchers have attempted to link latent variables with the physical properties of the generated designs~\cite{wang2020deep, chen2019synthesizing}, which shows promise in several design domains, such as functionally graded materials. Disentanglement can also make the generation process more interpretable, ideally serving as a tool for human designers to manually select or tune generated designs. Disentanglement metrics generally fall into one of three categories: Intervention-based metrics such as Z-diff~\cite{higgins2017beta}, predictor-based metrics such as attribute predictability score (SAP)~\cite{kumar2017variational}, and information-based metrics such as mutual information gap (MIG)~\cite{chen2018isolating}. We refer readers to reviews like~\cite{zaidi2020measuring} and~\cite{do2019theory} for more detailed discussions on disentanglement metrics. 

\subsubsection{Human Evaluation}
Though automated evaluation metrics are often the most practical, they seldom provide as valuable of an analysis as people. Human evaluation approaches can roughly be divided into crowdsourced evaluation frameworks and expert evaluation frameworks, where the primary tradeoff is cost versus evaluation quality.

\paragraph{Crowdsourced Evaluation}
Crowdsourced evaluation frameworks are common in computer vision fields since untrained humans typically suffice for determining the `realism' of images in computer vision problems. Metrics like Human Eye Perceptual Evaluation (HYPE)~\cite{zhou2019hype}, for example, quantify how easily humans can discern between real and fake samples. In contrast, designs may be difficult to properly evaluate since they are not always represented in a visual medium and may have infeasibilities or inefficiencies that require engineering expertise to discern. 

\paragraph{Expert Evaluation}
Expert evaluation methods are often used in various applications of deep generative models to assess the quality of the generated samples. Domain experts evaluate the samples based on specific criteria such as realism, coherence, and relevance. The scores are then averaged over a set of generated samples to compare different models. This approach is considered to be a gold standard for evaluating the quality of generated samples, but is time-consuming and costly. We refer the reader to a review~\cite{miller2021should} on these metrics in design literature for an in-depth discussion.

% Alternatively, human experts can also be used to provide feedback to the model during the generation process, allowing the model to learn from its mistakes and improve its performance over time.

% Another approach to evaluate deep generative models is the Consensual Assessment Technique (CAT)~\cite{amabile1982social}, which is commonly used in psychology design methodology literature. CAT ratings are a subjective measure of creativity based on the opinions of experts. It involves having experts rate designs on a scale on criteria such as originality, feasibility, and potential impact. CAT ratings can provide valuable insights into the perceived level of creativity of a design. However, one of the key challenges of using these metrics is scalability, as deep generative models are capable of generating millions of samples in a short time, while the cost of evaluating each sample is very high. In principle, most metrics (e.g., ranking, scoring) used to rate human-generated designs could also be used to rate DGM-generated designs. Similarly, many metrics in design, such as SVS~\cite{shah2003metrics} could be adapted for DGM output. We refer the reader to a review~\cite{miller2021should} on these metrics in design literature for an in-depth discussion. With this, we conclude our discussion on miscellaneous metrics for DGM evaluation and move on to a case study on the application of evaluation metrics in a real engineering design problem.

\subsubsection{Miscellaneous DGM evaluation areas} 
There are many other areas of DGM performance that include but are not limited to those listed below. Since these areas are not particularly unique to engineering design, we deem them out of scope and do not directly discuss them.
\begin{enumerate}
    \item Cost: Computational time, memory, and energy costs of training, inference and deployment.
    \item Robustness: The model's resilience against noise, adversarial attacks, or perturbations in input data.
    \item Transferability and data-efficiency: The capability of the model to generalize and perform well on unseen tasks with limited original data or fine-tuning.
    \item Stability and consistency: The degree to which the model produces similar results when provided with similar inputs and its ability to avoid issues like mode collapse or gradient vanishing/exploding during training.
    \item Bias and fairness: The resilience of the model to biased data.
    \item Privacy: The ease with which revealing information about training data can be reconstructed from a trained model.
\end{enumerate}

\subsection{Calculating Distances between Designs}
Many evaluation metrics necessitate the ability to calculate the distance between designs. This task is often nontrivial. Below, we provide some methods to calculate distances in various data modalities, as well as some strategies to leverage embeddings to calculate distances. 

\begin{enumerate}
    \item \textbf{Continuous variables}: Common distance metrics include Euclidean distance and Manhattan distance. These distances are sensitive to the relative scaling between parameters. As such, care must be taken to appropriately scale data. Cosine similarity is another common approach to measuring `distance' between variables.
    \item \textbf{Ordinal Discrete Variables}: For ordinal discrete variables (such as the number of teeth on a gear), $L_1$, $L_2$, and $L_\infty$ are all valid metrics.
    \item \textbf{Categorical Variables}: For discrete variables with no sequential significance (i.e., categorical data), one-hot encoding can transform the data into boolean data, to which the aforementioned distance methods apply. When directly working in categorical spaces, Hamming distance is a widely used metric. 
    \item \textbf{Pixels}: Calculating distances directly on images is challenging, largely due to their high dimensionality. Depending on the nature of the images, directly calculating pixel distance using a simple method like L2 may be a valid approach. For example, this may work on topologies represented as images, while on photographs of vehicles, pixel distance is unlikely to reflect similarity in the content of images. As a strong alternative, distances between images can be calculated using Structural Similarity (SSIM)~\cite{wang2004image}. Recently, calculating distances between images is most commonly done using learned embeddings.
    \item \textbf{3D geometry}: For many common 3D geometry representation methods like point clouds, meshes, and rasterized curves, point-set distances like Chamfer and Hausdorff distance are commonly used. We refer the reader to papers like~\cite{nguyen2021point} for further details on metrics used in computer graphics. 
    \item \textbf{Text}: Directly calculating distances between pieces of textual data is challenging. While simple categorical distances can be calculated on tokenized text data using cosine distance, learned text embeddings are usually the preferred approach to calculating distances on text data. 
    \item \textbf{Graphs}: Measuring distance between graphs (networks) is a common problem known as the `network comparison' problem. Typically methods depend on the type of graph (directed vs. undirected, weighted vs. unweighted, known vs. unknown nodes, etc.). We refer readers interested in calculating distances between graphs to~\cite{tantardini2019comparing}.
    \item \textbf{Multimodal Data} Calculating distances between designs represented using multimodal data is nontrivial and inherits the challenges of the individual modalities. Though this problem requires further research, there are a few strategies that researchers can select from. A simple approach takes the weighted sum of distances across different modalities but requires tuning of weights. Another approach constructs an embedding from the multimodal data over which to measure distances~\cite{radford2021learning}. 
\end{enumerate}

While calculating distances in the original data modality is often viable, a common alternative approach is to find features of the data and then calculate distances between those features. These features are often learned using a machine learning model instead of being hand-picked. These approaches are very common when working with images~\cite{khrulkov2020hyperbolic, faghri2017vse++}, 3D geometry~\cite{dai2018siamese}, text~\cite{devlin2018bert, cer2018universal}, and graphs~\cite{cai2018comprehensive}. However, learned embeddings suffer from a potential risk of poor generalizability. For example, a breadth of literature has documented how image embeddings struggle to generalize even across similar natural image datasets~\cite{morozov2020self, rosca2017variational, barratt2018note, zhou2019hype}. In general, great care must be taken when using learned embeddings to calculate distances, regardless of the modality of the original data.

\begin{figure}[!htb]
    \centering
    \includegraphics[width=0.7\textwidth]{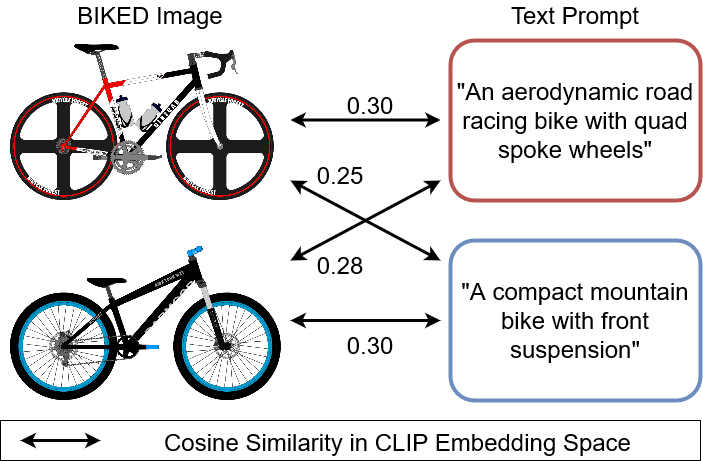}
    \caption{Cosine similarity between bicycle images from~\cite{regenwetter2022biked} and text prompts in CLIP embedding space. Shared embeddings allow for distance calculation between datapoints of different modalities. }
    \label{fig:clip}
\end{figure}

When working with multimodal data, calculating distances can be challenging. Shared embeddings are a viable approach to calculating distances between datapoints of different modalities. For example, Contrastive Language-Image Pretraining (CLIP)~\cite{radford2021learning} embeddings map images and text into a common embedding space in which distances can be calculated, as showcased in Figure~\ref{fig:clip}. 

\subsection{Model and Hyperparameter Settings for Evaluation Experiments}
We present the training settings for all models tested on the synthetic datasets and all evaluation metrics used. 
\paragraph{GAN and cGAN Models}
\begin{enumerate}
    \item Discriminator and generator models are 2 hidden-layer neural networks with 100 parameters with leaky ReLU activation with slope coefficient $\alpha=0.2$.
    \item The latent vectors are four-dimensional.
    \item Discriminator and generator optimizers are Adam~\cite{kingma2014adam} with learning rate $1e-3$.
    \item The models are trained over 5000 randomized batches of 32 samples. 
\end{enumerate}
\paragraph{VAE Model}
\begin{enumerate}
    \item Encoder and decoder models are 3 hidden-layer neural networks with 100 parameters with ReLU activation.
    \item The latent dimension is 4.
    \item The model is trained for 1000 epochs of batches of 100.
    \item the model's optimizer is Adam with a learning rate $1e-3$.
\end{enumerate}
\paragraph{MO-PaDGAN Model}
\begin{enumerate}
    \item Discriminator an generator architectures and optimizers are the same as in the GAN model. latent dimension and training epochs are also unchanged.
    \item All training parameters are identical to~\cite{chen2021mopadgan} except $\gamma_0 = 5$ and $\gamma_1=2$. 
\end{enumerate}

\paragraph{Experiment 1: Evaluating statistical similarity on 2D data (Sec.~\ref{simdemo}}
\begin{enumerate}
    \item Nearest datapoint, nearest generated sample, and rediscovery use Euclidean distance.
    \item F1, F10, F0.1, and AUC-PR use 20 clusters, angle resolution of 1000, and 10 clustering runs.
    \item ML efficacy uses an auxiliary regression task of predicting the objectives used in Sec.~\ref{perfdemo}. A K-nearest-neighbors regressor with $K=5$ is used and evaluated using the coefficient of determination ($R^2$). 
\end{enumerate}
\paragraph{Experiment 2: Evaluating design exploration on 2D data (Sec.~\ref{novdemo}}
\begin{enumerate}
    \item Nearest datapoint and inter-sample distance use Euclidean distance.
    \item DPP Diversity uses a subset size of 10 and an exponentiation parameter of 0.1.
\end{enumerate}
\paragraph{Experiment 3: Evaluating design constraints on 2D data (Sec.~\ref{constdemo}}
\begin{enumerate}
    \item Predicted constraint satisfaction uses K-nearest-neighbors with $K=5$.
    \item Nearest Invalid Datapoint uses Euclidean Distance.
\end{enumerate}
\paragraph{Experiment 4: Evaluating design quality on 2D data  (Sec.~\ref{perfdemo}}
\begin{enumerate}
    \item The hypervolume reference point is selected as the 99th quantile performance value from the dataset.
    \item The Pareto-optimal set for KNO1 is given as $x+y=.4705$.
    \item The target for all target-related scores is set as (0.5,0.5). Objectives are weighted equally. 
    \item DTAI parameters are $\alpha=\beta=[1,1]$.
\end{enumerate}
\paragraph{Experiment 5: Evaluating conditioning adherence on 2D data (Sec.~\ref{conddemo}}
\begin{enumerate}
    \item cVAE and cGAN architectures are identical to unconditional counterparts.
    \item Both marginal and conditional F0.1 and F10 use 20 clusters, and angle resolution of 1000, and 10 clustering runs.
    \item Conditional scores used the nearest 10\% of the dataset as the constraint-satisfying subset. 
    \item Condition adherence and reconstruction use mean squared error. 
    \item Condition reconstruction uses K-nearest-neighbors regression with $K=5$.
\end{enumerate}
\subsection{Model and Hyperparameter Settings for Bicycle Frame Design Problem}
\paragraph{Evaluation Metric Settings}
\begin{enumerate}
    \item Nearest datapoint, nearest generated sample, and inter-sample distance use Euclidean distance.
    \item F1 score uses 20 clusters, angle resolution of 1000, and 10 clustering runs.
    \item DPP Diversity uses a subset size of 10 and an exponentiation parameter of 0.1.
    \item The target for all target-related scores is set as 0.75 quantile performance value in each objective across the FRAMED dataset.
    \item DTAI parameters are $\alpha=[1, 1, 1, 1, 2, 1, 1, 4, 4, 3], \beta=[4, 4, 4, 4, 4, 4, 4, 4, 4, 2]$.
\end{enumerate}
\paragraph{Model Parameters}
\begin{enumerate}
    \item Discriminator and generator models for the GAN, MO-PaDGAN, and DTAI-GAN are 2 and 3 hidden-layer neural networks respectively with 128 parameters and leaky ReLU activation with slope coefficient $\alpha=0.2$.
    \item All PaDGAN and DTAI-GAN training parameters are identical to~\cite{chen2021mopadgan} except $\gamma_0 = 5$ and $\gamma_1=0.5$.
    \item The latent vectors are four-dimensional.
    \item Discriminator and generator optimizers are Adam~\cite{kingma2014adam} with learning rate $1e-4$.
    \item The models are trained over 50000 randomized batches of 8 samples. 
    \item MO-PaDGAN and DTAI-GAN used differentiable performance surrogate regressors for performance evaluations of generated designs. DTAI-GAN used a differentiable constraint satisfaction surrogate classifier for constraint satisfaction likelihoods. Performance prediction and constraint satisfaction prediction networks were optimal neural networks identified through Bayesian optimization in~\cite{regenwetter2022framed}. 
\end{enumerate}

\subsection{Public Code and Evaluation Metrics Package}
In our publicly released codebase we have provided the following:
\begin{enumerate}
    \item 6 model architectures
    \item 23+ evaluation metrics
    \item 14+ synthetic datasets
    \item Code to generate score reports with confidence bounds
    \item Code to generate training convergence plots
    \item Code to generate plots of generated samples on 2D data, as shown in the paper
    \item Code to generate animations of generated samples throughout the training process on 2D data
\end{enumerate}
These features are organized as an easy-to-use package that can be quickly used to evaluate other researchers' models. 
\end{document}